\documentclass{article}

\usepackage{microtype}
\usepackage{graphicx}
\usepackage{overpic}
\usepackage{subcaption}
\usepackage{colortbl}
\usepackage{multirow}
\usepackage{xcolor}
\definecolor{lightgray}{gray}{0.93}
\usepackage{makecell} 
\usepackage[table]{xcolor}

\usepackage{booktabs}
\usepackage{array} 
\usepackage{algorithm}
\usepackage{algorithmic}
\usepackage{amsmath}
\usepackage{wrapfig}
\usepackage{url}
\usepackage{xurl}
\usepackage[unicode,linktocpage,backref=page,hidelinks]{hyperref}

\usepackage{amsmath,amsfonts,bm}

\def\eqref#1{equation~\ref{#1}}

\def\1{\bm{1}}

\def\vk{{\bm{k}}}

\def\vu{{\bm{u}}}
\def\vv{{\bm{v}}}

\def\vx{{\bm{x}}}
\def\vy{{\bm{y}}}
\def\vz{{\bm{z}}}

\DeclareMathAlphabet{\mathsfit}{\encodingdefault}{\sfdefault}{m}{sl}
\SetMathAlphabet{\mathsfit}{bold}{\encodingdefault}{\sfdefault}{bx}{n}

\definecolor{CorrectMatch}{HTML}{E8E8E8}   
\definecolor{NewNeighbor}{HTML}{B8C9E8}    
\definecolor{WrongClass}{HTML}{E8CBB8}     
\definecolor{Forgotten}{HTML}{D4C4E0}      

\definecolor{PredCorrect}{HTML}{90EE90}    
\definecolor{PredWrong}{HTML}{FFB6C6}      
\usepackage{tikz}
\DeclareRobustCommand{\rbox}[2]{%
  \tikz[baseline=(X.base)]\node[
    fill=#1,
    rounded corners=1.5pt,
    inner xsep=2pt,
    inner ysep=0.5pt
  ] (X) {#2};%
}

\AtBeginDocument{%
    \renewcommand*{\PrintBackRefs}[1]{}%
}
\let \backreforig \backref
\renewcommand*{\backref}[1]{[Referenced on page \backreforig{#1}]}

\newcommand{\err}[1]{\scalebox{0.72}{$\pm\,#1$}}
\usepackage[main, final, nonatbib]{neurips_2026}

\usepackage{apacite}
\usepackage[round]{natbib}

\usepackage[utf8]{inputenc} 
\usepackage[T1]{fontenc}    
\usepackage{amsfonts}       
\usepackage{nicefrac}       
\usepackage{amssymb}
\usepackage{mathtools}
\usepackage{amsthm}

\usepackage[capitalize,noabbrev]{cleveref}

\theoremstyle{plain}

\theoremstyle{definition}

\theoremstyle{remark}

\usepackage[textsize=tiny]{todonotes}

\title{Models Designed to Forget: \\ Machine Unlearning via Key Deletion}

\author{
  Sonia Laguna\thanks{Correspondence to \texttt{slaguna@inf.ethz.ch}. $^\dagger$Shared senior authorship.}, 
  Jorge da Silva Gonçalves,  
  Moritz Vandenhirtz, 
  Alain Ryser,\\
  \textbf{ Irene Cannistraci$^\dagger$,  Julia E.~Vogt$^\dagger$ }\\
  Department of Computer Science, ETH Zurich, Switzerland
}

\begin{document}

\maketitle

\vspace*{-0.4cm}
\begin{abstract}
Machine unlearning for vision models is rapidly becoming a practical requirement, driven by privacy regulations, data errors, and the need to remove harmful or corrupted training images. Despite this, most existing approximate unlearning methods tackle the problem from a post-hoc perspective. They attempt to erase the influence of targeted samples through parameter updates that typically require access to the full training data. This creates a mismatch with real deployment scenarios where unlearning requests can be anticipated, revealing a fundamental limitation of post-hoc approaches. We motivate \textit{unlearning by design}, a novel paradigm for approximate methods in which models are directly trained to support forgetting as an inherent architectural capability. We instantiate this idea with Machine UNlearning via KEY deletion (MUNKEY), a memory-augmented transformer that decouples instance-specific memorization from model weights. Here, unlearning corresponds to removing the instance-identifying key, enabling zero-shot forgetting without weight updates or access to the original samples or labels. Across natural image benchmarks, fine-grained visual recognition, and medical datasets, MUNKEY outperforms all post-hoc baselines. Our results establish that \textit{unlearning by design} enables fast, deployment-oriented unlearning while preserving predictive performance.\looseness-1 
\vspace*{-0.1cm}
\end{abstract}

\section{Introduction}
\label{sec:intro}
 As machine learning systems are increasingly deployed in high-stakes domains, the ability to \emph{unlearn}~\citep{cao2015towards, bourtoule2021machine} specific training samples on demand has become essential. This necessity is driven both by the need to remove harmful, biased, or corrupted data and by the legal imperative to comply with privacy regulations, such as the European Union’s GDPR ``right to be forgotten''~\citep{mantelero2013eu} and the California Consumer Privacy Act~\citep{ccpa2018}. Because neural networks are capable of memorizing training samples~\citep{arpit2017closer}, the central challenge of machine unlearning is to surgically excise the influence of targeted data points. This process must eliminate the data, often verified via membership inference attacks~\citep{shokri2017membership}, without degrading the model's performance on the remaining ``retained'' and test data.\looseness-1

Current unlearning approaches generally fall into two categories: exact and approximate. Exact unlearning ensures that all influence of the target data is removed, mostly by retraining the model from scratch, a baseline that is often computationally prohibitive. To address this, approximate unlearning has emerged as a more scalable alternative, seeking to diminish the impact of forget-set data to a negligible level by updating the model's parameters. However, existing approximate methods~\citep{nguyen2025survey} are strictly \emph{post-hoc}: they rely on computationally intensive weight updates at the time of unlearning, such as gradient ascent, distillation, or saliency-based corrections, often requiring access to most of the data samples and corresponding labels. 

Beyond computational cost and data availability, the reliance on post-hoc interventions highlights a fundamental structural flaw in standard architectures: knowledge entanglement. When a neural network is trained, instance-specific details and generalizable features are fused into a monolithic set of static weights~\citep{arpit2017closer}, making it challenging and inefficient to remove the influence of a single instance \emph{a posteriori}.\looseness-1 


\begin{figure}[t]
    \centering
     \includegraphics[width=\linewidth]
     {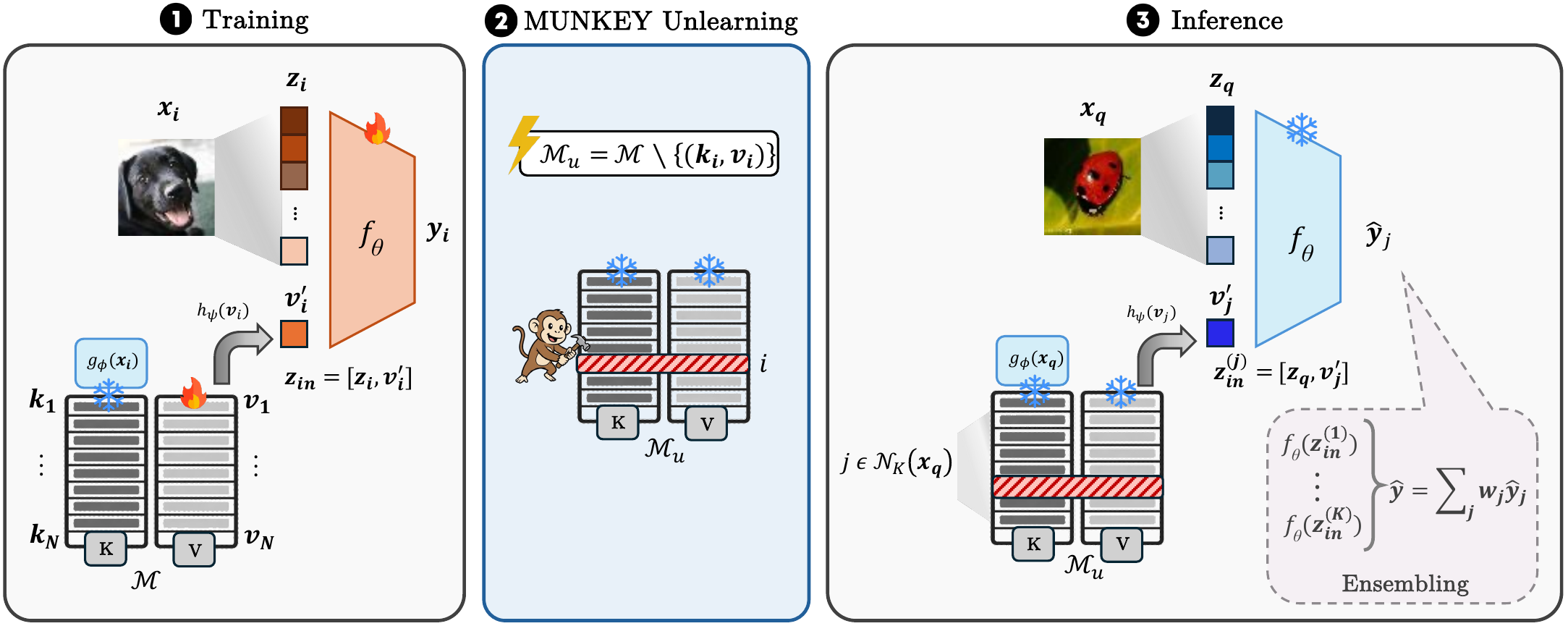}

    \caption{\textbf{Overview of MUNKEY.} \textbf{(1) Training:} The learnable exemplar token $\vv_i$ from the memory bank $\mathcal{M}$ is concatenated to image tokens $\vz_i$ to predict the label $\vy_i$. \textbf{(2) Unlearning} results in a simple instance-specific key deletion in $\mathcal{M}$. \textbf{(3) Inference:} For a query $x_q$, the model retrieves $K$-nearest neighbor tokens from the updated memory $\mathcal{M}_u$ and ensembles their predictions for the final output.\looseness-1}
    \label{fig:placeholder}
    \vspace*{-0.1cm}
\end{figure}

\textbf{Can the paradigm be shifted towards \emph{unlearning by design}? }In many applications, such as clinical studies where patients withdraw consent or consumer platforms where users opt out of data sharing, deletion requests are an anticipated part of the model lifecycle. By accounting for these requests \emph{a priori}, we can design models with unlearning in mind, where forgetting is streamlined and efficient.\looseness-1 
\vspace*{-0.15cm}
\paragraph{Machine UNlearning via KEY deletion}\hspace{-0.3cm}(MUNKEY) bridges this gap. We introduce MUNKEY, a vision method that externalizes image-specific information to separate it structurally from static model parameters. By associating each training instance with a learnable exemplar token, MUNKEY transforms unlearning from a complex optimization problem into a simple set-theoretic operation: deleting entries from an external memory. This method offers two primary advantages: \textit{(i)} efficiency, enabling ``zero-shot'' near-instantaneous unlearning without weight updates, and \textit{(ii)} privacy-centric deployment, requiring only instance identifiers rather than raw samples or labels. Crucially, decoupled memories have been studied to improve predictive performance, perception, or few-shot efficiency~\citep{jia2022visual, long2022retrieval, sandler2022fine}, while MUNKEY explicitly designs the memory for model adaptation and targeted data deletion. Our work builds on recent results showing that visual memories can scale to billions of samples~\citep{geirhostowards}. By repurposing memory-augmented structures for the ``inverse'' task of forgetting, we enable a new class of models that are not only high-performing but are natively compliant with the dynamic requirements of data privacy and integrity.\looseness-1

Finally, while a portion of the literature focuses on class-level removal~\citep{zhang2025targeted, gandikota2023erasing}, we target the general case of approximate instance-level unlearning as it more directly addresses individual privacy and consent. This work targets the vision domain, building upon the extensive body of literature~\citep{ebrahimpour2025amun,hsuunseen,khalilcoun25,fan2024salun,leeruago} that focuses on unlearning for vision. Across benchmarks spanning natural, fine-grained, and medical imagery, MUNKEY outperforms state-of-the-art post-hoc baselines, offering a superior balance of efficiency and robustness against membership inference attacks without compromising utility. The decoupling of representation (backbone) from editable knowledge (memory) provides additional advantages, such as dynamic scalability, allowing new data to be incorporated by expanding the memory bank without backbone retraining~\citep{li2024evcap}; or interpretability, where decisions are traceable to specific retrieved memory entries or “prototypes”~\citep{geirhostowards}.\looseness-1

\paragraph{Contributions.} Our contributions are threefold: \textit{(i) Paradigm Shift:} We propose \emph{Unlearning by Design} in approximate methods, shifting from costly post-hoc corrections to intrinsic architectural modularity. By anticipating deletions during training, unlearning can be streamlined. \textit{(ii) Algorithmic Framework:} We introduce MUNKEY, a vision unlearning method that leverages memory-augmented models for instance-level forgetting by formalizing a way to decouple instance-specific information from model weights, which enables optimization-free unlearning. \textit{(iii) Empirical Validation:} Extensive evaluations across natural, large-scale, and medical vision datasets show that MUNKEY outperforms state-of-the-art post-hoc baselines in efficiency, utility, and detectable membership signal.\looseness-1
\vspace*{-0.05cm}
\section{Related Work}
\label{sec:related_work}
\vspace*{-0.1cm}
\paragraph{Machine Unlearning.} 
Machine unlearning removes the influence of specific training samples from a trained model  \citep{cao2015towards}. While early work explores exact approaches with strong theoretical guarantees  \citep{bourtoule2021machine}, the prohibitive computational cost of these methods has shifted recent research to more efficient approximate alternatives \citep{nguyen2025survey}.

A prominent direction in approximate unlearning leverages fine-tuning to induce catastrophic forgetting (CF)~\citep{mccloskey1989catastrophic}. This concept has motivated more targeted strategies such as CF-$k$ and Exact Unlearning-$k$ (EU-$k$) \citep{goel2022towards}, which restrict unlearning to the final $k$ layers, as well as gradient ascent approaches such as NegGrad and NegGrad+ \citep{kurmanji2023towards}, or supervising on held-out sets~\citep{regun}. To improve model utility, prior works have proposed selectively adjusting specific layers or parameters: $\ell_1$-sparse unlearning \citep{jia2023model} promotes sparsity via $\ell_1$ regularization, while SSD \citep{foster2024fast} and SalUN \citep{fan2024salun} update only important parameters identified through Fisher information and gradient-based saliency, respectively.
Meanwhile, other approaches focus on data manipulation and model distillation. These include utilizing label perturbation \citep{graves2021amnesiac}, adversarial examples \citep{ebrahimpour2025amun,cha2024learning}, student-teacher methods such as Bad-Teacher \citep{chundawat2023can} and its later refinement SCRUB \citep{kurmanji2023towards}, as well as recent contrastive-learning-based approaches via CoUn~\citep{khalilcoun25}.\looseness-1

Despite their differences, these methods share a limitation: they operate post-hoc, requiring access to the forget set, retain set, labels, or a combination thereof to modify a trained model. Approaches like SISA~\citep{bourtoule2021machine} hint at unlearning by design by anticipating deletion through data partitioning, but remain exact methods and require optimization over affected shards upon deletion. Similarly, some approximate methods, including zero-shot~\citep{chundawat2023zero} approaches that remove the need for retain data, and sparsity-based methods that prepare models for later removal~\citep{lin2023erm}, still rely on post-hoc optimization.
We instead advocate for a paradigm shift where approximate unlearning is anticipated and enforced during training. We propose a memory-augmented method designed for instance-level removal: by externalizing instance-specific information into a memory, our approach enables targeted removal without destabilizing the core model parameters.\looseness-1
\vspace*{-0.125cm}
\paragraph{External Memory.} External memory mechanisms have been widely explored to augment neural networks with explicit storage and retrieval capabilities. Architectures such as Neural Turing Machines \citep{graves2014neural}, Memory Networks \citep{DBLP:journals/corr/WestonCB14, sukhbaatar2015end}, and Memorizing Transformers \citep{wu2022memorizing} traditionally rely on decoupling memory storage from the model's primary parameters. In parallel, complementary work focuses on associative memory, where learnable structures are embedded directly into network components to store patterns or knowledge~\citep{sandler2022fine, jia2022visual}. However, while both utilize memory to enhance predictive performance~\citep{li2024evcap} or facilitate few-shot learning, we repurpose the external memory framework for a distinct objective: efficient data localization and removal. To achieve this, we introduce architectural and training choices that encourage the model to route deletable information through external storage rather than entangle it in the backbone. To the best of our knowledge, this is the first external-memory framework designed specifically for machine unlearning. An extended discussion of related work is presented in Appendix~\ref{app:related_work_extended}.\looseness-1
\vspace*{-0.05cm}
\section{Methods}
\label{sec:methods}

\vspace*{-0.1cm}
\subsection{Problem Formalization: Machine Unlearning}
Let $f_\theta: \mathcal{X} \to \mathcal{Y}$ be a neural network parameterized by $\theta$. The model is initially optimized over a training dataset $\mathcal{D} = \{(\vx_i, \vy_i)\}_{i=1}^N$ via Empirical Risk Minimization with a task-specific loss, in our case cross-entropy for the classification task. Machine unlearning considers a request to remove a subset of data $\mathcal{D}_f \subset \mathcal{D}$, the \textit{forget set}, while preserving performance on $\mathcal{D}_r = \mathcal{D} \setminus \mathcal{D}_f$, the \textit{retain set}. The gold standard is the \textit{Retrain} model $f_r$, optimized exclusively on $\mathcal{D}_r$. The objective of an approximate machine unlearning method is to achieve a behavior such that $f_{\theta_u} \approx f_r$ without the prohibitive cost of full retraining, ensuring both \textit{Forgetting} (zero influence of $\mathcal{D}_f$) and \textit{Utility} (high performance on $\mathcal{D}_r$ and on the test set $\mathcal{D}_t$). In this work, we employ the Vision Transformer (ViT)~\citep{dosovitskiy2020image} as our architectural backbone, followed by a linear classification head to form $f_\theta$. A ViT partitions an input image $x_i$ into a sequence of $n$ flattened patches and projects them onto a latent space $\{\vz_i^{1}, \ldots, \vz_i^{n}\}$ through an embedding layer. These patches are then processed via attention layers. In this work, we will refer to $f_\theta$ as the model after the embedding layer. 

\vspace*{-0.05cm}
\subsection{MUNKEY: Machine UNlearning via KEY Deletion}
\label{sec:munkey}
\setcounter{footnote}{0}
We propose MUNKEY\footnote{The code is publicly available at \url{https://github.com/sonialagunac/MUNKEY}.} to reformulate unlearning as a modular data-deletion task within a memory bank $\mathcal{M}$. Unlike traditional post-hoc approximate methods, MUNKEY is constructed for \textit{unlearning by design} by conditioning classification on a joint sequence of image patch embeddings and retrieved exemplar tokens, as illustrated in~\Cref{fig:placeholder}. This structural separation encourages the influence of a specific sample to be confined to its corresponding entry in $\mathcal{M}$, turning unlearning from an optimization problem over $\theta$ into a deletion operation over $\mathcal{M}$. 
\vspace{-0.05cm}
\paragraph{External Exemplar Memory.}
MUNKEY exploits the permutation-invariant modularity of transformers by augmenting the standard patch sequence with a decoupled external exemplar token. By injecting instance-specific knowledge directly into the input sequence rather than the static weights $\theta$, the burden of memorization is shifted to the externalized memory, enabling unlearning via simple deletion of the memory entry.
To operationalize this decoupling, an explicit exemplar memory bank $\mathcal{M} = \{(\vk_i, \vv_i)\}_{i=1}^N$ is defined for all $\vx_i \in \mathcal{D}$. For each instance, $\vk_i=g_\phi(\vx_i) \in \mathbb{R}^{d_k}$ is a fixed key extracted via a frozen encoder $g_\phi$, while $\vv_i \in \mathbb{R}^{d_e}$ is a learnable exemplar token that is treated as a learnable parameter. $g_{\phi}$ can be instantiated with any suitable visual encoder. We evaluate both general-purpose pre-trained encoders, as well as encoders trained on same-distribution held-out data. The latter setting ensures that MUNKEY also operates under a controlled scenario where the encoder has not seen any samples that may later be forgotten, addressing cases where key-encoder leakage or overlap with general-purpose pre-training data is a concern. To ensure compatibility, an adapter layer $h_\psi$ projects the exemplar token into the  ViT’s latent space, treating it as an auxiliary input token. 

\vspace*{-0.05cm}
\paragraph{Training with Stochastic Pathway Dropout.}
To train MUNKEY, the image patch embeddings $\vz_i$ are concatenated with the learnable projected exemplar token $\vv'_i = h_\psi(\vv_i)$ to form the input sequence $\vz_{in}$. Subsequently, $\vz_{in}$ is passed through $f_\theta$ to predict the class, thereby not requiring any adaptation of the standard ViT architecture.
A fundamental challenge in such dual-stream processing is \textit{modality dominance}~\citep{wang2020makes, hussen2020modality}, where the model over-relies on the most predictive input. In our case, depending on the dataset nature, the backbone can over-rely on high-capacity exemplar tokens and fail to learn robust features from the raw image patches, or vice versa.\looseness-1 

To enforce a balanced information flow and prevent pathway dominance, a stochastic pathway dropout is applied to both image patch embeddings, $\vz_i$, and projected exemplar tokens, $\vv'_i$. A stochastic masking vector $\mathbf{m} = [\gamma_{img}, \gamma_{tok}] \in \{0, 1\}^2$ follows a categorical distribution,  where $P(\mathbf{m} = [0, 1]) = p_i$, $P(\mathbf{m} = [1, 0]) = p_t$,  $P(\mathbf{m} = [1, 1]) = 1 - p_i - p_t$, and $P(\mathbf{m} = [0, 0]) = 0$, with $p_i$ and $p_t$ being the probabilities for image and token dropout, respectively. Specifically, the dropout operator $\mathcal{P}$ is formulated for each pathway using the sampled mask component $\gamma$ as: $\mathcal{P}(\vu, \emptyset, \gamma) = \gamma \vu + (1-\gamma) \emptyset$, for any input token $\vu$ and a pathway-specific globally learned null token $\emptyset$, i.e. $\emptyset_{img}$ or $\emptyset_{tok}$.\looseness-1

\vspace*{-0.05cm}
\paragraph{Inference via Nearest Neighbors.}
At inference time, the specific exemplar token $\vv_q$ for an unseen query image $\vx_q$ is unavailable. To resolve this, the query key $\vk_q = g_\phi(\vx_q)$ is first extracted from the frozen encoder. Subsequently, the $K$ nearest neighbors $\mathcal{N}_K(\vx_q)$ are retrieved from the memory bank $\mathcal{M}$ based on the cosine similarity of $\vk_q$ with the stored keys  $\{\vk_i\}_{i=1}^N$. Note that following the efficiency paradigms of large-scale retrieval-augmented models~\citep{geirhostowards}, this operation can be scaled efficiently through approximate nearest neighbor indexing, i.e. ScaNN~\citep{guo2020accelerating}. The final prediction is an ensemble of the model's output for each of the neighbors. More concretely, for each retrieved exemplar token $\{\vv_j\}_{j \in \mathcal{N}_K(\vx_q)}$, an input sequence $\vz_{in}^{(j)}$ as $[\vz_q, \vv'_j]$ is constructed. The final output $\hat{\vy}$ is computed as a weighted average of the logits: $\hat{\vy} = \sum_{j \in \mathcal{N}_K(\vx_q)} w_j \cdot f_\theta\big(\vz_{in}^{(j)}\big)$, where the weights $w_j$ are determined by a softmax over key similarities: $w_j = \frac{\exp(\text{sim}(\vk_q, \vk_j)/\tau)}{\sum_{l \in \mathcal{N}_K(\vx_q)} \exp(\text{sim}(\vk_q, \vk_l)/\tau)}$. Here, $\text{sim}(\cdot, \cdot)$ denotes cosine similarity and $\tau$ a temperature parameter.\looseness-1 

\paragraph{Retrieval-based Regularization at Training.}
To ensure the model generalizes to retrieved context at inference, where the specific $\vv_q$ for a test sample is absent, this state is simulated during training. 
For each training sample $(\vx_i, \vy_i)$, a fixed key $\vk_i = g_\phi(\vx_i)$ is extracted using the frozen encoder. These keys populate the memory $\mathcal{M}$ and serve as the basis for neighborhood discovery. To construct the augmented input sequence $\vz_{in}$,  
either the instance-specific token $\vv'_i$, or the neighboring $\vv'_n$ are selected. $\vv'_n$ is the result of simulating the inference state by retrieving a randomly varying number $K'$ of neighbor tokens $\{\vv_j\}_{j \in \mathcal{N}_{K'}(\vx_i)}$ from $\mathcal{M}\setminus\{(\vk_i,\vv_i)\}$ and aggregating them via $\text{Agg}(\cdot)$ into $\vv_n$, where $\text{Agg}(\cdot)$ stands for a similarity-weighted average operation. Let $r \sim \text{Bernoulli}(p_r)$ be an indicator variable for the retrieval state. Combined with the masking vector, the final input sequence at training time $\vz_{in}$ is defined as:\looseness-1
\begin{equation}
    \vz_{in}(\vx_i,\mathbf{m},r)= \left[ \mathcal{P}(\vz_i, \emptyset_{img}, \gamma_{img}) \, , \, \mathcal{P}\big( (1-r)\vv'_i + r \vv'_n , \emptyset_{tok}, \gamma_{tok} \big) \right].
\end{equation}
 The model parameters $\{\theta, \psi\}$, and exemplar tokens $\{\vv_i\}_{i=1}^N$ are optimized jointly  by minimizing the expected masked training objective:
\begin{equation}
    \min_{\theta,\psi,\{\vv_i\}_{i=1}^N}
    \;
    \mathbb{E}_{(\vx_i,\vy_i)\sim\mathcal{D}}
    \;
    \mathbb{E}_{\mathbf{m}\sim p(\mathbf{m}),\, r \sim \mathrm{Bern}(p_r)}
    \left[
    \ell\left(f_\theta(\vz_{in}(\vx_i,\mathbf{m},r)),\vy_i\right)
    \right],
\end{equation} 
where the gradient $\nabla_{\vv_j} \ell$ is clipped to $0$ for all $j \in \mathcal{N}_{K'}(\vx_i)$ when the retrieval branch is active. This ensures the backbone learns to aggregate context, preventing neighboring tokens from drifting to accommodate sample-specific gradients. This objective can be viewed as a variational pathway-masking formulation: $\mathbf{m}$ and $r$ induce a distribution over information pathways, and training minimizes the expected risk under it. By requiring stable predictions when either pathway is partially suppressed, stochastic masking discourages brittle reliance on a single source. Consequently, it imposes an implicit information bottleneck on the exemplar pathway, pushing $\vv_i$ to encode complementary information rather than act as an unconstrained lookup. We note that, as in approximate unlearning, this decoupling should be interpreted as an architectural inductive bias rather than a strict formal guarantee. Following standard practice in approximate methods, we evaluate forgetting empirically using membership inference attacks as a proxy for residual membership signal.\looseness-1

\paragraph{Unlearning.}
Unlike traditional methods that rely on iterative optimization to erase information, unlearning in MUNKEY is reduced to a simple set-theoretic operation: $\mathcal{M}_u = \mathcal{M} \setminus \{(\vk_i, \vv_i) \mid (\vx_i, \vy_i) \in \mathcal{D}_f\}$. By removing the forget set entries from the memory bank, the model is rendered incapable of retrieving the specific tokens $v_i$ associated with the requested data. As the model is trained to rely only on the externalized memory for instance-specific details, this deletion results in ``zero-shot'' forgetting. 
All the steps are summarized in~\Cref{alg:munkey}, Appendix~\ref{app:algorithm}.\looseness-1

\paragraph{Pathway Sensitivity and Model Selection.}

To prevent \textit{pathway collapse}, where the model either ignores the memory (poor unlearning) or degenerates into a lookup table (poor generalization), we introduce the Pathway Sensitivity Score (${P_s}$). ${P_s}$ quantifies the balance of information flow by evaluating performance when one pathway is replaced by the learned null tokens $\emptyset$. We define the modality-specific accuracies $\mathcal{A}_{img}$ and $\mathcal{A}_{tok}$ by evaluating the model on the ablated input sequences $\vz_{in}^{(img)} = [\vz_i, \emptyset_{tok}]$ and $\vz_{in}^{(tok)} = [\emptyset_{img}, \vv'_i]$, respectively, and $\mathcal{A}_{both}$ as the standard joint accuracy using the complete sequence $\vz_{in} = [\vz_i, \vv'_i]$. The sensitivity score is then formulated as ${P_s} = \frac{|\mathcal{A}_{img} - \mathcal{A}_{tok}|}{\mathcal{A}_{both} + \epsilon}$, where $\epsilon$ is a small constant added for numerical stability. 

By utilizing the global null tokens $\emptyset$ learned during training, the sensitivity evaluation remains within the model's learned distribution. We define $\xi$ as a threshold that bounds the disparity between pathway contributions, preventing dominance. $P_s$ enables multi-stage hyperparameter selection for $p_i$ and $p_t$: (i) a ``health check'' filters configurations where $P_s \ge \xi$, and (ii) following the ``rewind'' procedure proposed in~\citet{kurmanji2023towards}, we prioritize models that minimize the validation utility gap, $\Delta_{\mathrm{val}}$, i.e., the accuracy difference between validation and disjoint proxy forget sets. We refer to Appendix~\ref{app:sensitivity} for extended details. Therefore, among candidate hyperparameter configurations $\lambda = (p_i,p_t)$, we select\looseness-1
\begin{equation}
 \lambda^*
    =
    \arg\min_{\lambda}
    \; \Delta_{\mathrm{val}}(\lambda)
    \quad
    \text{s.t.}
    \quad
    P_s(\lambda) < \xi. 
\end{equation}

\section{Experimental Setup}

\label{sec:experiments}

\paragraph{Datasets.} We evaluate MUNKEY across natural and medical imaging benchmarks. For natural images, we utilize CIFAR-10, CIFAR-100~\citep{krizhevsky2009learning}, and large-scale Tiny ImageNet~\citep{deng2009imagenet}. For the medical domain, we incorporate three datasets from the high-resolution MedMNIST v2~\citep{yang2023medmnist}: BloodMNIST~\citep{acevedo2020dataset}, with 17,092 images across 8 classes for fine-grained cellular morphology; PathMNIST~\citep{kather2019predicting}, a large-scale histology dataset with 107,180 images and 9 classes; and DermaMNIST~\citep{tschandl2018ham10000}, with 10,015 images across 7 imbalanced skin-disease classes, with multi-source noise, mimicking real-world clinical prevalence where common conditions outnumber rare pathologies. Due to its difficulty, DermaMNIST is featured as a primary benchmark alongside CIFAR-10, while evaluations for the remaining datasets are provided in Appendix~\ref{app:extra-results} and~\ref{app:variants}. Dataset details are included in Appendix~\ref{app:datasets}.\looseness-1

\paragraph{Baselines.} We compare against a diverse suite of eleven baselines whose mechanisms, while originally designed for CNNs, apply directly to Transformers. Recognizing ViTs as the current gold standard for vision, we implement both MUNKEY and all baselines using a ViT-Tiny backbone to ensure a fair comparison and to bridge the reporting gap in transformer-based unlearning. Additional model scales, e.g., ViT-Base, are evaluated in Appendix~\ref{app:scaling}. We include: \textit{Retrain from Scratch}  with $\mathcal{D}_r$, the oracle for exact unlearning on both the standard ViT-Tiny, referred to as \textit{Retrain (T)}, and the MUNKEY-specific model, \textit{Retrain (M)}. Catastrophic Forgetting (\textit{CF}), \textit{CF-k}, and \textit{EU-k}~\citep{goel2022towards} fine-tune the model on $\mathcal{D}_r$ while freezing or restoring parts of the network. Similarly, \textit{NegGrad+}~\citep{kurmanji2023towards} and \textit{Amnesiac}~\citep{foster2024fast}, utilize gradient ascent and label noise to induce forgetting. To represent weight-level manipulation, we include \textit{$\ell_1$-sparse} as in~\citet{jia2023model}, encouraging parameter sparsity. For distillation-based frameworks, we use \textit{SCRUB}~\citep{kurmanji2023towards}. Moreover, optimization-free parameter dampening is represented by \textit{SSD}~\citep{foster2024fast}, which utilizes Fisher Information to identify and damp relevant parameters without iterative training. We also evaluate \textit{AMUN}~\citep{ebrahimpour2025amun}, which generates adversarial samples to force forgetting, and CoUn~\citep{khalilcoun25}, a contrastive-learning-based method. Finally, we include a memory-based reference using \textit{KNN}~\citep{fix1985discriminatory}, as a point of comparison for retrieval-based inference. Further implementation details are provided in Appendix~\ref{app:baselines}. \looseness-1

\paragraph{Metrics.} Following prior work~\citep{patel2025learning, fan2024salun, ebrahimpour2025amun}, we evaluate our model using four primary metrics and an average gap to assess the trade-off between utility and forgetting: Test Accuracy (TA) and Retain Accuracy (RA) for model utility, and Forget Accuracy (FA) and Membership Inference Attack (MIA) to measure forgetting. For privacy evaluation, we use the SOTA Robust MIA~\citep{zarifzadeh2024low}. This attack compares the confidence score of a sample against reference distributions obtained from 64 shadow models. As in prior work~\citep{ebrahimpour2025amun}, we compute the confidence values using the Soft-Margin Taylor expansion of the Softmax with temperature $T=2$ and a ratio threshold $\gamma$ of 1. We report the AUC for distinguishing training samples from unseen samples. An MIA AUC close to 0.5 on the forget set $\mathcal{D}_f$ indicates that the unlearned model does not retain a detectable membership signal. We aggregate these results into the Average Gap (Avg Gap), which measures the mean absolute difference between each of these four metrics and their respective counterparts in the "gold-standard" oracles. For architectural fairness, baselines are compared against a retrained ViT-Tiny, \textit{Retrain (T)}, while our method is compared against a retrained MUNKEY, \textit{Retrain (M)}. \looseness-1
\paragraph{Implementation Details.} We evaluate MUNKEY under two forgetting regimes: a standard $10\%$ removal of random samples benchmark and a more conservative $2\%$, which we posit represents a more realistic scenario where data deletion requests are frequent but scarce. More extreme $50\%$ removal is shown in Appendix~\ref{app:extra-results}. We additionally evaluate full-class, class-imbalanced, and single-instance deletion requests on CIFAR-10 in Appendix~\ref{app:forget-structure}. To maintain label distribution and prevent the removal of entire classes, mainly in imbalanced datasets or with a high number of classes, we utilize stratified random sampling of the forget sets for all datasets except CIFAR-10. All experiments are conducted over three independent seeds. The training pipeline and dynamics are standardized across all base models and datasets, and further details are reported in Appendix~\ref{app:imp_details}. We provide ablations of all different subcomponents of MUNKEY in Appendix~\ref{app:ablations} to isolate the effect of each one of them on the final performance. In particular, we study the trade-off between performance and dimension $d_e$ of the exemplar token ($\vv_i \in \mathbb{R}^{d_e}$), the effect of the number of neighbors $K$ used at inference time, or the choice of key encoder. Beyond unlearning quality and runtime, there are system-level considerations associated with storing and querying the external memory bank. We discuss these practical costs, including memory requirements and retrieval overhead, in Appendix~\ref{app:system-costs}. \looseness-1
\section{Results}
\label{sec:results}

To validate the proposed \textit{unlearning by design}, we evaluate MUNKEY across visual domains, addressing three questions: \textit{(i) Decoupled Memorization:} Do instance-specific tokens enable zero-shot unlearning? \textit{(ii) Information Encoding:} Does the memory capture granular signals beyond class labels in representations? \textit{(iii) Architectural Interaction:} How does pathway aggregation impact performance?\looseness-1

\begin{table}[h!]
\centering
\vspace*{-0.2cm}
\caption{Comparison of machine unlearning performance across random forget rates (10\% and 2\%) on DermaMNIST and CIFAR-10. We present the accuracy on the test (TA), retain (RA), and forget (FA) sets, as well as the membership inference attack (MIA) and average gap values. Across all tables in the manuscript, when applicable, we \textbf{bold} the lowest Avg Gap and \underline{underline} the second lowest.\looseness-1}
\vspace{0.4em}
\label{tab:combined-unlearning-comparison}
\setlength{\tabcolsep}{2.4pt}
\fontsize{6}{6.7}\selectfont
\begin{tabular}{@{}l cccc >{\columncolor{lightgray}[0pt][0pt]}c cccc >{\columncolor{lightgray}[0pt][0pt]}c@{}}
\toprule
& \multicolumn{5}{c}{\textbf{Random Forget (10\%)}} & \multicolumn{5}{c}{\textbf{Random Forget (2\%)}} \\
\cmidrule(lr){2-6} \cmidrule(lr){7-11}
\textbf{Method} & \textbf{TA} & \textbf{RA} & \textbf{FA} & \textbf{MIA} & \textbf{Avg Gap $\downarrow$} & \textbf{TA} & \textbf{RA} & \textbf{FA} & \textbf{MIA} & \textbf{Avg Gap $\downarrow$} \\
\midrule
\multicolumn{11}{c}{\textbf{DermaMNIST}} \\
\midrule
Retrain (T) & $76.46_{\err{0.22}}$ & $100.0_{\err{0.00}}$ & $74.86_{\err{1.77}}$ & $48.88_{\err{0.69}}$ & -- & $76.49_{\err{0.26}}$ & $100.0_{\err{0.00}}$ & $75.71_{\err{2.02}}$ & $50.72_{\err{1.60}}$ & -- \\
Retrain (M) & $81.25_{\err{0.66}}$ & $99.70_{\err{0.08}}$ & $80.29_{\err{1.41}}$ & $50.04_{\err{1.05}}$ & -- & $81.80_{\err{0.04}}$ & $99.75_{\err{0.05}}$ & $81.43_{\err{2.10}}$ & $49.36_{\err{2.86}}$ & -- \\
\midrule
KNN & $74.60_{\err{0.24}}$ & $85.54_{\err{0.14}}$ & $74.81_{\err{1.23}}$ & $49.94_{\err{0.90}}$ & -- & $75.01_{\err{0.07}}$ & $85.84_{\err{0.09}}$ & $74.52_{\err{0.67}}$ & $50.32_{\err{3.30}}$ & -- \\
CF & $75.28_{\err{0.98}}$ & $97.50_{\err{1.55}}$ & $94.48_{\err{2.23}}$ & $64.41_{\err{1.96}}$ & $9.71_{\err{0.99}}$ & $76.38_{\err{0.28}}$ & $97.50_{\err{1.16}}$ & $95.48_{\err{1.21}}$ & $61.20_{\err{1.80}}$ & $8.22_{\err{0.90}}$ \\
CF-k & $76.94_{\err{0.42}}$ & $99.86_{\err{0.13}}$ & $99.90_{\err{0.13}}$ & $67.10_{\err{3.73}}$ & $10.97_{\err{1.05}}$ & $76.96_{\err{0.43}}$ & $99.86_{\err{0.13}}$ & $100.0_{\err{0.00}}$ & $66.73_{\err{1.38}}$ & $10.23_{\err{0.74}}$ \\
EU-k & $76.89_{\err{0.58}}$ & $99.60_{\err{0.23}}$ & $99.52_{\err{0.18}}$ & $69.27_{\err{0.23}}$ & $11.47_{\err{0.51}}$ & $76.97_{\err{0.47}}$ & $99.63_{\err{0.22}}$ & $99.76_{\err{0.34}}$ & $68.54_{\err{1.16}}$ & $10.68_{\err{0.73}}$ \\
$\ell_1$-sparse & $75.81_{\err{0.45}}$ & $92.36_{\err{1.80}}$ & $86.81_{\err{1.17}}$ & $57.69_{\err{1.13}}$ & $7.26_{\err{0.78}}$ & $75.61_{\err{0.67}}$ & $89.46_{\err{3.01}}$ & $80.24_{\err{1.21}}$ & $54.02_{\err{0.48}}$ & $4.81_{\err{1.06}}$ \\
NegGrad+ & $76.26_{\err{0.37}}$ & $96.31_{\err{1.03}}$ & $83.48_{\err{1.25}}$ & $55.83_{\err{1.14}}$ & $4.87_{\err{0.69}}$ & $75.73_{\err{0.96}}$ & $98.08_{\err{0.42}}$ & $74.05_{\err{2.99}}$ & $48.63_{\err{1.79}}$ & \underline{$1.61_{\err{1.12}}$} \\
Amnesiac & $75.56_{\err{0.23}}$ & $98.30_{\err{0.85}}$ & $82.00_{\err{2.38}}$ & $47.01_{\err{1.11}}$ & \underline{$2.93_{\err{0.73}}$} & $75.89_{\err{0.29}}$ & $98.92_{\err{0.17}}$ & $69.76_{\err{1.47}}$ & $36.59_{\err{1.82}}$ & $5.44_{\err{0.88}}$ \\
SCRUB & $73.97_{\err{1.81}}$ & $76.02_{\err{2.48}}$ & $74.68_{\err{1.23}}$ & $49.62_{\err{0.25}}$ & $6.85_{\err{0.96}}$ & $76.38_{\err{0.76}}$ & $99.55_{\err{0.53}}$ & $86.94_{\err{1.70}}$ & $57.51_{\err{1.00}}$ & $4.65_{\err{0.85}}$ \\
SSD & $75.07_{\err{2.08}}$ & $92.07_{\err{10.2}}$ & $89.73_{\err{12.9}}$ & $63.29_{\err{8.28}}$ & $9.65_{\err{4.67}}$ & $72.58_{\err{5.56}}$ & $87.87_{\err{16.3}}$ & $78.52_{\err{30.3}}$ & $60.98_{\err{9.33}}$ & $7.28_{\err{9.07}}$ \\
AMUN & $76.46_{\err{0.69}}$ & $98.33_{\err{0.94}}$ & $95.05_{\err{0.64}}$ & $63.70_{\err{0.27}}$ & $9.17_{\err{0.59}}$ & $76.46_{\err{0.69}}$ & $97.99_{\err{0.91}}$ & $98.57_{\err{1.01}}$ & $65.95_{\err{1.01}}$ & $10.03_{\err{0.79}}$ \\
CoUn
& $79.50_{\err{0.35}}$ & $99.80_{\err{0.05}}$ & $87.24_{\err{1.33}}$ & $54.92_{\err{1.29}}$ & $5.42_{\err{0.67}}$
& $79.50_{\err{0.61}}$ & $99.81_{\err{0.06}}$ & $89.52_{\err{1.47}}$ & $55.11_{\err{2.04}}$ & $5.35_{\err{0.92}}$ \\
\textbf{MUNKEY} & $81.06_{\err{0.44}}$ & $99.71_{\err{0.08}}$ & $84.00_{\err{1.22}}$ & $51.19_{\err{1.19}}$ & $\mathbf{1.27}_{\err{0.64}}$ & $81.23_{\err{0.53}}$ & $99.71_{\err{0.10}}$ & $85.48_{\err{2.76}}$ & $50.96_{\err{2.97}}$ & $\mathbf{1.57}_{\err{1.22}}$ \\

\midrule
\multicolumn{11}{c}{\textbf{CIFAR-10}} \\
\midrule
Retrain (T) & $78.13_{\err{0.64}}$ & $91.62_{\err{0.55}}$ & $77.57_{\err{0.83}}$ & $49.77_{\err{0.32}}$ & -- & $78.87_{\err{1.02}}$ & $91.83_{\err{0.61}}$ & $79.60_{\err{0.86}}$ & $50.30_{\err{0.78}}$ & -- \\
Retrain (M) & $92.38_{\err{0.06}}$ & $99.93_{\err{0.03}}$ & $92.05_{\err{0.23}}$ & $49.90_{\err{0.34}}$ & -- & $92.59_{\err{0.20}}$ & $99.93_{\err{0.02}}$ & $92.83_{\err{0.54}}$ & $50.67_{\err{2.02}}$ & -- \\
\midrule
KNN & $91.06_{\err{0.05}}$ & $94.56_{\err{0.06}}$ & $91.02_{\err{0.13}}$ & $50.16_{\err{0.10}}$ & -- & $91.28_{\err{0.08}}$ & $94.58_{\err{0.02}}$ & $92.10_{\err{0.28}}$ & $51.19_{\err{0.66}}$ & --\\
CF & $77.66_{\err{0.50}}$ & $91.74_{\err{0.48}}$ & $87.25_{\err{1.04}}$ & $55.86_{\err{0.16}}$ & $4.09_{\err{0.44}}$ & $78.53_{\err{0.95}}$ & $91.97_{\err{0.91}}$ & $87.10_{\err{0.59}}$ & $56.38_{\err{1.35}}$ & $3.52_{\err{0.65}}$ \\
CF-k & $79.60_{\err{0.63}}$ & $94.30_{\err{0.55}}$ & $94.16_{\err{0.61}}$ & $58.93_{\err{0.40}}$ & $7.48_{\err{0.41}}$ & $79.53_{\err{0.60}}$ & $94.30_{\err{0.55}}$ & $94.53_{\err{0.29}}$ & $59.51_{\err{0.71}}$ & $6.82_{\err{0.50}}$ \\
EU-k & $79.44_{\err{0.72}}$ & $94.22_{\err{0.55}}$ & $94.00_{\err{0.65}}$ & $58.87_{\err{0.30}}$ & $7.36_{\err{0.42}}$ & $79.36_{\err{0.64}}$ & $94.20_{\err{0.58}}$ & $94.63_{\err{0.12}}$ & $59.16_{\err{0.69}}$ & $6.69_{\err{0.50}}$ \\
$\ell_1$-sparse & $52.80_{\err{0.55}}$ & $53.47_{\err{0.75}}$ & $52.97_{\err{0.86}}$ & $50.79_{\err{0.79}}$ & $22.28_{\err{0.48}}$ & $52.70_{\err{0.73}}$ & $53.79_{\err{0.89}}$ & $53.70_{\err{2.00}}$ & $50.70_{\err{0.77}}$ & $22.63_{\err{0.74}}$ \\
NegGrad+ & $77.29_{\err{0.27}}$ & $91.08_{\err{0.42}}$ & $83.13_{\err{0.89}}$ & $53.65_{\err{0.32}}$ & \underline{$2.71_{\err{0.41}}$} & $77.03_{\err{0.48}}$ & $90.72_{\err{0.74}}$ & $72.03_{\err{1.06}}$ & $48.62_{\err{0.75}}$ & $3.05_{\err{0.57}}$ \\
Amnesiac & $78.31_{\err{0.35}}$ & $92.17_{\err{0.75}}$ & $85.67_{\err{1.05}}$ & $52.51_{\err{0.21}}$ & $2.89_{\err{0.46}}$ & $77.77_{\err{0.29}}$ & $91.11_{\err{0.68}}$ & $76.97_{\err{1.21}}$ & $44.95_{\err{0.86}}$ & $\underline{2.45_{\err{0.59}}}$ \\
SCRUB & $76.51_{\err{0.78}}$ & $85.31_{\err{0.71}}$ & $83.72_{\err{0.94}}$ & $54.18_{\err{0.47}}$ & $4.62_{\err{0.48}}$ & $77.74_{\err{1.19}}$ & $89.83_{\err{1.11}}$ & $85.61_{\err{0.84}}$ & $56.03_{\err{1.31}}$ & $3.72_{\err{0.70}}$ \\
SSD & $78.54_{\err{1.08}}$ & $93.29_{\err{0.71}}$ & $92.84_{\err{0.96}}$ & $58.91_{\err{0.43}}$ & $6.62_{\err{0.52}}$ & $79.02_{\err{1.06}}$ & $93.63_{\err{0.62}}$ & $94.00_{\err{0.22}}$ & $58.75_{\err{0.78}}$ & $6.20_{\err{0.55}}$ \\
AMUN & $77.71_{\err{0.46}}$ & $91.25_{\err{0.56}}$ & $86.85_{\err{3.46}}$ & $55.90_{\err{1.94}}$ & $4.05_{\err{1.05}}$ & $77.71_{\err{0.46}}$ & $90.89_{\err{0.94}}$ & $87.27_{\err{3.74}}$ & $56.37_{\err{2.08}}$ & $3.96_{\err{1.18}}$ \\
CoUn
& $80.68_{\err{0.44}}$ & $95.72_{\err{0.01}}$ & $91.32_{\err{0.54}}$ & $55.75_{\err{0.34}}$ & $6.60_{\err{0.36}}$
& $81.01_{\err{0.53}}$ & $95.84_{\err{0.01}}$ & $92.03_{\err{0.17}}$ & $55.16_{\err{0.43}}$ & $5.86_{\err{0.45}}$ \\
\textbf{MUNKEY} & $92.63_{\err{0.20}}$ & $99.94_{\err{0.01}}$ & $92.53_{\err{0.52}}$ & $51.40_{\err{0.44}}$ & $\textbf{0.56}_{\err{0.21}}$ & $92.69_{\err{0.21}}$ & $99.94_{\err{0.01}}$ & $93.57_{\err{0.26}}$ & $50.05_{\err{2.13}}$ & $\textbf{0.38}_{\err{0.75}}$ \\

\bottomrule
\vspace*{-0.225cm}
\end{tabular}
\end{table}
\paragraph{Decoupled Memorization: Forgetting and Utility Results.}
\Cref{tab:combined-unlearning-comparison} reports unlearning performance in two datasets across two random forget rates, 10\% and 2\%. Across all evaluations, MUNKEY consistently achieves the lowest Avg Gap, which quantifies the trade-off between model utility and unlearning faithfulness. Results on four additional datasets, full-class and class-imbalanced forget sets, and larger forget rates, i.e., 50\%, are provided in Appendix~\ref{app:extra-results}, confirming that these performance improvements generalize to diverse tasks. We include CIFAR-10 and DermaMNIST in our primary analysis to present two distinct trends. On CIFAR-10, where visual distributions align closely with the ImageNet-based features of our frozen key encoder, MUNKEY achieves predictive accuracies significantly higher than all post-hoc baselines. These results demonstrate that when the key encoder captures the underlying data manifold effectively, we can leverage these features for superior predictive performance while maintaining robust ``black-box'' unlearning, as evidenced by the MIA scores. We emphasize that the key encoder does not constitute an unfair advantage as each method's unlearning performance is evaluated against its respective gold standard, i.e., Retrain (T) or Retrain (M). Shifting to DermaMNIST, where medical morphologies differ significantly from the pre-training data of the key encoder, we still outperform baselines, not only in Avg Gap, but also in predictive performance. This indicates that the method also prevails in settings where the key encoder's pre-training does not align with the dataset. Overall, high forget accuracy does not necessarily indicate unsuccessful unlearning, as forgotten samples can still be classified through features shared with retained data, just as unseen test samples. Indeed, Retrain (M), which never trains on the forget set, also achieves high forget accuracy, while MUNKEY remains close to its oracle in both forget accuracy and membership inference.

To isolate the contribution of the pre-trained key encoder, Appendix~\ref{app:key-encoder} includes a controlled setting where the encoder is trained on a same-dataset held-out 20\% split. This highlights a flexible design choice: when a pre-trained encoder is available, it can be directly leveraged; otherwise, or if a potential overlap between encoder and forgotten samples is a concern, an encoder trained on disjoint data remains sufficient. We further explore the effect of other key encoders, showing that MUNKEY remains effective under alternative self-supervised~\citep{oquab2024dinov2} and multimodal~\citep{radford2021learning} pre-trained models.\looseness-1 

In both datasets, while pure retrieval classifiers, i.e., KNN, achieve good MIA scores, their predictive performance is lower. This disparity shows that MUNKEY’s image pathway boosts generalization beyond simple memory retrieval, a critical requirement for applications where both high utility and reliable unlearning are paramount. Beyond our results, we observe performance variability among baselines. While recent contrastive, distillation, and adversarial-based methods are generally competitive, they are occasionally outperformed by simpler gradient ascent (NegGrad+) or label-noising strategies. These fluctuations suggest that unlearning techniques originally optimized for CNNs may not seamlessly transfer to ViTs, highlighting a need for more specialized research in unlearning within transformer-based architectures.
We further compare against a RAC-inspired~\citep{long2022retrieval} retrieval-fusion baseline in Appendix~\ref{app:retrieval-fusion}, showing that an editable retrieval component alone does not achieve comparable unlearning performance without deletion-oriented training.\looseness-1

\begin{wraptable}{h}{0.40\columnwidth}
\vspace{-0.05cm}
\centering
    \caption{Runtime efficiency, full timings in~\Cref{tab:times-stdev}. \textit{Unlearn} is post-training deletion time. \textit{Train/Ep.} denotes per-epoch training time for the base models. \looseness-1}    
\label{tab:runtime-summary}
\scriptsize
\setlength{\tabcolsep}{1pt}
\begin{tabular}{l l cc}
\toprule
\textbf{Method} & \textbf{Cost} & \textbf{DermaMNIST} & \textbf{CIFAR-10} \\
\midrule
Standard ViT & Train/Ep. & $24.7$s & $174.6$s \\
MUNKEY & Train/Ep. & $25.9$s & $178.5$s \\
\midrule
Fastest baseline & Unlearn & $23.1$s & $105.8$s \\
Slowest baseline & Unlearn & $313.4$s & $1940.5$s \\
\textbf{MUNKEY} & Unlearn & $\mathbf{\approx 0}$s & $\mathbf{\approx 0}$s \\
\bottomrule
\end{tabular}
\vspace{-0.23cm}
\end{wraptable}

\vspace{-0.2cm}
Beyond end-to-end MIA, Appendix~\ref{app:privacy_components} studies the membership signal of the isolated backbone by removing the exemplar pathway at evaluation time. Predictive performance drops, confirming that the memory pathway contributes to utility, while the membership inference signal remains close to random ($\approx 0.5$), indicating low detectable instance-specific leakage in the backbone alone. Finally, a core advantage of the \textit{unlearning by design} paradigm is the elimination of iterative post-hoc optimization.~\Cref{tab:runtime-summary} shows runtime efficiency across methods, with the detailed per-method timings included in~\Cref{tab:times-stdev}. Post-hoc baselines span a wide unlearning runtime range, where even the fastest remains slower than MUNKEY, that requires O(1) computations, hence negligible time ($\approx 0$s). Forgetting is reduced to simply dropping the forgotten samples' key-value pairs from the memory bank, rendering the process near-instantaneous even at scale. We also report the per-epoch training time for both standard ViTs and MUNKEY. The results indicate that the overhead of maintaining and learning the exemplar memory bank is minimal. Given that this specialized training occurs only once in a compute-rich environment, the marginal increase in training time is a negligible cost for the permanent benefit of zero-shot unlearning throughout the model's deployment lifecycle. In Appendix~\ref{app:system-costs} we include an overview of system requirements, i.e., additional memory cost.\looseness-1 
\vspace{0.15cm}
\begin{figure}[h!]
    \centering
    \begin{subfigure}[t]{0.49\linewidth}
        \centering
        \begin{overpic}[width=\linewidth]{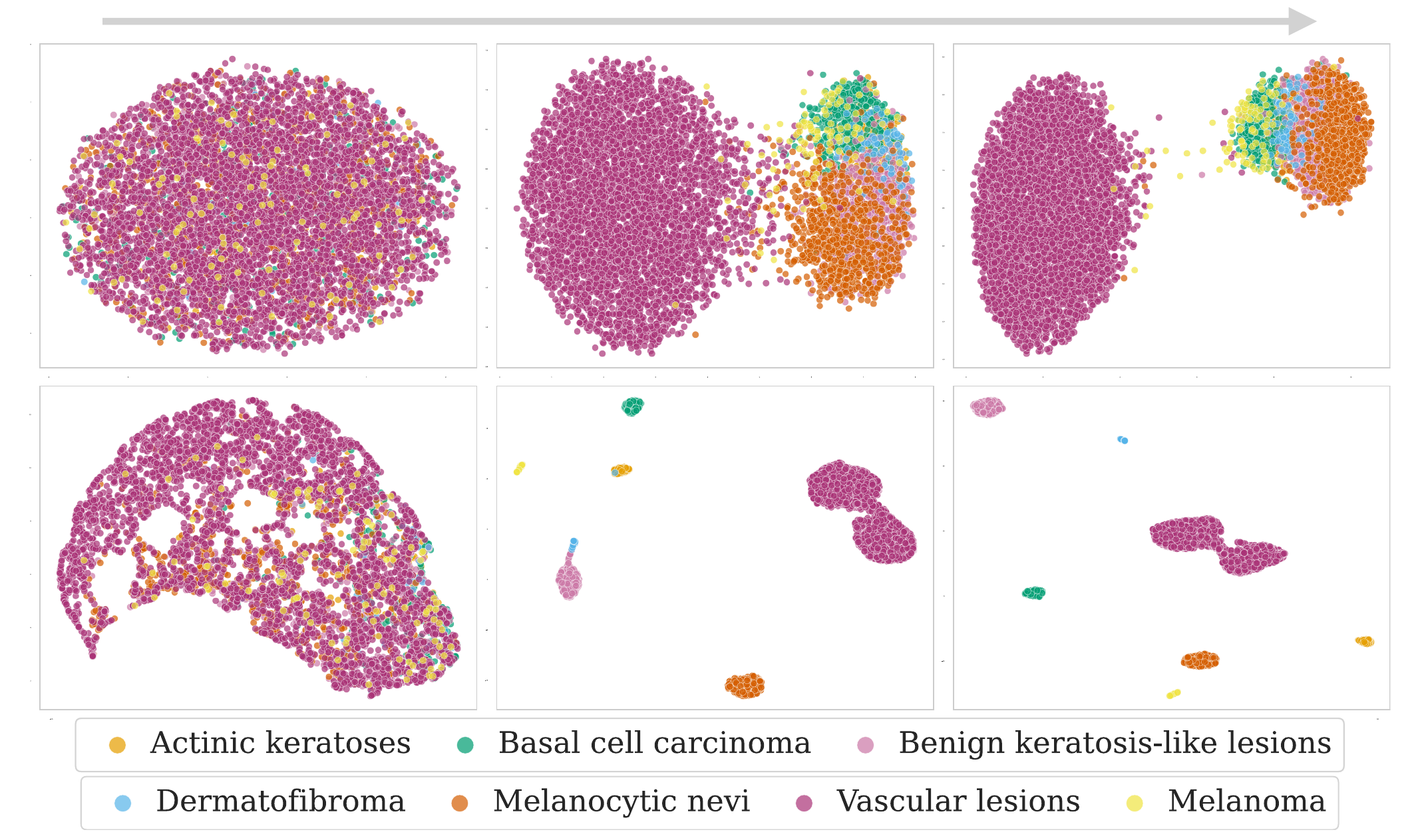}
            \put(-1,36){\scalebox{0.8}{\rotatebox{90}{\texttt{Exemplar}}}}
            \put(-1,18){\scalebox{0.8}{\rotatebox{90}{\texttt{[CLS]}}}}
            \put(37,59){\scalebox{0.75}{Training Time}}
        \end{overpic}
        \vspace*{-0.5cm}
        \caption{DermaMNIST}
        \label{fig:umap-learned-cls-dermamnist}
    \end{subfigure}
    \hfill
    \begin{subfigure}[t]{0.49\linewidth}
        \centering
        \raisebox{1.75mm}{
        \begin{overpic}[width=\linewidth]{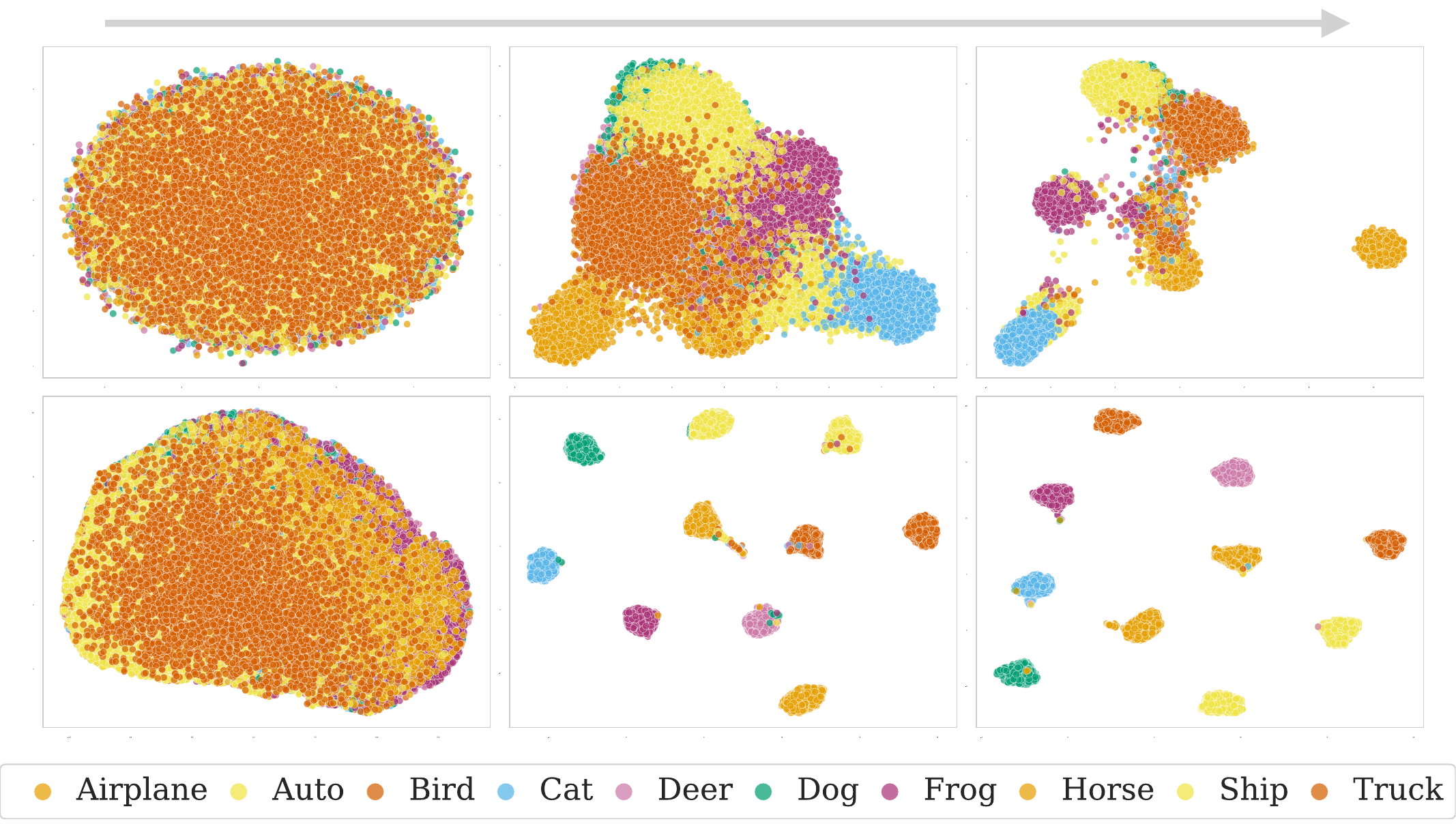}
            \put(-1,33){\scalebox{0.8}{\rotatebox{90}{\texttt{Exemplar}}}}
            \put(-1,14){\scalebox{0.8}{\rotatebox{90}{\texttt{[CLS]}}}}
            \put(37,56.5){\scalebox{0.75}{Training Time}}
        \end{overpic}}
        \vspace*{-0.5cm}
        \caption{CIFAR-10}
        \label{fig:umap-learned-cls-cifar10}
    \end{subfigure}
    \caption{UMAP visualization of learned exemplar tokens in the memory and $[CLS]$ tokens at epochs 0, 50, and 100 of training. Left: DermaMNIST. Right: CIFAR-10.}
    \label{fig:umap-learned-cls-both}
    \vspace*{-0.25cm}
\end{figure}
\paragraph{Information Encoded in the Memory.}

We investigate whether the learned exemplar tokens capture instance-specific information beyond generic class labels by analyzing the representation space of the memory. Figure~\ref{fig:umap-learned-cls-both} contrasts the 2D UMAP~\citep{mcinnes2018umap} projections of the Exemplar Memory Bank $\{\vv_i\}_{i=1}^N$ with those of the $[CLS]$ tokens. The $[CLS]$ tokens are extracted from the model backbone's output for each training point and serve as features for classification, therefore capturing a class-guided baseline representation. We observe two distinct patterns consistent with our quantitative results: In CIFAR-10, where the key encoder's pre-trained features are highly relevant to the class, the memory bank exhibits significant class-based clustering, albeit substantially weaker than the $[CLS]$. Conversely, in DermaMNIST, the memory bank representations are more diffuse, reflecting the distribution shift from the encoder's pre-training data. As shown in our temporal visualizations across three training stages, the $[CLS]$ tokens rapidly form tight class-specific clusters to minimize the objective loss. In contrast, the exemplar tokens evolve more granularly; they do not merely replicate class separation but instead organize into a richer topology mirroring instance-specific features.\looseness-1 

\begin{figure}[h!]
    \centering
    \vspace*{0.15cm}
\hspace*{1.5mm}
    \begin{minipage}{0.48\linewidth}
        \centering
        \begin{overpic}[width=\linewidth]{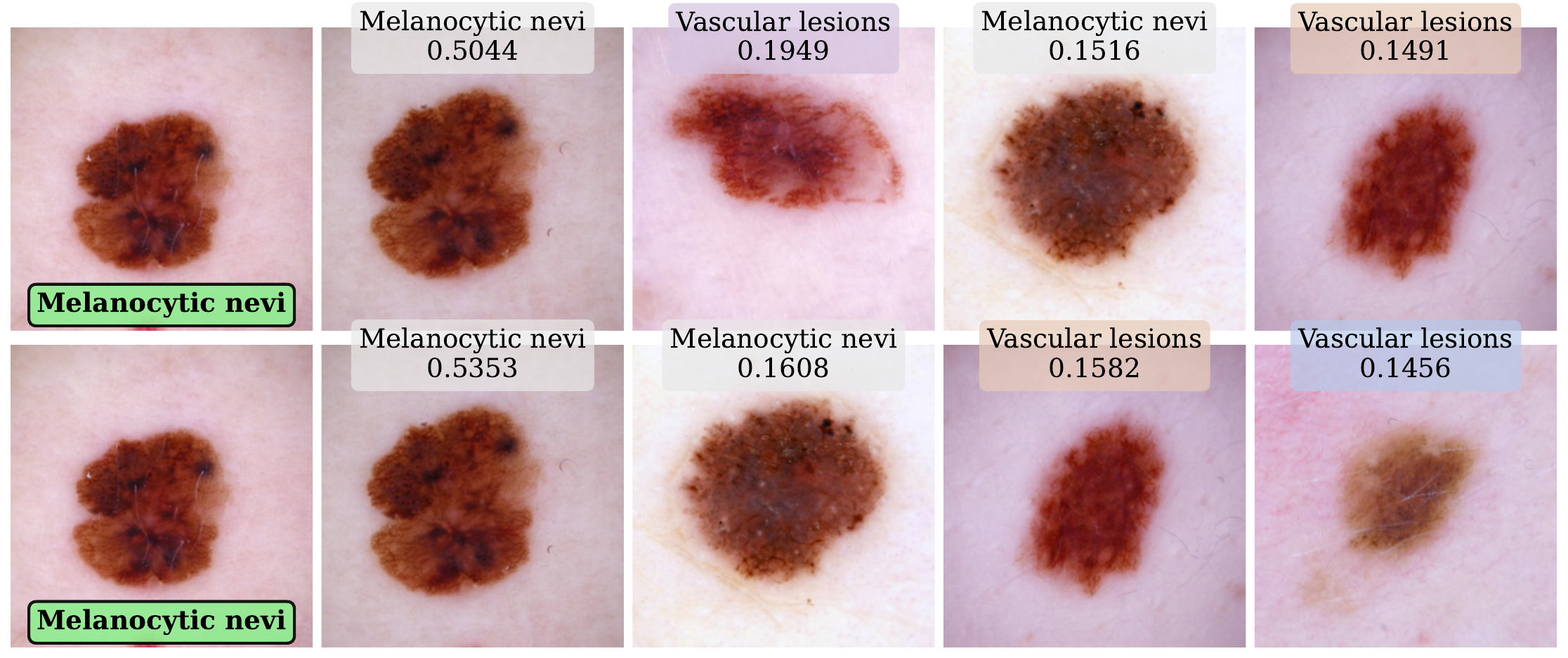}
            \put(-2.5,9){\scalebox{0.7}{\rotatebox{90}{\texttt{Test Sample}}}}
            \put(-6.2,23.5){\scalebox{0.7}{\rotatebox{90}{\texttt{Original}}}}
            \put(-6.2,1.4){\scalebox{0.7}{\rotatebox{90}{\texttt{Unlearning}}}}
            \put(9,43.5){\scalebox{0.7}{$\vx_i$}}
            \put(54.5,43.5){\scalebox{0.7}{$\mathcal{N}_K(\vx_i)$}}
        \end{overpic}
    \end{minipage}
   \hspace*{1mm}
    \begin{minipage}{0.48\linewidth}
        \centering
        \begin{overpic}[width=\linewidth]{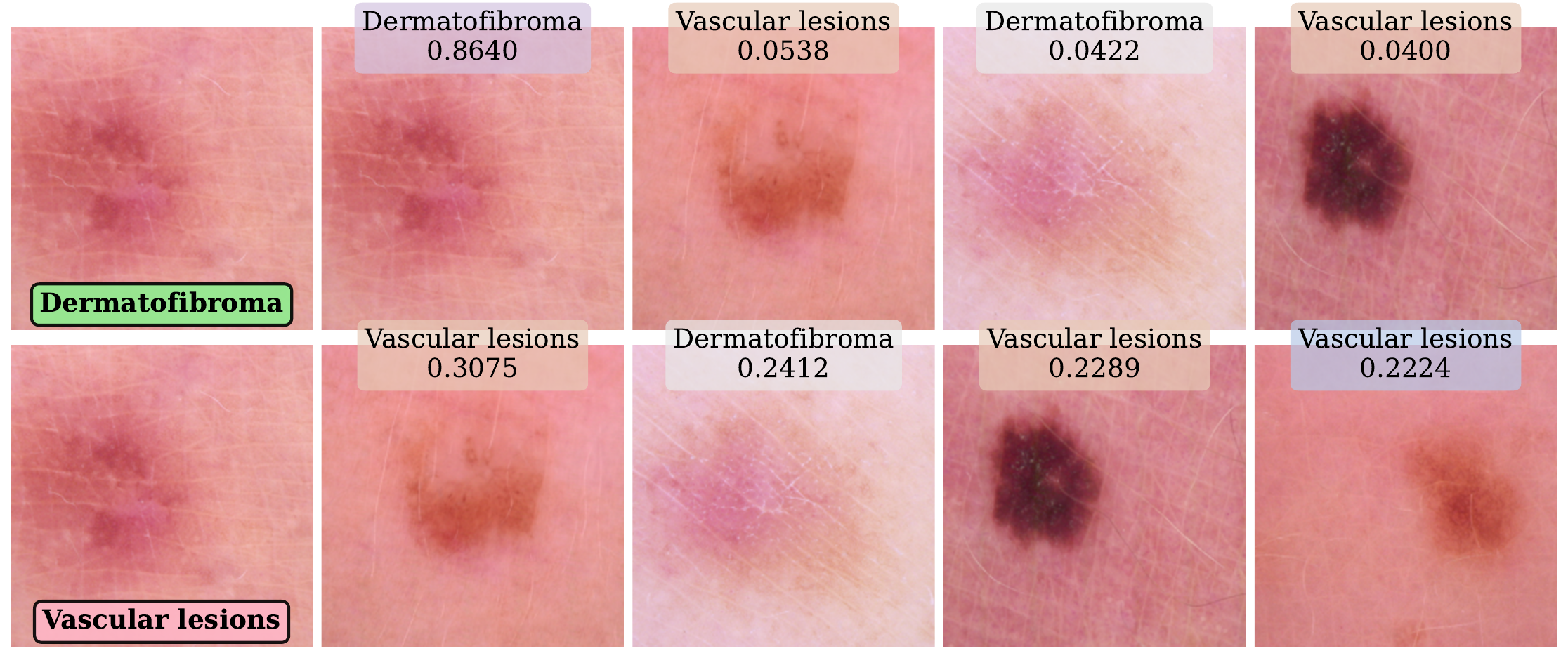}
            \put(-2.5,7.5){\scalebox{0.7}{\rotatebox{90}{\texttt{Forget Sample}}}}
            \put(9,43.5){\scalebox{0.7}{$\vx_i$}}
            \put(54.5,43.5){\scalebox{0.7}{$\mathcal{N}_K(\vx_i)$}}
        \end{overpic}
    \end{minipage}
    \caption{Retrieval visualizations on DermaMNIST on a test (left) and a forget (right) sample. Rows show the original and unlearned behavior, and their retrieved counterparts. Labels: \rbox{NewNeighbor}{new neighbors}, \rbox{CorrectMatch}{correct matches}, \rbox{WrongClass}{wrong-class neighbors}, \rbox{Forgotten}{forgotten samples}, \rbox{PredCorrect}{correct} and \rbox{PredWrong}{incorrect} predictions.\looseness-1}
    \label{fig:knn-dermamnist}
\end{figure}

To further illustrate the mechanics of unlearning within the memory bank, Figure~\ref{fig:knn-dermamnist} presents retrieval examples from the DermaMNIST test and forget sets, with $K=4$ neighbors. By identifying a targeted instance and removing its key-value pair from the memory bank, we observe distinct behaviors: for a test sample $\vx_i$, e.g., left rows, the key encoder continues to retrieve visually similar images $\mathcal{N}_K(\vx_i)$, even if belonging to different classes, and MUNKEY retrieves all the updated $\vv_j$ from its memory without changing the final predicted class. In contrast, for a training sample $\vx_i$ marked for forgetting, e.g., right rows, the retrieved neighbor with the strongest weight prior to unlearning is the instance itself, measured through the softmax of cosine similarities. However, once unlearning is performed, a new instance is retrieved. This shift in neighbors, now without the instance-specific information, effectively alters the content and reweighting of all the retrieved $\vv_j$ and changes the final class prediction. We further quantify the effect of neighboring key deletions on retained samples in Appendix~\ref{app:neighbor-deletion}. \looseness-1

\paragraph{Effect of Aggregation Strategies.} 
\begin{wraptable}{r}{0.5\columnwidth}
\centering
\vspace*{-0.4cm}
\caption{Ablation of aggregation strategies on DermaMNIST and CIFAR-10. $P_s$ denotes Pathway Sensitivity (lower is more balanced), Avg Gap denotes unlearning faithfulness (lower is better).}
\vspace{-0.1cm}
\label{tab:ablation-summary}
\scriptsize
\setlength{\tabcolsep}{1pt}
\begin{tabular}{l cc cc}
\toprule
& \multicolumn{2}{c}{\textbf{DermaMNIST}} & \multicolumn{2}{c}{\textbf{CIFAR-10}} \\
\cmidrule(lr){2-3} \cmidrule(lr){4-5}
\textbf{Strategy} & $P_s$ ($\downarrow$) & Avg Gap ($\downarrow$) & $P_s$ ($\downarrow$) & Avg Gap ($\downarrow$) \\
\midrule
Direct Key      & $0.212_{\pm .012}$ & $8.23_{\pm .70}$ & $0.239_{\pm .007}$ & $3.34_{\pm .18}$ \\
Deep Prompt     & $0.172_{\pm .006}$ & $1.79_{\pm .69}$ & $0.188_{\pm .006}$ & $0.92_{\pm .52}$ \\
Cross Attn.     & $0.208_{\pm .005}$ & $\mathbf{1.02}_{\pm .46}$ & $0.292_{\pm .005}$ & $\mathbf{0.29}_{\pm .13}$ \\
\midrule
\textbf{Ensemble} & $0.169_{\pm .004}$ & \underline{1.52}$_{\pm .56}$ & $0.184_{\pm .006}$ & \underline{0.58}$_{\pm .22}$ \\
Softmax         & $0.168_{\pm .006}$ & $2.01_{\pm .89}$ & $0.186_{\pm .006}$ & $0.71_{\pm .23}$ \\
Rank            & $0.166_{\pm .006}$ & $2.58_{\pm .78}$ & $0.184_{\pm .006}$ & $0.67_{\pm .21}$ \\
\bottomrule
\end{tabular}
\vspace{-0.2cm}
\end{wraptable}
As established in~\Cref{sec:munkey}, MUNKEY takes advantage of the synergistic benefits of image and memory pathways. While our architecture employs a pathway ensemble during inference, we investigate alternative aggregation strategies to determine how instance-specific information is integrated. These variants include Cross Attention, integrated at varying transformer depths; Deep Prompting~\citep{jia2022visual}, where the exemplar token is re-inserted as a prefix for every transformer block; and Direct Key, which bypasses the learnable memory by embedding the key encoder's output. Furthermore, within the standard input-level aggregation framework, we explore late-fusion techniques of the retrieved neighbors, comparing ensembling, rank-based mixing, and softmax-based weighting of exemplar tokens. We view MUNKEY not merely as a fixed model, but as a realization within the proposed paradigm \textit{unlearning by design}, where different architectural choices yield distinct trade-offs and an analysis of these strategies is provided in Appendix~\ref{app:variants}.

We report a subset of results on DermaMNIST and CIFAR-10 in Table~\ref{tab:ablation-summary}, focused on the Pathway Sensitivity Score, $P_s$, and Avg Gap to represent unlearning and pathway reliance. We see that both direct key and cross attention result in elevated $P_s$, indicating pathway imbalance. In the case of cross attention, there are utility improvements on Avg Gap, but those do not outweigh the added model complexity and worse $P_s$. Similarly, the limitations of the direct key approach empirically motivate the need for MUNKEY's instance-specific exemplar memory. Conversely, late-fusion methods and deep prompting maintain consistently lower $P_s$, signaling a more balanced integration of features. Among these, ensembling is the most robust, with the smallest Avg Gap and a stable $P_s$ across both natural and medical domains. Although we observe fluctuations across datasets, the results suggest that pathway aggregation serves as a critical structural lever for unlearning, while the choice of late-fusion strategy is less impactful. Crucially, in scenarios where the computational overhead of $K$ forward passes makes ensembling challenging, aggregating neighbors at the token level and doing a single inference pass, as in softmax and rank, offers a cost-effective alternative with competitive performance.\looseness-1

\section{Conclusion}
\label{sec:conclusion}
In this work, we advocate for a shift toward the \textit{unlearning by design} paradigm in approximate methods. Current research focuses largely on post-hoc methods to adapt models that were never intended to forget. As data privacy regulations and security concerns intensify, there is a growing need for architectures that are inherently modular by construction. Within this paradigm, we propose MUNKEY, a framework that externalizes instance-specific knowledge into an explicitly editable memory bank. We reformulate the training and interaction of memory-augmented transformers so that the burden of instance-specific memorization is shifted from static weights to a learnable memory, enabling near-instantaneous, zero-shot forgetting. Results across diverse domains, including natural imagery and medical datasets, highlight the robustness of this approach. More broadly, our findings open new doors to a rich design space of modular architectures centered on adaptability for settings where stored information must be selectively adapted, updated, or removed over time. Beyond unlearning, this structural separation enables a form of interpretability, critical to high-stakes domains: by identifying retrieved ``prototypes'', i.e., influential neighbors for a decision, the model provides explicit evidence for its predictions, supporting transparent, case-based reasoning often absent in standard black-box systems. This design also entails practical trade-offs, including dataset-scaling memory overhead, typically modest relative to model size, and reliance on a key encoder, which trades additional storage for instantaneous and targeted unlearning. As with other approximate unlearning methods, our results provide empirical evidence under black-box membership inference rather than formal guarantees of forgetting. Ultimately, this work motivates exploration into approximate models that treat the ability to forget not as an afterthought, but as a fundamental architectural requirement, through decoupled memories or other instantiations of the \textit{unlearning by design} paradigm.\looseness-1

\begin{ack}

MV and SL are supported by the Swiss State Secretariat for Education, Research, and Innovation (SERI) under contract number MB22.00047. JSG is supported by the grant \#2021-911 of the Strategic Focal Area “Personalized Health and Related Technologies (PHRT)” of the ETH Domain (Swiss Federal Institutes of Technology). AR is supported by the StimuLoop grant \#1-007811-002 and the Vontobel Foundation. Computational data analysis was performed at Leonhard Med (\url{https://sis.id.ethz.ch/services/sensitiveresearchdata/}) secure trusted research environment at ETH Zurich.\looseness-1
 \end{ack}

\newpage
\bibliographystyle{apacite}
\bibliography{bibliography}

\newpage \appendix
\section{Extended Related Work}
\label{app:related_work_extended}

\paragraph{Machine Unlearning.} 

In this section, we describe additional work tangential to MUNKEY but relevant to the broader picture of machine unlearning. Beyond approximate methods discussed in Section~\ref{sec:related_work}, exact unlearning has also been studied, typically with formal theoretical guarantees~\citep{ginart2019making,guo2020certified, wuunlearned}. Retraining from scratch is sometimes considered the only truly exact unlearning method, as it fully guarantees that information from the forget set is not retained in the model~\citep{nguyen2025survey}; however, this approach is often impractical. To mitigate this, prior work proposes data sharding and slicing strategies, where models are trained on disjoint partitions and ensembled to enable exact unlearning at the shard level~\citep{cao2015towards,bourtoule2021machine,guo2020certified}. Such approaches have been applied in domains including graph neural networks~\citep{chen2022graph} and recommendation systems~\citep{chen2022recommendation}. Nevertheless, exact unlearning methods often incur substantial computational costs due to retraining on large retain sets, or high storage costs from maintaining multiple models trained on disjoint data subsets. Further lines of research address unlearning from alternative optimization perspectives, aiming to better balance forgetting and retention~\citep{leeruago}. This includes formulations based on game theory, i.e., where unlearning is cast as a Stackelberg game between the unlearner and an auditor~\citep{diadversarial}. Other works~\citep{kimunlearning} adopt meta-learning perspectives to jointly optimize forgetting and remembering~\citep{huang2024learning}, or explicitly mitigate gradient conflicts between unlearning and retention objectives through implicit regularization mechanisms~\citep{patel2025learning}. Incremental unlearning has also been explored in pre-trained vision models via feature subspace decomposition and dynamic subspace adaptation~\citep{feng2025fg,kawamuraapproximate, wang2025machine}. Finally, a growing body of work studies unsupervised unlearning using variational inference~\citep{shen2024label}; while such methods can achieve performance comparable to supervised approaches, they are generally not superior, and we therefore do not include them in our comparisons.\looseness-1

Unlearning approaches can be distinguished by the level of information they aim to remove. While many of the aforementioned methods focus on instance- or class-level unlearning, i.e.~\citep{zhou2025decoupled}, a growing body of work targets concept unlearning, particularly in generative settings. For example, recent work studies concept removal and relearning prevention in diffusion models~\citep{gandikota2023erasing,gao2025meta,biswas2025cure}, and in large foundation models such as CLIP, Stable Diffusion, or VLMs, by identifying and modifying relevant layers with efficient gradient-based localization~\citep{caitargeted,zhang2025targeted,dong2025machine}. Similarly, a large body of work targets large language models, including approaches based on distillation~\citep{lee2025distillation}, mechanistic localization~\citep{guomechanistic}, or alternative supervision strategies~\citep {mekala2025alternate,wangllm}. From another angle, SAeUron~\citep{cywinskisaeuron} adopts sparse autoencoders from mechanistic interpretability to remove specific concepts in diffusion models. While highly relevant for generative modeling, these approaches address a different unlearning setting from our discriminative vision task and are therefore beyond the scope of this work.

\paragraph{Associative and External Memories.}
To give a broader context to the exemplar memory we propose in MUNKEY, we discuss here associative and external memory mechanisms that have been widely explored to augment neural networks with explicit storage and retrieval capabilities. Associative (internal) memory approaches embed learnable memory structures directly into network components, including product key memory layers~\citep{lample2019large} and fine-tuning methods that introduce learnable memory tokens to provide persistent, task-specific context in Vision Transformer encoders~\citep{sandler2022fine}. Similarly, Visual Prompt Tuning~\citep{jia2022visual} introduces learnable tokens that are embedded into the model to adapt its behavior. In contrast, as described in~\Cref{sec:related_work}, external memory architectures decouple memory from the model parameters~\citep{graves2014neural, DBLP:journals/corr/WestonCB14,sukhbaatar2015end,santoro2016meta,wu2022memorizing}. 
More recent work on retrieval-augmented models, such as Retrieval-Augmented Classification (RAC)~\citep{long2022retrieval}, combines a parametric image classifier with an external retrieval memory. However, since training samples influence both the classifier weights and the retrieval memory, deleting an entry does not remove its influence from the learned parameters. In contrast, Tip-Adapter~\citep{zhang2022tip} keeps CLIP frozen and stores downstream examples only in an external cache, allowing their contribution to be removed through deletion. However, this is a frozen-model adaptation setting rather than unlearning, as these examples never contributed to parameter training. MUNKEY instead explicitly trains the model for subsequent deletion, encouraging instance-specific information to be localized in removable memory rather than shared model parameters.
 
\newpage
\section{Detailed Experimental Setting}
\label{app:exp-setup}

\subsection{MUNKEY Algorithm}
\label{app:algorithm}

Algorithm~\ref{alg:munkey} summarizes the MUNKEY pipeline, from training with exemplar memory to unlearning by key deletion and inference via nearest-neighbor retrieval, as introduced in~\Cref{sec:munkey}.

    \begin{algorithm}[h!]
    \small
    \caption{\small{MUNKEY: Machine UNlearning via KEY Deletion}}
    \label{alg:munkey}
    \begin{algorithmic}[1]
    \STATE \textbf{Input:} Training set $\mathcal{D}$, frozen key encoder $g_\phi$, $p_i, p_t, p_r$
    \STATE \textbf{Initialize:} Model $\theta$, adapter $\psi$, and memory bank $\mathcal{M} = \{(\vk_i, \vv_i)\}_{i=1}^N$
    \STATE \textbf{Phase 1: Training}
    \STATE Precompute $\vk_i = g_\phi(\vx_i)$ for all $\vx_i \in \mathcal{D}$
    \WHILE{not converged}
        \FOR{each $(\vx_i, \vy_i) \in \mathcal{D}$}
            \STATE $\vz_i \gets \text{ViTEmbed}(\vx_i)$
            \STATE Sample $\mathbf{m} = [\gamma_{img}, \gamma_{tok}]$ $\sim$ Cat$(p_i, p_t, 1 - p_i - p_t,0)$
            \STATE Sample $r \sim \text{Bern}(p_r)$
            \IF{$r = 1$}
                \STATE Sample $K' \sim \mathcal{U}\{2, 16\}$
                \STATE $\vv_n \gets \text{Agg}(\{\vv_j \mid j \in \mathcal{N}_{K'}(\vx_i; \mathcal{M}\setminus\{(\vk_i,\vv_i)\})\})$

                \STATE $\text{Exemplar} \gets h_\psi(\vv_n)$
            \ELSE
                \STATE $\text{Exemplar} \gets h_\psi(\vv_i)$
            \ENDIF
            \STATE $\vz_{in} \gets [\mathcal{P}(\vz_i, \emptyset_{img}, \gamma_{img}), \mathcal{P}(\text{Exemplar}, \emptyset_{tok}, \gamma_{tok})]$
            \STATE $\mathcal{L} \gets \ell(f_\theta(\vz_{in}), \vy_i)$
            \IF{$r = 1$}
                \STATE Update $\theta,\psi$  by minimizing $\mathcal{L}$; set $\nabla_{\vv_j}\mathcal{L}=0$ for retrieved neighbors $j$
            \ELSE
                \STATE Update $\theta, \psi, \vv_i$ by minimizing $\mathcal{L}$
            \ENDIF
        \ENDFOR
    \ENDWHILE
    
    \STATE \textbf{Phase 2: Zero-Shot Unlearning}
    \STATE \textbf{Input:} Forget set $\mathcal{D}_f$
    \STATE $\mathcal{M}_u \gets \mathcal{M} \setminus \{(\vk_i, \vv_i) \mid (\vx_i, \vy_i)  \in \mathcal{D}_f\}$ \COMMENT{Instance key deletion}
    
    \STATE \textbf{Phase 3: Inference}
    \STATE \textbf{Input:} Query $\vx_q$, Updated Memory $\mathcal{M}_u$
    \STATE Retrieve $\mathcal{N}_K(\vx_q)$ from $\mathcal{M}_u$ based on $\text{sim}(g_\phi(\vx_q), \vk_j)$
    \STATE  $\vz_q \gets \text{ViTEmbed}(\vx_q)$
    \STATE $\hat{\vy} \gets \sum_{j \in \mathcal{N}_K(\vx_q)} w_j \cdot f_\theta([\vz_q, h_\psi(\vv_j)])$
    \STATE \textbf{return} $\hat{\vy}$
    \end{algorithmic}
    \end{algorithm}

\subsection{Datasets}
\label{app:datasets}

To evaluate the performance and robustness of our proposed method, we conduct experiments across a diverse range of benchmarks, spanning both natural and medical imaging domains. All datasets used in this work are publicly available and were obtained from their official sources. We use them exclusively for research and benchmarking purposes, and follow the corresponding dataset terms of use.\looseness-1

\paragraph{Natural Image Benchmarks.} We first evaluate our model on the CIFAR-10 and CIFAR-100 datasets~\citep{krizhevsky2009learning}. Both are established computer vision benchmarks consisting of 60,000 $32 \times 32$ color images, 50,000 in the training set and 10,000 in the validation set. CIFAR-10 contains 10 mutually exclusive classes, whereas CIFAR-100 provides a more complex challenge with 100 fine-grained categories. Additionally, we utilize Tiny ImageNet, a subset\footnote{Tiny ImageNet Visual Recognition Challenge: \url{http://cs231n.stanford.edu/tiny-imagenet-200.zip}} of the larger ImageNet database~\citep{deng2009imagenet}. This dataset comprises 200 object classes with a total of 100,000 training, 10,000 validation, and 10,000 test unlabeled images. For all three natural image benchmarks, we perform a random split of the original validation data to create a custom validation set (40\%) and a test set (60\%). All images in these datasets are resized to $224 \times 224$ resolution.  

\paragraph{Biomedical Image Benchmarks.} To assess the method's efficacy in specialized domains, we incorporate high-resolution ($224 \times 224$) tasks from the MedMNIST v2 collection~\citep{yang2023medmnist}. We include three datasets of varying clinical nature: First, BloodMNIST~\citep{acevedo2020dataset} is a collection of 17,092 microscopic images of normal blood cells across 8 classes with moderate class imbalance. It serves as a benchmark for morphological recognition, requiring the model to distinguish subtle structural features like nuclear shape and cytoplasmic granules against controlled backgrounds. We utilize the predefined splits: 11,959 training, 1,712 validation, and 3,421 test samples. Second, PathMNIST~\citep{kather2019predicting, kather2018100} comprises 107,180 histology patches for 9-class colorectal cancer tissue classification. This dataset is chosen to represent a large-scale medical imaging task. We follow the standard training split (89,996 samples); however, as the original test set (7,180 samples) is collected from different clinical centers to test out-of-distribution generalization, which is outside the scope of this unlearning study, we utilize the validation set (10,004 samples) for testing to maintain focus on the core methodology. Lastly, DermaMNIST~\citep{tschandl2018ham10000,codella2019skin} contains 10,015 dermatoscopic images, 7007 for training, 1003 for validation, and 2005 for testing, used for 7-class skin disease classification. It mimics real-world clinical prevalence, where common conditions like \textit{melanocytic nevi} significantly outnumber rarer, critical pathologies such as \textit{melanoma}. Furthermore, its multi-source nature introduces hardware and photographic noise, making it a challenging benchmark for imbalanced, real-world diagnosis. Given its representative difficulty, DermaMNIST is included in our main manuscript alongside CIFAR-10 as primary benchmarks, while the results for the remaining datasets are provided in Appendices~\ref{app:extra-results}, \ref{app:variants}.

\subsection{Baselines Setup}
\label{app:baselines}

To provide a comprehensive evaluation of MUNKEY, we compare against a diverse suite of baselines spanning traditional optimization, weight sparsification, distillation-based unlearning, parameter dampening, and recent adversarial unlearning. We first note that MUNKEY is Transformer-based; however, most post-hoc unlearning methods are primarily reported on CNN or ResNet backbones, despite being applicable to Transformers. Since ViTs are the current gold standard backbone for vision, we implement \emph{all} baselines using a ViT-Tiny architecture, as in MUNKEY, to ensure a fair comparison and to bridge this reporting gap, bringing results of the state-of-the-art into transformer-based unlearning. For all optimization-based baselines adapted to ViT-Tiny, we replace the commonly used SGD optimizer with AdamW (betas $[0.9,0.99]$, weight decay $0.05$). Additionally, we perform a learning-rate grid search using multiplicative factors in $[0.1, 1.0]$ relative to the original training rate of the base model $1.5\times 10^{-4}$. In this section, we list the baselines considered and their implementation details, and report their training and unlearning runtimes in Table~\ref{tab:times-stdev}. While gradient-based methods like NegGrad+ and SCRUB require significant compute for parameter updates, and methods like SSD require time to calculate the Fisher Information Matrix, MUNKEY’s unlearning time is negligible ($\approx 0$s), hence O(1) computations.\looseness-1
\paragraph{Retrained oracle.}
Retrain serves as the fundamental oracle baseline. For all baselines, this corresponds to training a ViT-Tiny from scratch on the retain data $\mathcal{D}_r$. For our method, we additionally report an oracle that matches the MUNKEY ViT-Tiny architecture with its hyperparameters and additional components, also trained uniquely on $\mathcal{D}_r$, providing an upper bound aligned with our design.\looseness-1

\paragraph{Traditional optimization-based baselines.} These include earlier machine unlearning approaches. Catastrophic Forgetting (CF) fine-tunes the model on $\mathcal{D}_r$. CF-k~\citep{goel2022towards} uses the same fine-tuning procedure but freezes the backbone and updates only the classifier weights. Moreover, Exact Unlearning-k (EU-k)~\citep{goel2022towards} resets the classifier parameters before fine-tuning on $\mathcal{D}_r$. NegGrad+~\citep{kurmanji2023towards} is an adaptation of the more primitive NegGrad that performs gradient ascent on the forget set $\mathcal{D}_f$ while simultaneously performing gradient descent on $\mathcal{D}_r$; a grid search over the forget/utility trade-off found $0.95/0.05$ to work best as the weight ratio of both loss terms.\looseness-1

\paragraph{Weight sparsification.} $\ell_1$-sparse~\citep{jia2023model} applies an $\ell_1$-penalty to the loss, with scaling parameter $\lambda=0.01$ to the knowledge weights in the backbone to sparsify memorized information, while excluding biases and embeddings for stability. 

\paragraph{Noise-based baseline.} Amnesiac unlearning, implemented as in~\citet{foster2024fast}, applies random labels as noise supervision on the forget set $\mathcal{D}_f$ during fine-tuning with the rest of the data. 

Following recent literature~\citep{ebrahimpour2025amun}, and our own empirical findings, we run all the above methods for 10 epochs to achieve the best trade-off between forgetting and utility. \looseness-1
 
 \paragraph{Teacher-Student distillation baseline.} SCRUB~\citep{kurmanji2023towards} is a common representative of the distillation-based methods, where a student model is trained to selectively inherit knowledge from the original model (the teacher), specifically by minimizing KL-divergence to stay close to the teacher on $\mathcal{D}_r$ while maximizing that divergence to move away from it on $\mathcal{D}_f$. We follow the original codebase and optimization steps, unlearning for 10 epochs, and using their \textit{rewinding} procedure to select the checkpoint with the smallest validation utility gap.

 \paragraph{Optimization-free parameter dampening.} This category of methods includes SSD (Selective Synaptic Dampening)~\citep{foster2024fast}. Unlike fine-tuning approaches, and similar to MUNKEY, these methods do not require iterative optimization at unlearning time; instead, they identify and dampen the weights most responsible for the forget set. In particular, SSD relies on computing the Fisher Information Matrix (FIM) to identify relevant parameters. While this allows for near-instant unlearning, the FIM calculation introduces a computational overhead prior to the unlearning phase.
 
\paragraph{Adversarial-based unlearning.}  AMUN (Adversarial Machine Unlearning)~\citep{ebrahimpour2025amun} represents a new line of recent work. It generates an adversarial set of images for the forget samples and then fine-tunes the model on $\mathcal{D}_r$ together with these adversarial samples to induce forgetting. We remained faithful to their adversarial set generation pipeline by using an untargeted PGD-50 attack under an $\ell_2$-constraint, starting with step size $0.1\varepsilon$ and increasing $\varepsilon$ until misclassification.

\paragraph{Contrastive-based Unlearning.} CoUn (Contrastive Machine Unlearning)~\citep{khalilcoun25} frames unlearning as representation-space adjustment through contrastive learning on the retain data, combining a SimCLR-style InfoNCE loss with supervised cross-entropy. We follow the official protocol and reported hyperparameters of the original work, unlearning for 50 epochs with the data augmentation pipeline used for the contrastive branch, i.e., stochastic image transformations such as random cropping, horizontal flipping, and normalization.

\paragraph{Memory-based baseline.}
Finally, KNN~\citep{fix1985discriminatory} provides a simple retrieval baseline motivated by the memory component of MUNKEY. We use the same key-encoded embeddings as the search space and predict class probabilities by counting labels among the nearest neighbors. This provides a point of comparison for the simplest form of memory-based inference.

\begin{table}[h]
    \vspace{-0.8\baselineskip}
    \centering
    \caption{Run-Time Efficiency comparison (seconds) on DermaMNIST and CIFAR-10. \textit{Unlearn} is the time to remove data post-training. \textit{Train/Ep.} is the per-epoch training cost of base models.}
    \vspace{0.2cm}
\label{tab:times-stdev}
    \scriptsize
    \setlength{\tabcolsep}{2pt}
    \begin{tabular}{l cc cc}
    \toprule
    & \multicolumn{2}{c}{\tiny \textit{DermaMNIST}} & \multicolumn{2}{c}{\tiny \textit{CIFAR-10}} \\
    \cmidrule(lr){2-3} \cmidrule(lr){4-5}
    \textbf{Model} & \textbf{Unlearn} & \textbf{Train/Ep.} & \textbf{Unlearn} & \textbf{Train/Ep.} \\
    \midrule
    KNN & $\approx 0$ & $\approx 0$ & $\approx 0$ & $\approx 0$ \\
    CF & $114.1_{\pm 1.9}$ & $24.7_{\pm 0.2}$ & $669.5_{\pm 1.6}$ & $174.6_{\pm 1.6}$ \\
    CF-k & $58.5_{\pm 2.8}$ & $24.7_{\pm 0.2}$ & $311.4_{\pm 1.6}$ & $174.6_{\pm 1.6}$ \\
    EU-k & $58.2_{\pm 0.9}$ & $24.7_{\pm 0.2}$ & $311.1_{\pm 1.1}$ & $174.6_{\pm 1.6}$ \\
    $\ell_1$-sparse & $114.3_{\pm 0.8}$ & $24.7_{\pm 0.2}$ & $694.2_{\pm 0.9}$ & $174.6_{\pm 1.6}$ \\
    NegGrad+ & $124.3_{\pm 1.8}$ & $24.7_{\pm 0.2}$ & $761.7_{\pm 1.1}$ & $174.6_{\pm 1.6}$ \\
    Amnesiac & $123.8_{\pm 1.3}$ & $24.7_{\pm 0.2}$ & $760.5_{\pm 1.4}$ & $174.6_{\pm 1.6}$ \\
    SCRUB & $154.4_{\pm 0.3}$ & $24.7_{\pm 0.2}$ & $1478.2_{\pm 27.3}$ & $174.6_{\pm 1.6}$ \\
    SSD & $23.1_{\pm 4.0}$ & $24.7_{\pm 0.2}$ & $105.8_{\pm 13.6}$ & $174.6_{\pm 1.6}$ \\
    AMUN & $313.4_{\pm 6.8}$ & $24.7_{\pm 0.2}$ & $1940.5_{\pm 3.7}$ & $174.6_{\pm 1.6}$ \\
    CoUn & $191.75_{\pm 5.3}$ & $24.7_{\pm 0.2}$ & $1255.88_{\pm 11.4}$ & $174.6_{\pm 1.6}$  \\
    \textbf{MUNKEY} & $\mathbf{\approx 0}$ & $25.9_{\pm 0.3}$ & $\mathbf{\approx 0}$ & $178.5_{\pm 2.7}$ \\
    \bottomrule
    \end{tabular}
    \vspace{-1.0\baselineskip}
\end{table}

\subsection{Implementation Details}
\label{app:imp_details}

The training pipeline and dynamics are standardized across all base models and datasets. We employ a ViT-Tiny (ViT-T/16) backbone with a patch size of 16 and a hidden dimension of 192 for both MUNKEY and the baseline architectures, and we fit a linear classifier on the final $[CLS]$ token. At training time, natural images are augmented using random resize cropping to $224 \times 224$ pixels, while biomedical images are directly resized to $224 \times 224$ pixels. All models are trained for 100 epochs using the AdamW optimizer with $\beta_1=0.9, \beta_2=0.99$, a weight decay of 0.05, and a learning rate that follows a cosine annealing schedule starting at $1.5 \times 10^{-4}$ until the end of training. Regarding MUNKEY-specific hyperparameters, the regularizing $p_r$ is 0.2, the memory bank utilizes a token dimension of 128, and we include a linear layer adapter to map it to the 192 ViT-T/16 latent space. We retrieve $K = 4$ nearest neighbors during inference in the experiments in the main text, and show results on the effect of different numbers of $K$ in Appendix~\ref{app:neighbors}.

The temperature parameter $\tau$ in the softmax operation for neighbor retrieval is set to 0.07 following standard practice in similarity-based representation learning and retrieval architectures \citep{wu2018unsupervised, he2020momentum}. Moreover, the key encoder used for retrieval is a ViT-B/16 pre-trained on ImageNet (sourced from HuggingFace\footnote{https://huggingface.co/google/vit-base-patch16-224}), and again we provide further ablations regarding the choice of key encoder in Appendix~\ref{app:key-encoder}. The dropout probabilities for each dataset, selected based on Appendix~\ref{app:sensitivity}, are $p_i = 0.1, p_t = 0.3$ for CIFAR-10/100; $p_i = 0.1, p_t = 0.1$ for PathMNIST and Tiny ImageNet; and $p_i = 0.3, p_t = 0.1$ for DermaMNIST and BloodMNIST. All experiments were performed on an NVIDIA GeForce RTX 3080 GPU.\looseness-1

\subsection{System Costs}
\label{app:system-costs}

A key motivation of the unlearning by design paradigm is that the cost of forgetting is shifted away from iterative optimization and into lightweight memory operations. In this section, we discuss the system characteristics of MUNKEY and the practical implications of this design.

\paragraph{Memory Footprint.}
The storage requirements of MUNKEY scale linearly with dataset size, as the external memory stores only key-value pairs. Concretely, the memory complexity is $O(N(d_k + d_e))$, where $N$ is the number of stored samples, $d_k$ is the key dimension, and $d_e$ is the exemplar token dimension. We study the effect of the token dimensionality in Appendix~\ref{app:exemplarmemory} and find that strong performance is maintained even with compact tokens of 64 dimensions. 
As an example, for CIFAR-10 ($N=50{,}000$), using a key dimension of 192 and an exemplar token dimension of 64, the total uncompressed float16 memory overhead is 24.41 MiB, consisting of 18.31 MiB for keys and 6.10 MiB for exemplar tokens, with 3.20M additional learnable exemplar parameters. For DermaMNIST ($N=7{,}007$), the total overhead corresponds to 3.42 MiB and 0.45M learnable exemplar parameters.
For larger deployments, this footprint can be further reduced through standard compression and approximate retrieval techniques such as Product Quantization, as shown in~\citep{geirhostowards}. 
Beyond compression, GPU memory requirements can be further reduced by storing the memory bank on CPU or SSD, transferring only the retrieved exemplar tokens to the accelerator. This reduces the memory footprint associated with retrieval to $O(Kd_e)$ per query, independent of dataset size, at the cost of additional retrieval latency.

\paragraph{Training Cost.}
Although the number of learnable exemplar parameters grows linearly with dataset size, this does not necessarily translate into increased per-step training cost. To evaluate this, we measure training throughput with memory banks ranging from 1K to 100K entries on Tiny ImageNet with $d_e = 64$, keeping the backbone and batch size fixed. As shown in Table~\ref{tab:training-scaling}, MUNKEY's per-step training time remains stable across the evaluated memory sizes. This is because key embeddings are precomputed before training, and each minibatch accesses only its own exemplar tokens and a fixed number of retrieved neighbors. Nevertheless, the total number of steps per epoch still scales with dataset size, as in regular learning, and exemplar parameters and their optimizer states require $O(Nd_e)$ storage.

\begin{table}[h!]
\centering
\caption{Training scalability of MUNKEY across different memory bank sizes. We report the number of learnable exemplar parameters and steady-state training step time on an RTX 3080, keeping the backbone and batch size fixed.}
\label{tab:training-scaling}
\vspace*{0.2cm}
\footnotesize
\setlength{\tabcolsep}{4.5pt}
\begin{tabular}{c c >{\columncolor{lightgray}[3pt][3pt]}c}
\toprule
\textbf{Memory Samples ($N$)} & \textbf{Exemplar Parameters} & \textbf{Step Time (ms)} \\
\midrule
1,000   & 0.06M & $27.32_{\pm \text{\tiny 0.03}}$ \\
5,000   & 0.32M & $27.60_{\pm \text{\tiny 0.02}}$ \\
10,000  & 0.64M & $29.06_{\pm \text{\tiny 0.02}}$ \\
25,000  & 1.60M & $28.22_{\pm \text{\tiny 0.02}}$ \\
50,000  & 3.20M & $28.41_{\pm \text{\tiny 0.04}}$ \\
100,000 & 6.40M & $27.70_{\pm \text{\tiny 0.03}}$ \\
\bottomrule
\end{tabular}
\end{table}

\paragraph{Unlearning and Update Costs.}
The main systems advantage of MUNKEY is that forgetting does not require retraining, gradient updates, or expensive parameter editing. Instead, unlearning is implemented through a lightweight masking operation over the memory bank, where forgotten entries are assigned $-\infty$ in the similarity computation and can no longer be retrieved. This yields a deletion complexity of $O(1)$ per deleted instance, in contrast to optimization-based approaches that scale with dataset size, training epochs, and backpropagation cost. Importantly, multiple deletion requests can be handled simultaneously in a single masking operation, rather than sequentially processing each request. In practice, this makes forgetting operationally simple and substantially faster than post-hoc alternatives.\looseness-1

\paragraph{Inference Cost.}
Compared to a standard black-box classifier, inference in MUNKEY introduces an additional retrieval stage. The overall complexity can be written as
$
O(C_{\text{enc}} + N \cdot d_k + C_{\text{pred}})
$ for the non-ensemble variants,
where $C_{\text{enc}}$ is the cost of computing the query key, $N \cdot d_k$ is the similarity search over the memory bank, and $C_{\text{pred}}$ is the cost of the final forward pass over the backbone. Operationally, inference proceeds in three steps: (i) encode the query, (ii) retrieve relevant memory entries through similarity search, and (iii) process the retrieved tokens for the final output. The main source of inference overhead compared to a standard ViT is the additional forward pass through the frozen key encoder, while nearest-neighbor retrieval is relatively inexpensive, requiring only a matrix multiplication against the memory bank. This overhead is common to retrieval-based architectures and could be reduced by using a smaller or faster key encoder. While the retrieval term scales with $N$, it is highly optimized in practice as a matrix similarity operation and can be reduced to sub-linear time through approximate nearest neighbor indexing, as in~\citep{geirhostowards}.

\paragraph{Scaling Considerations.}
Although our experiments are conducted at benchmark scale, larger deployments could benefit from established retrieval optimizations, such as product quantization~\citep{jegou2010product}, approximate nearest-neighbor indexing with FAISS~\citep{johnson2019billion}, or SSD-based retrieval with DiskANN~\citep{jayaram2019diskann}.~\citet{geirhostowards} demonstrate billion-scale retrieval in related memory-based vision systems. This suggests that the core advantages of MUNKEY persist under scale: memory grows predictably, inference can be accelerated with standard retrieval systems, and unlearning remains a lightweight update rather than a new optimization problem. While these techniques are compatible with MUNKEY's retrieval component, their accuracy–latency–storage trade-offs remain unexplored in our setting, and we leave efficient large-scale retrieval to future work.

\subsection{Discussion Beyond Transformers}

Although MUNKEY is implemented using transformers, the underlying principle of externalizing instance-specific information into editable memory is not inherently transformer-specific. Coupling parametric networks with externally addressable memory predates transformers~\citep{graves2014neural,DBLP:journals/corr/WestonCB14} and has also been explored with conventional neural architectures for visual recognition~\citep{santoro2016meta}.

For instance, a CNN could extract a feature representation $u=f_{\mathrm{CNN}}(x)$ and aggregate retrieved exemplar tokens into $e(x)=\operatorname{Agg}(\{v_j\}_{j\in\mathcal N_K(x)})$. A classification head $h_{\mathrm{cls}}$ could then combine these representations through concatenation $h_{\mathrm{cls}}([u;e(x)])$, projected addition $h_{\mathrm{cls}}(u+We(x))$, where $W$ maps the exemplar representation to the CNN feature space, or learned gating. However, the main challenge lies not in integrating external memory, but in encouraging the separation of instance-specific information from shared model parameters. In MUNKEY, attention enables interactions between image patches and exemplar tokens, while pathway dropout regularizes their respective contributions. Standard CNNs do not provide an equivalent token-level interaction mechanism, and dropping the retrieved representation at the fusion point provides coarser regularization. Investigating suitable fusion mechanisms and dropout strategies to achieve comparable separation in other architectures remains an interesting direction for future work. Moreover, extending MUNKEY to generative language models would additionally require defining the granularity of removable memories and how retrieved information conditions generation, which we likewise leave to future work.

\newpage
\section{Additional Benchmarks}
\label{app:extra-results}

\paragraph{Extended Datasets.}
In this section, we provide detailed results for the remaining benchmarks, CIFAR-100, BloodMNIST, PathMNIST, and Tiny ImageNet, using the suite of metrics defined in~\Cref{sec:experiments}, comprising Test Accuracy, Retain Accuracy, Forget Accuracy, MIA, and Average Gap to evaluate performance, utility, and forgetting. We note that MUNKEY consistently outperforms all baselines across these datasets, with the exception of PathMNIST in~\Cref{tab:unlearning-pathmnist}, where it remains highly competitive and on par with baselines within a standard deviation, as most methods perform highly on this task. BloodMNIST, seen in~\Cref{tab:unlearning-bloodmnist}, mirrors the performance trends observed for DermaMNIST in the main text, where MUNKEY outperforms baselines in both accuracy and MIA. This is particularly interesting as another example of non-natural image data that differs significantly from the distribution used for the key encoder. \looseness-1

\begin{table}[h!]
\centering
\caption{Unlearning performance on a random forget set of 10\% of the training data on BloodMNIST.}
\vspace*{0.1cm}
\label{tab:unlearning-bloodmnist}
\footnotesize
\setlength{\tabcolsep}{5.2pt} 
\begin{tabular}{l cccc >{\columncolor{lightgray}}c}
\toprule
\textbf{Model} & \textbf{TA} & \textbf{RA} & \textbf{FA} & \textbf{MIA} & \textbf{Avg Gap $\downarrow$} \\
\midrule
Retrain \text{\tiny (T)} & $96.88_{\pm \text{\tiny 0.16}}$ & $99.99_{\pm \text{\tiny 0.00}}$ & $96.43_{\pm \text{\tiny 0.42}}$ & $49.60_{\pm \text{\tiny 0.28}}$ & -- \\
Retrain \text{\tiny (M)}  & $97.55_{\pm \text{\tiny 0.15}}$ & $99.49_{\pm \text{\tiny 0.13}}$ & $97.21_{\pm \text{\tiny 0.22}}$ & $49.60_{\pm \text{\tiny 0.41}}$ & -- \\
\midrule
KNN & $93.05_{\pm \text{\tiny 0.05}}$ & $95.23_{\pm \text{\tiny 0.04}}$ & $92.25_{\pm \text{\tiny 0.58}}$ & $49.80_{\pm \text{\tiny 0.61}}$ & -- \\
CF & $96.65_{\pm \text{\tiny 0.21}}$ & $99.74_{\pm \text{\tiny 0.12}}$ & $99.50_{\pm \text{\tiny 0.41}}$ & $54.10_{\pm \text{\tiny 0.12}}$ & $2.01_{\pm \text{\tiny 0.18}}$ \\
CF-k & $96.93_{\pm \text{\tiny 0.25}}$ & $99.96_{\pm \text{\tiny 0.04}}$ & $100.00_{\pm \text{\tiny 0.00}}$ & $52.50_{\pm \text{\tiny 0.47}}$ & $1.64_{\pm \text{\tiny 0.19}}$ \\
EU-k & $96.94_{\pm \text{\tiny 0.26}}$ & $99.94_{\pm \text{\tiny 0.07}}$ & $99.97_{\pm \text{\tiny 0.04}}$ & $52.90_{\pm \text{\tiny 0.42}}$ & $1.74_{\pm \text{\tiny 0.18}}$ \\
$\ell_1$-sparse & $90.92_{\pm \text{\tiny 1.65}}$ & $92.14_{\pm \text{\tiny 1.87}}$ & $91.55_{\pm \text{\tiny 1.48}}$ & $50.99_{\pm \text{\tiny 0.62}}$ & $5.02_{\pm \text{\tiny 0.75}}$ \\
NegGrad+ & $96.48_{\pm \text{\tiny 0.25}}$ & $99.77_{\pm \text{\tiny 0.05}}$ & $97.80_{\pm \text{\tiny 0.04}}$ & $51.45_{\pm \text{\tiny 0.56}}$ & $\underline{0.96}_{\pm \text{\tiny 0.20}}$ \\
Amnesiac & $96.27_{\pm \text{\tiny 0.47}}$ & $99.63_{\pm \text{\tiny 0.42}}$ & $97.49_{\pm \text{\tiny 0.43}}$ & $40.85_{\pm \text{\tiny 1.57}}$ & $2.70_{\pm \text{\tiny 0.46}}$ \\
SCRUB & $95.44_{\pm \text{\tiny 0.52}}$ & $97.27_{\pm \text{\tiny 0.88}}$ & $96.52_{\pm \text{\tiny 0.76}}$ & $51.01_{\pm \text{\tiny 0.25}}$ & $1.42_{\pm \text{\tiny 0.35}}$ \\
SSD & $96.84_{\pm \text{\tiny 0.28}}$ & $99.88_{\pm \text{\tiny 0.16}}$ & $99.90_{\pm \text{\tiny 0.10}}$ & $51.72_{\pm \text{\tiny 0.25}}$ & $1.44_{\pm \text{\tiny 0.17}}$ \\
AMUN & $96.34_{\pm \text{\tiny 0.08}}$ & $99.61_{\pm \text{\tiny 0.19}}$ & $98.94_{\pm \text{\tiny 0.34}}$ & $51.92_{\pm \text{\tiny 0.55}}$ & $1.44_{\pm \text{\tiny 0.22}}$ \\
CoUn & $97.90_{\pm \text{\tiny 0.23}}$ & $99.99_{\pm \text{\tiny 0.01}}$ & $98.49_{\pm \text{\tiny 0.20}}$ & $51.30_{\pm \text{\tiny 0.74}}$ & $1.20_{\pm \text{\tiny 0.24}}$ \\
\midrule
\textbf{MUNKEY} & $97.72_{\pm \text{\tiny 0.08}}$ & $99.63_{\pm \text{\tiny 0.12}}$ & $99.50_{\pm \text{\tiny 0.18}}$ & $50.66_{\pm \text{\tiny 0.50}}$ & $\textbf{0.92}_{\pm \text{\tiny 0.19}}$ \\
\bottomrule
\end{tabular}
\end{table}

\begin{table}[h!]
\centering
\caption{Unlearning performance on a random forget set of 10\% of the training data on PathMNIST.}
\vspace*{0.1cm}
\label{tab:unlearning-pathmnist}
\footnotesize
\setlength{\tabcolsep}{5.2pt} 
\begin{tabular}{l cccc >{\columncolor{lightgray}}c}
\toprule
\textbf{Model} & \textbf{TA} & \textbf{RA} & \textbf{FA} & \textbf{MIA} & \textbf{Avg Gap $\downarrow$} \\
\midrule
Retrain \text{\tiny (T)} & $98.64_{\pm \text{\tiny 0.01}}$ & $100.00_{\pm \text{\tiny 0.00}}$ & $98.53_{\pm \text{\tiny 0.05}}$ & $49.95_{\pm \text{\tiny 0.15}}$ & -- \\
Retrain \text{\tiny (M)} & $98.96_{\pm \text{\tiny 0.07}}$ & $100.00_{\pm \text{\tiny 0.00}}$  & $98.78_{\pm \text{\tiny 0.17}}$ & $50.09_{\pm \text{\tiny 0.24}}$ & -- \\
\midrule
KNN & $96.69_{\pm \text{\tiny 0.08}}$ & $97.10_{\pm \text{\tiny 0.01}}$ & $96.81_{\pm \text{\tiny 0.17}}$ & $50.27_{\pm \text{\tiny 0.17}}$ & -- \\
CF & $98.30_{\pm \text{\tiny 0.08}}$ & $99.87_{\pm \text{\tiny 0.08}}$ & $99.77_{\pm \text{\tiny 0.12}}$ & $51.31_{\pm \text{\tiny 0.27}}$ & $0.77_{\pm \text{\tiny 0.09}}$ \\
CF-k & $98.56_{\pm \text{\tiny 0.08}}$ & $99.99_{\pm \text{\tiny 0.00}}$ & $100.00_{\pm \text{\tiny 0.01}}$ & $51.89_{\pm \text{\tiny 0.24}}$ & $0.88_{\pm \text{\tiny 0.07}}$ \\
EU-k & $98.54_{\pm \text{\tiny 0.04}}$ & $99.99_{\pm \text{\tiny 0.00}}$ & $100.00_{\pm \text{\tiny 0.01}}$ & $51.92_{\pm \text{\tiny 0.26}}$ & $0.89_{\pm \text{\tiny 0.08}}$ \\
$\ell_1$-sparse & $92.12_{\pm \text{\tiny 0.70}}$ & $92.43_{\pm \text{\tiny 0.66}}$ & $91.89_{\pm \text{\tiny 0.65}}$ & $49.88_{\pm \text{\tiny 0.24}}$ & $5.20_{\pm \text{\tiny 0.30}}$ \\
NegGrad+ & $97.92_{\pm \text{\tiny 0.25}}$ & $99.80_{\pm \text{\tiny 0.11}}$ & $98.26_{\pm \text{\tiny 0.19}}$ & $50.74_{\pm \text{\tiny 0.12}}$ & $\textbf{0.50}_{\pm \text{\tiny 0.10}}$ \\
Amnesiac & $97.79_{\pm \text{\tiny 0.16}}$ & $99.93_{\pm \text{\tiny 0.02}}$ & $94.14_{\pm \text{\tiny 0.23}}$ & $32.59_{\pm \text{\tiny 1.51}}$ & $5.67_{\pm \text{\tiny 0.39}}$ \\
SCRUB & $98.08_{\pm \text{\tiny 0.15}}$ & $99.69_{\pm \text{\tiny 0.07}}$ & $99.33_{\pm \text{\tiny 0.12}}$ & $51.19_{\pm \text{\tiny 0.16}}$  & $0.73_{\pm \text{\tiny 0.07}}$ \\
SSD & $98.31_{\pm \text{\tiny 0.20}}$ & $99.95_{\pm \text{\tiny 0.04}}$ & $99.96_{\pm \text{\tiny 0.03}}$ & $51.49_{\pm \text{\tiny 0.36}}$ & $0.84_{\pm \text{\tiny 0.11}}$ \\
AMUN & $98.16_{\pm \text{\tiny 0.00}}$ & $99.87_{\pm \text{\tiny 0.03}}$ & $99.71_{\pm \text{\tiny 0.02}}$ & $51.14_{\pm \text{\tiny 0.18}}$ & $0.75_{\pm \text{\tiny 0.06}}$ \\
CoUn & $98.29_{\pm \text{\tiny 0.67}}$ & $99.90_{\pm \text{\tiny 0.03}}$ & $99.38_{\pm \text{\tiny 0.05}}$ & $51.90_{\pm \text{\tiny 0.57}}$ & $0.81_{\pm \text{\tiny 0.22}}$ \\
\midrule
\textbf{MUNKEY} & $98.75_{\pm \text{\tiny 0.12}}$ & $100.00_{\pm \text{\tiny 0.00}}$ & $99.95_{\pm \text{\tiny 0.02}}$ & $50.77_{\pm \text{\tiny 0.08}}$ & $\underline{0.52}_{\pm \text{\tiny 0.08}}$ \\
\bottomrule
\end{tabular}
\end{table}

Results on CIFAR-100 and Tiny ImageNet, shown in~\Cref{tab:unlearning-cifar100} and~\Cref{tab:unlearning-tinyimagenet}, respectively, highlight our performance on fine-grained data with a high number of classes, a generally more challenging setting where MUNKEY maintains a superior base performance and utility boost (in task accuracy) thanks to its memory-augmented architecture. While we evaluate final unlearning based on the Average Gap, a fair metric with respect to the baselines since their Retrain \text{\tiny (T)} oracle also lacks the external memory, it is notable that the memory bank further enhances absolute accuracies. Tiny ImageNet serves as an extreme case example where the data has been seen, in a different resolution, by the pre-trained key encoder; we show that MUNKEY is able to outperform in this setting as well. The relatively low baseline accuracies here are due to the lack of extensive data augmentation, as our primary objective is evaluating unlearning capabilities rather than maximizing task accuracy. Notably, MUNKEY still performs better than the KNN baseline on Tiny ImageNet, demonstrating that the backbone encoding effectively improves performance and makes MUNKEY behave as ``the best of both worlds''. To ensure clarity regarding our unlearning claims in this last dataset, we emphasize that it occurs at the output level and distinguish between two levels of membership: pre-training membership, where the key encoder may have seen an image during its initial training, is considered out of scope for our algorithm; however, task-specific membership, which refers to whether an image was used to train the current model for its specific task, is exactly what we target for unlearning. This distinction is critical as we assume a practical deployment setting where a black-box adversary only has access to the model's output space; by utilizing the black-box membership inference attack at the output level as claimed throughout this work, we verify that task-specific influence is successfully neutralized.

\begin{table}[h!]
\centering
\caption{Unlearning performance on a random forget set of 10\% of the training data on CIFAR-100.}
\vspace*{0.1cm}
\label{tab:unlearning-cifar100}
\footnotesize
\setlength{\tabcolsep}{5.2pt} 
\begin{tabular}{l cccc >{\columncolor{lightgray}}c}
\toprule
\textbf{Model} & \textbf{TA} & \textbf{RA} & \textbf{FA} & \textbf{MIA} & \textbf{Avg Gap $\downarrow$} \\
\midrule
Retrain \text{\tiny (T)} & $47.17_{\pm \text{\tiny 0.61}}$ & $87.37_{\pm \text{\tiny 0.41}}$ & $47.00_{\pm \text{\tiny 0.39}}$ & $49.88_{\pm \text{\tiny 0.26}}$ & -- \\
Retrain \text{\tiny (M)} & $76.19_{\pm \text{\tiny 0.48}}$ & $99.84_{\pm \text{\tiny 0.05}}$ & $76.05_{\pm \text{\tiny 0.17}}$ & $49.91_{\pm \text{\tiny 0.30}}$ & -- \\

\midrule
KNN & $71.86_{\pm \text{\tiny 0.31}}$ & $81.67_{\pm \text{\tiny 0.07}}$ & $71.65_{\pm \text{\tiny 0.36}}$ & $49.88_{\pm \text{\tiny 0.35}}$ & -- \\
CF & $46.69_{\pm \text{\tiny 1.04}}$ & $90.68_{\pm \text{\tiny 0.67}}$ & $76.94_{\pm \text{\tiny 0.95}}$ & $68.79_{\pm \text{\tiny 0.17}}$ & $13.16_{\pm \text{\tiny 0.45}}$ \\
CF-k & $49.03_{\pm \text{\tiny 0.48}}$ & $94.71_{\pm \text{\tiny 0.19}}$ & $94.51_{\pm \text{\tiny 0.44}}$ & $77.88_{\pm \text{\tiny 0.45}}$ & $21.18_{\pm \text{\tiny 0.30}}$ \\
EU-k & $49.04_{\pm \text{\tiny 0.56}}$ & $92.75_{\pm \text{\tiny 0.26}}$ & $92.61_{\pm \text{\tiny 0.41}}$ & $77.58_{\pm \text{\tiny 0.42}}$ & $20.14_{\pm \text{\tiny 0.30}}$ \\
$\ell_1$-sparse & $33.40_{\pm \text{\tiny 0.58}}$ & $35.49_{\pm \text{\tiny 0.77}}$ & $33.48_{\pm \text{\tiny 1.22}}$ & $50.12_{\pm \text{\tiny 0.79}}$ & $19.85_{\pm \text{\tiny 0.49}}$ \\
NegGrad+ & $46.70_{\pm \text{\tiny 0.57}}$ & $89.80_{\pm \text{\tiny 0.13}}$ & $61.94_{\pm \text{\tiny 0.71}}$ & $60.50_{\pm \text{\tiny 0.45}}$ & $\underline{7.12}_{\pm \text{\tiny 0.34}}$ \\
Amnesiac & $46.28_{\pm \text{\tiny 0.64}}$ & $90.99_{\pm \text{\tiny 0.43}}$ & $67.50_{\pm \text{\tiny 0.81}}$ & $62.64_{\pm \text{\tiny 0.53}}$ & $9.44_{\pm \text{\tiny 0.38}}$ \\
SCRUB & $47.47_{\pm \text{\tiny 0.64}}$ & $84.93_{\pm \text{\tiny 0.15}}$ & $76.73_{\pm \text{\tiny 0.79}}$ & $68.70_{\pm \text{\tiny 0.45}}$ & $12.82_{\pm \text{\tiny 0.36}}$ \\
SSD & $46.60_{\pm \text{\tiny 1.69}}$ & $89.58_{\pm \text{\tiny 3.41}}$ & $87.55_{\pm \text{\tiny 3.93}}$ & $76.23_{\pm \text{\tiny 1.52}}$ & $17.42_{\pm \text{\tiny 1.44}}$ \\
AMUN & $46.82_{\pm \text{\tiny 1.26}}$ & $90.04_{\pm \text{\tiny 1.22}}$ & $75.09_{\pm \text{\tiny 1.06}}$ & $67.38_{\pm \text{\tiny 0.45}}$ & $12.15_{\pm \text{\tiny 0.57}}$ \\
CoUn & $50.39_{\pm \text{\tiny 0.46}}$ & $99.98_{\pm \text{\tiny 0.00}}$ & $81.01_{\pm \text{\tiny 0.54}}$ & $65.87_{\pm \text{\tiny 0.11}}$ & $16.46_{\pm \text{\tiny 0.28}}$ \\
\midrule
\textbf{MUNKEY} & $76.62_{\pm \text{\tiny 0.51}}$ & $99.85_{\pm \text{\tiny 0.03}}$ & $80.80_{\pm \text{\tiny 0.36}}$ & $53.31_{\pm \text{\tiny 0.11}}$ & $\textbf{2.15}_{\pm \text{\tiny 0.22}}$ \\
\bottomrule
\end{tabular}
\end{table}

\begin{table}[h!]
\centering
\caption{Unlearning performance on a random forget set of 10\% of the training data on Tiny ImageNet.}
\vspace*{0.1cm}
\label{tab:unlearning-tinyimagenet}
\footnotesize
\setlength{\tabcolsep}{5.2pt} 
\begin{tabular}{l cccc >{\columncolor{lightgray}}c}
\toprule
\textbf{Model} & \textbf{TA} & \textbf{RA} & \textbf{FA} & \textbf{MIA} & \textbf{Avg Gap $\downarrow$} \\
\midrule
Retrain \text{\tiny (T)} & $36.74_{\pm \text{\tiny 0.76}}$ & $76.78_{\pm \text{\tiny 0.29}}$ & $36.30_{\pm \text{\tiny 0.49}}$ & $49.87_{\pm \text{\tiny 0.23}}$ & -- \\
Retrain \text{\tiny (M)} & $70.63_{\pm \text{\tiny 0.52}}$ & $99.94_{\pm \text{\tiny 0.00}}$ & $70.62_{\pm \text{\tiny 0.43}}$ & $50.10_{\pm \text{\tiny 0.42}}$  & -- \\
\midrule
KNN & $66.56_{\pm \text{\tiny 0.31}}$ & $77.46_{\pm \text{\tiny 0.08}}$ & $66.78_{\pm \text{\tiny 0.52}}$ & $49.99_{\pm \text{\tiny 0.24}}$ & -- \\
CF & $35.46_{\pm \text{\tiny 0.37}}$ & $88.20_{\pm \text{\tiny 0.57}}$ & $68.24_{\pm \text{\tiny 0.97}}$ & $72.71_{\pm \text{\tiny 0.25}}$ & $16.87_{\pm \text{\tiny 0.39}}$ \\
CF-k & $36.98_{\pm \text{\tiny 0.55}}$ & $92.58_{\pm \text{\tiny 0.32}}$ & $92.26_{\pm \text{\tiny 0.54}}$ & $83.12_{\pm \text{\tiny 0.36}}$ & $26.31_{\pm \text{\tiny 0.33}}$ \\
EU-k & $36.84_{\pm \text{\tiny 0.40}}$ & $88.64_{\pm \text{\tiny 0.33}}$ & $88.01_{\pm \text{\tiny 0.60}}$ & $81.54_{\pm \text{\tiny 0.09}}$ & $23.84_{\pm \text{\tiny 0.32}}$ \\
$\ell_1$-sparse & $21.19_{\pm \text{\tiny 0.63}}$ & $22.26_{\pm \text{\tiny 0.53}}$ & $21.23_{\pm \text{\tiny 0.46}}$ & $50.37_{\pm \text{\tiny 0.15}}$ & $21.41_{\pm \text{\tiny 0.34}}$ \\
NegGrad+ & $34.73_{\pm \text{\tiny 0.25}}$ & $86.54_{\pm \text{\tiny 0.61}}$ & $49.34_{\pm \text{\tiny 0.21}}$ & $62.31_{\pm \text{\tiny 0.05}}$ & $\underline{9.31}_{\pm \text{\tiny 0.30}}$ \\
Amnesiac & $34.94_{\pm \text{\tiny 0.45}}$ & $88.36_{\pm \text{\tiny 0.34}}$ & $58.47_{\pm \text{\tiny 0.31}}$ & $68.11_{\pm \text{\tiny 0.21}}$ & $13.45_{\pm \text{\tiny 0.30}}$ \\
SCRUB & $36.28_{\pm \text{\tiny 0.37}}$ & $86.10_{\pm \text{\tiny 1.03}}$ & $73.64_{\pm \text{\tiny 0.69}}$ & $74.94_{\pm \text{\tiny 0.33}}$ & $18.05_{\pm \text{\tiny 0.41}}$ \\
SSD & $36.48_{\pm \text{\tiny 0.60}}$ & $91.01_{\pm \text{\tiny 0.41}}$ & $90.59_{\pm \text{\tiny 0.49}}$ & $82.86_{\pm \text{\tiny 0.35}}$ & $25.44_{\pm \text{\tiny 0.34}}$ \\
AMUN & $35.32_{\pm \text{\tiny 1.17}}$ & $86.86_{\pm \text{\tiny 1.22}}$ & $65.30_{\pm \text{\tiny 1.40}}$ & $70.80_{\pm \text{\tiny 0.20}}$ & $15.36_{\pm \text{\tiny 0.60}}$ \\
CoUn & $36.86_{\pm \text{\tiny 0.31}}$ & $99.98_{\pm \text{\tiny 0.00}}$ & $69.92_{\pm \text{\tiny 0.12}}$ & $65.62_{\pm \text{\tiny 0.25}}$ & $18.17_{\pm \text{\tiny 0.27}}$ \\
\midrule
\textbf{MUNKEY} & $70.86_{\pm \text{\tiny 0.42}}$ & $99.94_{\pm \text{\tiny 0.00}}$ & $71.73_{\pm \text{\tiny 0.51}}$ & $51.19_{\pm \text{\tiny 0.30}}$ & \textbf{$\text{0.61}_{\pm \text{\tiny 0.27}}$} \\
\bottomrule
\end{tabular}
\end{table}
\newpage
\paragraph{Stress-Test Evaluation with Larger Forget Sets.} To evaluate whether the proposed unlearning by design paradigm remains effective beyond standard settings, we report results on the primary benchmarks under 50\% random forgetting in~\Cref{tab:unlearning-derma-cifar-vittiny-50}, providing a stress test of utility preservation and forgetting quality.

\begin{table}[h!]
\centering
\caption{Comparison of MUNKEY and studied baselines using ViT-Tiny backbone on a random forget set of 50\% and three seeds on DermaMNIST and CIFAR-10.}
\label{tab:unlearning-derma-cifar-vittiny-50}
\vspace{0.2em}
\setlength{\tabcolsep}{2.4pt}
\fontsize{6}{6.7}\selectfont
\begin{tabular}{@{}l cccc >{\columncolor{lightgray}[0pt][0pt]}c cccc >{\columncolor{lightgray}[0pt][0pt]}c@{}}
\toprule
& \multicolumn{5}{c}{\textbf{DermaMNIST}} & \multicolumn{5}{c}{\textbf{CIFAR-10}} \\
\cmidrule(lr){2-6} \cmidrule(lr){7-11}
\textbf{Model}
& \textbf{TA} & \textbf{RA} & \textbf{FA} & \textbf{MIA} & \textbf{Avg Gap $\downarrow$}
& \textbf{TA} & \textbf{RA} & \textbf{FA} & \textbf{MIA} & \textbf{Avg Gap $\downarrow$} \\
\midrule

Retrain \text{\tiny (T)}
& $73.60_{\err{0.47}}$ & $99.98_{\err{0.03}}$ & $73.39_{\err{0.46}}$ & $50.09_{\err{0.61}}$ & --
& $71.62_{\err{0.27}}$ & $87.57_{\err{0.71}}$ & $71.16_{\err{0.44}}$ & $49.80_{\err{0.28}}$ & -- \\

Retrain \text{\tiny (M)}
& $78.84_{\err{0.51}}$ & $99.33_{\err{0.38}}$ & $77.70_{\err{0.46}}$ & $49.86_{\err{0.36}}$ & --
& $91.32_{\err{0.13}}$ & $99.76_{\err{0.05}}$ & $91.41_{\err{0.02}}$ & $49.73_{\err{0.18}}$ & -- \\
\midrule

KNN
& $73.82_{\err{0.00}}$ & $83.19_{\err{0.00}}$ & $71.77_{\err{0.00}}$ & $50.22_{\err{0.00}}$ & --
& $90.57_{\err{0.04}}$ & $94.18_{\err{0.09}}$ & $90.85_{\err{0.06}}$ & $50.36_{\err{0.13}}$ & -- \\

CF
& $77.19_{\err{0.60}}$ & $100.0_{\err{0.00}}$ & $99.75_{\err{0.08}}$ & $67.02_{\err{0.15}}$ & $11.73_{\err{0.27}}$
& $79.82_{\err{0.31}}$ & $97.08_{\err{0.43}}$ & $93.38_{\err{0.54}}$ & $59.54_{\err{1.33}}$ & $12.42_{\err{0.45}}$ \\

CF-k
& $77.04_{\err{0.42}}$ & $99.77_{\err{0.17}}$ & $99.83_{\err{0.12}}$ & $67.28_{\err{0.13}}$ & $11.82_{\err{0.26}}$
& $79.44_{\err{0.66}}$ & $94.13_{\err{0.54}}$ & $94.15_{\err{0.64}}$ & $60.36_{\err{1.29}}$ & $11.98_{\err{0.48}}$ \\

EU-k
& $72.59_{\err{2.35}}$ & $88.57_{\err{3.55}}$ & $88.67_{\err{3.46}}$ & $56.89_{\err{0.62}}$ & $8.63_{\err{1.40}}$
& $79.08_{\err{1.09}}$ & $93.68_{\err{0.61}}$ & $93.77_{\err{0.63}}$ & $59.67_{\err{0.68}}$ & $11.51_{\err{0.45}}$ \\

$\ell_1$-sparse
& $76.33_{\err{0.12}}$ & $95.24_{\err{0.37}}$ & $89.69_{\err{0.52}}$ & $59.69_{\err{0.41}}$ & $8.34_{\err{0.30}}$
& $25.16_{\err{0.52}}$ & $24.09_{\err{0.44}}$ & $23.90_{\err{0.33}}$ & $50.47_{\err{0.57}}$ & $39.47_{\err{0.33}}$ \\

NegGrad+
& $77.09_{\err{0.50}}$ & $100.0_{\err{0.00}}$ & $96.02_{\err{0.37}}$ & $64.75_{\err{0.45}}$ & $10.20_{\err{0.30}}$
& $79.38_{\err{0.71}}$ & $96.71_{\err{0.43}}$ & $91.51_{\err{0.58}}$ & $58.79_{\err{1.24}}$ & $11.56_{\err{0.46}}$ \\

Amnesiac
& $76.09_{\err{0.17}}$ & $100.0_{\err{0.00}}$ & $98.10_{\err{0.30}}$ & $60.70_{\err{0.99}}$ & $9.46_{\err{0.35}}$
& $79.11_{\err{0.56}}$ & $97.01_{\err{0.48}}$ & $91.74_{\err{0.61}}$ & $58.57_{\err{0.98}}$ & $11.57_{\err{0.41}}$ \\

SCRUB
& $69.79_{\err{0.70}}$ & $69.71_{\err{1.24}}$ & $69.35_{\err{0.88}}$ & $50.61_{\err{0.39}}$ & $9.66_{\err{0.48}}$
& $76.75_{\err{1.41}}$ & $85.24_{\err{1.96}}$ & $84.13_{\err{2.09}}$ & $55.62_{\err{0.90}}$ & $\underline{6.56}_{\err{0.86}}$ \\

SSD
& $76.76_{\err{0.49}}$ & $99.62_{\err{0.30}}$ & $99.70_{\err{0.23}}$ & $63.35_{\err{0.39}}$ & $10.77_{\err{0.29}}$
& $79.03_{\err{1.04}}$ & $93.61_{\err{0.58}}$ & $93.64_{\err{0.65}}$ & $58.95_{\err{0.10}}$ & $11.27_{\err{0.41}}$ \\

AMUN
& $76.96_{\err{0.25}}$ & $100.0_{\err{0.00}}$ & $99.49_{\err{0.28}}$ & $66.92_{\err{0.12}}$ & $11.58_{\err{0.25}}$
& $78.18_{\err{0.42}}$ & $94.79_{\err{0.22}}$ & $89.07_{\err{0.29}}$ & $57.11_{\err{0.70}}$ & $9.75_{\err{0.23}}$ \\

CoUn
& $78.05_{\err{0.77}}$ & $99.84_{\err{0.05}}$ & $86.11_{\err{0.38}}$ & $53.80_{\err{0.40}}$ & $\underline{5.26}_{\err{0.33}}$
& $79.24_{\err{0.63}}$ & $100.0_{\err{0.00}}$ & $90.98_{\err{0.55}}$ & $57.15_{\err{0.32}}$ & $11.81_{\err{0.32}}$ \\

\midrule

\textbf{MUNKEY}
& $80.17_{\err{0.76}}$ & $99.78_{\err{0.07}}$ & $81.82_{\err{0.28}}$ & $51.48_{\err{0.66}}$ & $\mathbf{1.88}_{\err{0.34}}$
& $92.01_{\err{0.07}}$ & $99.94_{\err{0.01}}$ & $92.60_{\err{0.13}}$ & $50.45_{\err{0.07}}$ & $\mathbf{0.70}_{\err{0.07}}$ \\

\bottomrule
\end{tabular}
\end{table}

\paragraph{Effect of Forget-Set Structure.}
\label{app:forget-structure}

Beyond varying the fraction of forgotten random samples, we investigate the effect of different forget-set structures on CIFAR-10. We consider three settings: (i) full-class removal, where all 5,000 training instances of one class are forgotten, averaged over the 10 classes; (ii) class-imbalanced forgetting, where 80\% of a 5,000-sample forget set comes from one class and the remaining 20\% from the other nine classes, again repeated over all classes; and (iii) single-instance removal, evaluated over 20 independent leave-one-out experiments. Each run is compared against its corresponding retain-only retraining oracle, with class-matched members and non-members for membership inference to avoid class-composition biases. Across the different forget-set structures, MUNKEY remains close to its corresponding retraining oracle. In particular, full-class removal leads to a substantial reduction in forget accuracy, closely matching retraining, and class-imbalanced forgetting achieves a small Avg Gap of 0.95. For single-instance removal, where Forget Accuracy (FA) denotes the number of correctly classified deleted samples across 20 leave-one-out runs, membership is evaluated using the deleted sample's percentile within a same-class panel of 50 examples, rather than MIA AUC, with 50 corresponding to chance. Overall, these results support the effectiveness of MUNKEY across different deletion structures.

\begin{table}[h!]
\centering
\caption{Effect of forget-set structure on CIFAR-10. Full-class and class-imbalanced removal use 5,000 forgotten samples, while single-instance removal uses one. Each method is evaluated against its corresponding retraining oracle.}
\label{tab:forget-structure}
\vspace*{0.2cm}
\footnotesize
\setlength{\tabcolsep}{4.5pt}
\begin{tabular}{l l c c c c >{\columncolor{lightgray}[3pt][3pt]}c}
\toprule
\textbf{Forget set} & \textbf{Model} & \textbf{TA} & \textbf{RA} & \textbf{FA} & \textbf{Membership} & \textbf{Avg Gap} \\
\midrule

\multirow{2}{*}{Full-class}
& \textbf{MUNKEY}
& $84.53_{\pm \text{\tiny 0.59}}$
& $99.94_{\pm \text{\tiny 0.00}}$
& $3.70_{\pm \text{\tiny 1.89}}$
& $50.92_{\pm \text{\tiny 3.12}}$
& $\mathbf{1.37}_{\pm \text{\tiny 0.99}}$ \\

& Retrain \text{\tiny (M)}
& $84.13_{\pm \text{\tiny 0.77}}$
& $99.95_{\pm \text{\tiny 0.02}}$
& $0.00_{\pm \text{\tiny 0.00}}$
& $49.56_{\pm \text{\tiny 1.40}}$
& -- \\
\midrule

\multirow{2}{*}{80/20 imbalance}
& \textbf{MUNKEY}
& $92.16_{\pm \text{\tiny 0.36}}$
& $99.94_{\pm \text{\tiny 0.00}}$
& $86.57_{\pm \text{\tiny 6.19}}$
& $49.39_{\pm \text{\tiny 2.20}}$
& $\mathbf{0.95}_{\pm \text{\tiny 0.68}}$ \\

& Retrain \text{\tiny (M)}
& $91.79_{\pm \text{\tiny 0.41}}$
& $99.94_{\pm \text{\tiny 0.02}}$
& $83.48_{\pm \text{\tiny 6.92}}$
& $49.07_{\pm \text{\tiny 1.48}}$
& -- \\
\midrule

\multirow{2}{*}{Single-instance}
& \textbf{MUNKEY}
& $92.75_{\pm \text{\tiny 0.00}}$
& $99.94_{\pm \text{\tiny 0.00}}$
& $19/20$
& $53.9$
& -- \\

& Retrain \text{\tiny (M)}
& $92.80_{\pm \text{\tiny 0.11}}$
& $99.94_{\pm \text{\tiny 0.01}}$
& $18/20$
& $49.7$
& -- \\

\bottomrule
\end{tabular}
\end{table}

\clearpage

\section{Robustness Analysis} 
\label{app:ablations}
\subsection{Effect of Pathway Aggregation}
\label{app:variants}

As discussed in~\Cref{sec:results}, the modular nature of the ViT architecture allows for multiple ways to integrate information from the exemplar tokens and the image patches. In this section, we provide a detailed investigation into the architectural paradigms for combining the image sequence $\vz_i$ with the exemplar token $\vv'_i$, as well as the late-fusion strategies of neighbors employed at inference time. We primarily investigate four approaches: (1) \textbf{Input-level Aggregation}, our default strategy where $\vv'_i$ is simply prepended to the input sequence as $\vz_{in} = [\vz_i, \vv'_i]$; (2) \textbf{Deep Prompt}~\citep{jia2022visual}, where the token is re-inserted as a prefix before every transformer block to allow for multi-scale conditioning; (3) \textbf{Cross Attention}, which involves inserting a dedicated layer after the self-attention blocks in the beginning, middle, and last layers to derive keys and values from $\vv'_i$ while using $\vz_i$ as the query; and (4) \textbf{Direct Key}, a baseline where the fixed key from the frozen encoder is directly projected into the input space.

During inference, when multiple neighbors $\mathcal{N}_K(\vx_i)$ are retrieved from the memory bank, we must determine how to aggregate their information. One approach is \textbf{softmax-weighted average}, where we compute a softmax over the cosine similarity scores of the retrieved keys to form a single weighted average token. Alternatively, \textbf{rank-based weighting} derives these weights purely from the ranking of the neighbors to provide a more stable distribution independent of absolute similarity values. Finally, \textbf{ensemble voting} serves as a more robust but computationally intensive method, requiring $K$ separate inference passes to aggregate the resulting logits.

An evaluation across the six studied datasets in~\Cref{tab:aggregation-dermamnist,tab:aggregation-bloodmnist,tab:aggregation-cifar10,tab:aggregation-cifar100,tab:aggregation-pathmnist,tab:aggregation-tinyimagenet} shows that while trends remain largely similar, the choice of strategy significantly impacts the modularity of the model's information flow measured in $P_s$, introduced in~\Cref{sec:methods}. We highlight $P_s$ and Avg Gap as they are the center of focus in this discussion.  We observe that cross attention generally achieves a low Avg Gap but suffers from high $P_s$, indicating an unhealthy over-reliance on specific pathways despite the added architectural complexity. Moreover, the direct key approach often leads not only to high $P_s$ but also to larger Avg Gaps, which empirically motivates the necessity of the learnable exemplar token in our framework. While deep prompting shows promise, it occasionally exhibits high $P_s$ on datasets like Tiny ImageNet and generally yields slightly lower utility than late-fusion methods. Consequently, we adopt the simpler input-level aggregation as our primary method, as it offers the best trade-off between unlearning, simplicity, and balanced feature integration. Finally, regarding late-fusion, all perform generally similarly, with Ensembling providing the smallest Avg Gap and most stable pathway balance. However, if inference time or compute is a constraint, we argue that the Softmax-weighted average performs closely to ensembling and remains superior to standard baselines while being more computationally efficient. For simplicity and ease of exploration and compute requirements, all results shown in this section are evaluated with a simpler membership inference attack that does not require training a large number of shadow models per model variant. This MIA uses a logistic regression classifier on the model's output space, and is trained on the loss, entropy, confidence, and prediction margin to distinguish between train and test samples. As in the main text, an MIA AUROC near 0.5 on $\mathcal{D}_f$ indicates successful unlearning. 

\begin{table}[h!]
\centering
\caption{Results on the effect of pathway aggregation and late-fusion of neighbors on DermaMNIST.}
\vspace*{0.1cm}
\label{tab:aggregation-dermamnist}
\footnotesize
\setlength{\tabcolsep}{4pt}
\begin{tabular}{l l cccc>{\columncolor{lightgray}}c >{\columncolor{lightgray}}c}
\toprule
\textbf{Category} & \textbf{Model Variant} & \textbf{TA} & \textbf{RA} & \textbf{FA} & \textbf{MIA} & \textbf{$P_{s}$} & \textbf{Avg Gap $\downarrow$} \\
\midrule
\multirow{6}{*}{\rotatebox[origin=c]{90}{Retrained}} 
& Direct Key & $81.83_{\pm \text{\tiny 0.68}}$ & $99.25_{\pm \text{\tiny 0.22}}$ & $80.05_{\pm \text{\tiny 1.10}}$ & $48.30_{\pm \text{\tiny 0.98}}$ & -- & -- \\
& Deep Prompt & $79.57_{\pm \text{\tiny 0.43}}$ & $93.85_{\pm \text{\tiny 1.30}}$ & $79.10_{\pm \text{\tiny 0.44}}$ & $50.34_{\pm \text{\tiny 0.84}}$ & -- & -- \\
& Cross Attention & $80.62_{\pm \text{\tiny 0.13}}$ & $97.23_{\pm \text{\tiny 0.05}}$ & $80.33_{\pm \text{\tiny 1.17}}$ & $50.31_{\pm \text{\tiny 0.65}}$ & -- & -- \\
\cmidrule{2-8}
& Ensembling & $80.80_{\pm \text{\tiny 0.54}}$ & $97.47_{\pm \text{\tiny 0.15}}$ & $80.10_{\pm \text{\tiny 0.99}}$ & $50.53_{\pm \text{\tiny 0.89}}$ & -- & -- \\
& Softmax & $79.70_{\pm \text{\tiny 0.82}}$ & $93.37_{\pm \text{\tiny 1.75}}$ & $78.95_{\pm \text{\tiny 0.18}}$ & $50.44_{\pm \text{\tiny 0.77}}$ & -- & -- \\
& Rank & $79.32_{\pm \text{\tiny 0.58}}$ & $88.34_{\pm \text{\tiny 1.37}}$ & $78.57_{\pm \text{\tiny 0.35}}$ & $48.90_{\pm \text{\tiny 0.93}}$ & -- & -- \\
\midrule
\multirow{6}{*}{\rotatebox[origin=c]{90}{\textbf{MUNKEY}}} 
& Direct Key & $81.70_{\pm \text{\tiny 0.74}}$ & $99.30_{\pm \text{\tiny 0.45}}$ & $99.33_{\pm \text{\tiny 0.64}}$ & $61.76_{\pm \text{\tiny 2.00}}$ & $0.2118_{\pm \text{\tiny 0.0120}}$ & $8.23_{\pm \text{\tiny 0.70}}$ \\
& Deep Prompt & $79.60_{\pm \text{\tiny 0.60}}$ & $93.32_{\pm \text{\tiny 1.87}}$ & $83.57_{\pm \text{\tiny 0.53}}$ & $52.47_{\pm \text{\tiny 0.88}}$ & $0.1724_{\pm \text{\tiny 0.0058}}$ & $1.79_{\pm \text{\tiny 0.69}}$ \\
& Cross Attention & $80.47_{\pm \text{\tiny 0.28}}$ & $97.17_{\pm \text{\tiny 0.17}}$ & $83.43_{\pm \text{\tiny 0.95}}$ & $51.08_{\pm \text{\tiny 0.75}}$ & $0.2076_{\pm \text{\tiny 0.0047}}$ & \textbf{$\text{1.02}_{\pm \text{\tiny 0.46}}$} \\
\cmidrule{2-8}
& Ensembling & $81.06_{\pm \text{\tiny 0.41}}$ & $97.64_{\pm \text{\tiny 0.22}}$ & $84.10_{\pm \text{\tiny 1.58}}$ & $52.16_{\pm \text{\tiny 0.53}}$ & $0.1690_{\pm \text{\tiny 0.0037}}$ & $\underline{1.52}_{\pm \text{\tiny 0.56}}$ \\
& Softmax & $79.83_{\pm \text{\tiny 1.33}}$ & $93.14_{\pm \text{\tiny 2.44}}$ & $84.29_{\pm \text{\tiny 0.47}}$ & $52.76_{\pm \text{\tiny 0.58}}$ & $0.1681_{\pm \text{\tiny 0.0055}}$ & $2.01_{\pm \text{\tiny 0.89}}$ \\
& Rank & $79.65_{\pm \text{\tiny 0.86}}$ & $88.42_{\pm \text{\tiny 2.10}}$ & $84.38_{\pm \text{\tiny 0.99}}$ & $53.00_{\pm \text{\tiny 0.56}}$ & $0.1657_{\pm \text{\tiny 0.0062}}$ & $2.58_{\pm \text{\tiny 0.78}}$ \\
\bottomrule
\end{tabular}
\end{table}

\begin{table}[h!]
\centering
\caption{Results on the effect of pathway aggregation and late-fusion of neighbors on CIFAR-10.}
\vspace*{0.1cm}
\label{tab:aggregation-cifar10}
\footnotesize
\setlength{\tabcolsep}{4pt}
\begin{tabular}{l l cccc>{\columncolor{lightgray}}c >{\columncolor{lightgray}}c}
\toprule
\textbf{Category} & \textbf{Model Variant} & \textbf{TA} & \textbf{RA} & \textbf{FA} & \textbf{MIA} & \textbf{$P_{s}$} & \textbf{Avg Gap $\downarrow$} \\
\midrule
\multirow{6}{*}{\rotatebox[origin=c]{90}{Retrained}} 
& Direct Key & $92.41_{\pm \text{\tiny 0.16}}$ & $99.56_{\pm \text{\tiny 0.07}}$ & $92.46_{\pm \text{\tiny 0.10}}$ & $50.11_{\pm \text{\tiny 0.58}}$ & -- & -- \\
& Deep Prompt & $91.56_{\pm \text{\tiny 1.02}}$ & $99.76_{\pm \text{\tiny 0.14}}$ & $91.32_{\pm \text{\tiny 1.05}}$ & $50.22_{\pm \text{\tiny 0.53}}$ & -- & -- \\
& Cross Attention & $92.79_{\pm \text{\tiny 0.02}}$ & $100.0_{\pm \text{\tiny 0.00}}$ & $92.71_{\pm \text{\tiny 0.19}}$ & $50.13_{\pm \text{\tiny 0.19}}$ & -- & -- \\
\cmidrule{2-8}
& Ensembling & $92.93_{\pm \text{\tiny 0.16}}$ & $99.93_{\pm \text{\tiny 0.02}}$ & $92.76_{\pm \text{\tiny 0.45}}$ & $49.97_{\pm \text{\tiny 0.25}}$ & -- & -- \\
& Softmax & $92.53_{\pm \text{\tiny 0.14}}$ & $99.85_{\pm \text{\tiny 0.05}}$ & $92.29_{\pm \text{\tiny 0.46}}$ & $49.95_{\pm \text{\tiny 0.20}}$ & -- & -- \\
& Rank & $92.36_{\pm \text{\tiny 0.03}}$ & $95.69_{\pm \text{\tiny 0.07}}$ & $92.12_{\pm \text{\tiny 0.33}}$ & $49.86_{\pm \text{\tiny 0.14}}$ & -- & -- \\
\midrule
\multirow{6}{*}{\rotatebox[origin=c]{90}{\textbf{MUNKEY}}} 
& Direct Key & $92.08_{\pm \text{\tiny 0.10}}$ & $99.47_{\pm \text{\tiny 0.12}}$ & $99.35_{\pm \text{\tiny 0.18}}$ & $56.16_{\pm \text{\tiny 0.32}}$ & $0.2385_{\pm \text{\tiny 0.0068}}$ & $3.34_{\pm \text{\tiny 0.18}}$ \\
& Deep Prompt & $92.09_{\pm \text{\tiny 0.83}}$ & $99.80_{\pm \text{\tiny 0.13}}$ & $92.99_{\pm \text{\tiny 1.04}}$ & $51.64_{\pm \text{\tiny 0.19}}$ & $0.1880_{\pm \text{\tiny 0.0063}}$ & $0.92_{\pm \text{\tiny 0.52}}$ \\
& Cross Attention & $92.93_{\pm \text{\tiny 0.01}}$ & $99.99_{\pm \text{\tiny 0.00}}$ & $93.15_{\pm \text{\tiny 0.26}}$ & $50.69_{\pm \text{\tiny 0.38}}$ & $0.2921_{\pm \text{\tiny 0.0052}}$ & \textbf{$\text{0.29}_{\pm \text{\tiny 0.13}}$} \\
\cmidrule{2-8}
& Ensembling & $93.12_{\pm \text{\tiny 0.08}}$ & $99.94_{\pm \text{\tiny 0.01}}$ & $93.41_{\pm \text{\tiny 0.58}}$ & $51.45_{\pm \text{\tiny 0.39}}$ & $0.1842_{\pm \text{\tiny 0.0060}}$ & $\underline{0.58}_{\pm \text{\tiny 0.22}}$ \\
& Softmax & $92.48_{\pm \text{\tiny 0.19}}$ & $99.80_{\pm \text{\tiny 0.06}}$ & $93.46_{\pm \text{\tiny 0.64}}$ & $51.51_{\pm \text{\tiny 0.40}}$ & $0.1857_{\pm \text{\tiny 0.0056}}$ & $0.71_{\pm \text{\tiny 0.23}}$ \\
& Rank & $92.38_{\pm \text{\tiny 0.43}}$ & $95.76_{\pm \text{\tiny 0.23}}$ & $93.22_{\pm \text{\tiny 0.42}}$ & $51.33_{\pm \text{\tiny 0.40}}$ & $0.1838_{\pm \text{\tiny 0.0060}}$ & $0.67_{\pm \text{\tiny 0.21}}$ \\
\bottomrule
\end{tabular}
\end{table}

\begin{table}[h!]
\centering
\caption{Results on the effect of pathway aggregation and late-fusion of neighbors on BloodMNIST.}
\vspace*{0.1cm}
\label{tab:aggregation-bloodmnist}
\footnotesize
\setlength{\tabcolsep}{4pt}
\begin{tabular}{l l cccc>{\columncolor{lightgray}}c >{\columncolor{lightgray}}c}
\toprule
\textbf{Category} & \textbf{Model Variant} & \textbf{TA} & \textbf{RA} & \textbf{FA} & \textbf{MIA} & \textbf{$P_{s}$} & \textbf{Avg Gap $\downarrow$} \\
\midrule
\multirow{6}{*}{\rotatebox[origin=c]{90}{Retrained}} 
& Direct Key & $97.97_{\pm \text{\tiny 0.12}}$ & $99.93_{\pm \text{\tiny 0.03}}$ & $98.38_{\pm \text{\tiny 0.48}}$ & $49.25_{\pm \text{\tiny 0.66}}$ & -- & -- \\
& Deep Prompt & $97.52_{\pm \text{\tiny 0.27}}$ & $99.55_{\pm \text{\tiny 0.18}}$ & $97.10_{\pm \text{\tiny 0.26}}$ & $49.29_{\pm \text{\tiny 1.04}}$ & -- & -- \\
& Cross Attention & $97.36_{\pm \text{\tiny 0.04}}$ & $98.57_{\pm \text{\tiny 0.12}}$ & $97.27_{\pm \text{\tiny 0.42}}$ & $49.73_{\pm \text{\tiny 0.25}}$ & -- & -- \\
\cmidrule{2-8}
& Ensembling & $97.55_{\pm \text{\tiny 0.15}}$ & $99.49_{\pm \text{\tiny 0.13}}$ & $97.21_{\pm \text{\tiny 0.22}}$ & $49.60_{\pm \text{\tiny 0.41}}$ & -- & -- \\
& Softmax & $97.62_{\pm \text{\tiny 0.21}}$ & $99.51_{\pm \text{\tiny 0.19}}$ & $97.49_{\pm \text{\tiny 0.24}}$ & $49.63_{\pm \text{\tiny 0.72}}$ & -- & -- \\
& Rank & $97.56_{\pm \text{\tiny 0.10}}$ & $99.62_{\pm \text{\tiny 0.15}}$ & $97.27_{\pm \text{\tiny 0.10}}$ & $49.53_{\pm \text{\tiny 0.49}}$ & -- & -- \\
\midrule
\multirow{6}{*}{\rotatebox[origin=c]{90}{\textbf{MUNKEY}}} 
& Direct Key & $98.13_{\pm \text{\tiny 0.15}}$ & $99.94_{\pm \text{\tiny 0.02}}$ & $99.97_{\pm \text{\tiny 0.04}}$ & $51.43_{\pm \text{\tiny 0.89}}$ & $0.0273_{\pm \text{\tiny 0.0080}}$ & $0.99_{\pm \text{\tiny 0.31}}$ \\
& Deep Prompt & $97.50_{\pm \text{\tiny 0.42}}$ & $99.71_{\pm \text{\tiny 0.09}}$ & $99.55_{\pm \text{\tiny 0.28}}$ & $51.21_{\pm \text{\tiny 0.70}}$ & $0.0388_{\pm \text{\tiny 0.0110}}$ & $1.14_{\pm \text{\tiny 0.35}}$ \\
& Cross Attention & $97.66_{\pm \text{\tiny 0.06}}$ & $98.61_{\pm \text{\tiny 0.05}}$ & $97.94_{\pm \text{\tiny 0.28}}$ & $50.26_{\pm \text{\tiny 0.17}}$ & $0.0432_{\pm \text{\tiny 0.0044}}$ & $\textbf{0.39}_{\pm \text{\tiny 0.15}}$ \\
\cmidrule{2-8}
& Ensembling & $97.72_{\pm \text{\tiny 0.08}}$ & $99.63_{\pm \text{\tiny 0.12}}$ & $99.50_{\pm \text{\tiny 0.18}}$ & $50.66_{\pm \text{\tiny 0.50}}$ & $0.0435_{\pm \text{\tiny 0.0067}}$ & $\underline{0.92}_{\pm \text{\tiny 0.19}}$ \\
& Softmax & $97.60_{\pm \text{\tiny 0.13}}$ & $99.73_{\pm \text{\tiny 0.07}}$ & $99.55_{\pm \text{\tiny 0.04}}$ & $51.11_{\pm \text{\tiny 0.51}}$ & $0.0243_{\pm \text{\tiny 0.0109}}$ & $0.95_{\pm \text{\tiny 0.24}}$ \\
& Rank & $97.73_{\pm \text{\tiny 0.29}}$ & $99.70_{\pm \text{\tiny 0.09}}$ & $99.47_{\pm \text{\tiny 0.16}}$ & $50.99_{\pm \text{\tiny 0.66}}$ & $0.0441_{\pm \text{\tiny 0.0133}}$ & $0.98_{\pm \text{\tiny 0.23}}$ \\
\bottomrule
\end{tabular}
\end{table}

\begin{table}[h!]
\centering
\caption{Results on the effect of pathway aggregation and late-fusion of neighbors on PathMNIST.}
\vspace*{0.1cm}
\label{tab:aggregation-pathmnist}
\footnotesize
\setlength{\tabcolsep}{4pt}
\begin{tabular}{l l cccc>{\columncolor{lightgray}}c >{\columncolor{lightgray}}c}
\toprule
\textbf{Category} & \textbf{Model Variant} & \textbf{TA} & \textbf{RA} & \textbf{FA} & \textbf{MIA} & \textbf{$P_{s}$} & \textbf{Avg Gap $\downarrow$} \\
\midrule
\multirow{6}{*}{\rotatebox[origin=c]{90}{Retrained}} 
& Direct Key & $98.98_{\pm \text{\tiny 0.04}}$ & $100.0_{\pm \text{\tiny 0.00}}$ & $98.94_{\pm \text{\tiny 0.11}}$ & $49.88_{\pm \text{\tiny 0.30}}$ & -- & -- \\
& Deep Prompt & $99.04_{\pm \text{\tiny 0.03}}$ & $100.0_{\pm \text{\tiny 0.00}}$ & $98.90_{\pm \text{\tiny 0.11}}$ & $49.98_{\pm \text{\tiny 0.25}}$ & -- & -- \\
& Cross Attention & $98.68_{\pm \text{\tiny 0.07}}$ & $99.97_{\pm \text{\tiny 0.01}}$ & $98.55_{\pm \text{\tiny 0.12}}$ & $49.98_{\pm \text{\tiny 0.07}}$ & -- & -- \\
\cmidrule{2-8}
& Ensembling & $98.96_{\pm \text{\tiny 0.07}}$ & $100.00_{\pm \text{\tiny 0.00}}$  & $98.78_{\pm \text{\tiny 0.17}}$ & $50.09_{\pm \text{\tiny 0.24}}$ & -- & -- \\
& Softmax & $99.24_{\pm \text{\tiny 0.08}}$ & $100.0_{\pm \text{\tiny 0.00}}$ & $99.23_{\pm \text{\tiny 0.05}}$ & $49.74_{\pm \text{\tiny 0.11}}$ & -- & -- \\
& Rank & $99.05_{\pm \text{\tiny 0.04}}$ & $100.0_{\pm \text{\tiny 0.00}}$ & $98.91_{\pm \text{\tiny 0.10}}$ & $49.86_{\pm \text{\tiny 0.22}}$ & -- & -- \\
\midrule
\multirow{6}{*}{\rotatebox[origin=c]{90}{\textbf{MUNKEY}}} 
& Direct Key & $98.99_{\pm \text{\tiny 0.05}}$ & $100.0_{\pm \text{\tiny 0.00}}$ & $100.0_{\pm \text{\tiny 0.01}}$ & $51.22_{\pm \text{\tiny 0.22}}$ & $0.0206_{\pm \text{\tiny 0.0265}}$ & $0.60_{\pm \text{\tiny 0.10}}$ \\
& Deep Prompt & $99.03_{\pm \text{\tiny 0.05}}$ & $100.0_{\pm \text{\tiny 0.00}}$ & $100.0_{\pm \text{\tiny 0.01}}$ & $51.17_{\pm \text{\tiny 0.30}}$ & $0.0636_{\pm \text{\tiny 0.0201}}$ & $0.58_{\pm \text{\tiny 0.10}}$ \\
& Cross Attention & $98.76_{\pm \text{\tiny 0.01}}$ & $99.97_{\pm \text{\tiny 0.01}}$ & $99.97_{\pm \text{\tiny 0.02}}$ & $50.81_{\pm \text{\tiny 0.14}}$ & $0.0040_{\pm \text{\tiny 0.0006}}$ & $0.58_{\pm \text{\tiny 0.05}}$ \\
\cmidrule{2-8}
& Ensembling & $98.75_{\pm \text{\tiny 0.12}}$ & $100.00_{\pm \text{\tiny 0.00}}$ & $99.95_{\pm \text{\tiny 0.02}}$ & $50.77_{\pm \text{\tiny 0.08}}$ & $0.0743_{\pm \text{\tiny 0.0050}}$ & $\underline{0.52}_{\pm \text{\tiny 0.08}}$ \\
& Softmax & $99.30_{\pm \text{\tiny 0.03}}$ & $100.0_{\pm \text{\tiny 0.00}}$ & $100.0_{\pm \text{\tiny 0.00}}$ & $50.74_{\pm \text{\tiny 0.24}}$ & $0.0049_{\pm \text{\tiny 0.0010}}$ & \textbf{$\text{0.46}_{\pm \text{\tiny 0.07}}$} \\
& Rank & $99.08_{\pm \text{\tiny 0.05}}$ & $100.0_{\pm \text{\tiny 0.00}}$ & $100.0_{\pm \text{\tiny 0.01}}$ & $51.43_{\pm \text{\tiny 0.13}}$ & $0.0737_{\pm \text{\tiny 0.0180}}$ & $0.67_{\pm \text{\tiny 0.07}}$ \\
\bottomrule
\end{tabular}
\end{table}

\begin{table}[h!]
\centering
\caption{Results on the effect of pathway aggregation and late-fusion of neighbors on CIFAR-100.}
\vspace*{0.1cm}
\label{tab:aggregation-cifar100}
\footnotesize
\setlength{\tabcolsep}{4pt}
\begin{tabular}{l l cccc>{\columncolor{lightgray}}c >{\columncolor{lightgray}}c}
\toprule
\textbf{Category} & \textbf{Model Variant} & \textbf{TA} & \textbf{RA} & \textbf{FA} & \textbf{MIA} & \textbf{$P_{s}$} & \textbf{Avg Gap $\downarrow$} \\
\midrule
\multirow{6}{*}{\rotatebox[origin=c]{90}{Retrained}} 
& Direct Key & $75.40_{\pm \text{\tiny 0.47}}$ & $99.48_{\pm \text{\tiny 0.33}}$ & $74.77_{\pm \text{\tiny 0.53}}$ & $49.44_{\pm \text{\tiny 0.66}}$ & -- & -- \\
& Deep Prompt & $73.31_{\pm \text{\tiny 0.78}}$ & $99.77_{\pm \text{\tiny 0.05}}$ & $73.44_{\pm \text{\tiny 0.69}}$ & $49.91_{\pm \text{\tiny 0.27}}$ & -- & -- \\
& Cross Attention & $76.23_{\pm \text{\tiny 0.35}}$ & $99.92_{\pm \text{\tiny 0.01}}$ & $76.17_{\pm \text{\tiny 0.30}}$ & $50.02_{\pm \text{\tiny 0.27}}$ & -- & -- \\
\cmidrule{2-8}
& Ensembling & $76.19_{\pm \text{\tiny 0.48}}$ & $99.84_{\pm \text{\tiny 0.05}}$ & $76.05_{\pm \text{\tiny 0.17}}$ & $49.91_{\pm \text{\tiny 0.30}}$ & -- & -- \\
& Softmax & $73.28_{\pm \text{\tiny 0.55}}$ & $99.76_{\pm \text{\tiny 0.07}}$ & $73.24_{\pm \text{\tiny 0.43}}$ & $49.88_{\pm \text{\tiny 0.03}}$ & -- & -- \\
& Rank & $72.94_{\pm \text{\tiny 0.50}}$ & $87.11_{\pm \text{\tiny 0.46}}$ & $72.58_{\pm \text{\tiny 0.40}}$ & $49.85_{\pm \text{\tiny 0.21}}$ & -- & -- \\
\midrule
\multirow{6}{*}{\rotatebox[origin=c]{90}{\textbf{MUNKEY}}} 
& Direct Key & $75.92_{\pm \text{\tiny 0.46}}$ & $99.46_{\pm \text{\tiny 0.51}}$ & $99.48_{\pm \text{\tiny 0.50}}$ & $69.04_{\pm \text{\tiny 0.49}}$ & $0.4162_{\pm \text{\tiny 0.0285}}$ & $11.21_{\pm \text{\tiny 0.35}}$ \\
& Deep Prompt & $73.49_{\pm \text{\tiny 0.69}}$ & $99.74_{\pm \text{\tiny 0.05}}$ & $81.99_{\pm \text{\tiny 0.20}}$ & $55.04_{\pm \text{\tiny 0.52}}$ & $0.3084_{\pm \text{\tiny 0.0008}}$ & $3.47_{\pm \text{\tiny 0.35}}$ \\
& Cross Attention & $76.42_{\pm \text{\tiny 0.45}}$ & $99.91_{\pm \text{\tiny 0.01}}$ & $78.86_{\pm \text{\tiny 0.26}}$ & $51.63_{\pm \text{\tiny 0.22}}$ & $0.5019_{\pm \text{\tiny 0.0055}}$ & \textbf{$\text{1.13}_{\pm \text{\tiny 0.19}}$} \\
\cmidrule{2-8}
& Ensembling & $76.62_{\pm \text{\tiny 0.51}}$ & $99.85_{\pm \text{\tiny 0.03}}$ & $80.80_{\pm \text{\tiny 0.36}}$ & $53.31_{\pm \text{\tiny 0.11}}$ & $0.2993_{\pm \text{\tiny 0.0053}}$ & $\underline{2.15}_{\pm \text{\tiny 0.22}}$ \\
& Softmax & $73.32_{\pm \text{\tiny 1.2}}$ & $93.19_{\pm \text{\tiny 4.63}}$ & $79.77_{\pm \text{\tiny 1.69}}$ & $54.20_{\pm \text{\tiny 0.43}}$ & $0.2966_{\pm \text{\tiny 0.0001}}$ & $4.37_{\pm \text{\tiny 1.29}}$ \\
& Rank & $73.49_{\pm \text{\tiny 0.82}}$ & $87.63_{\pm \text{\tiny 0.27}}$ & $81.74_{\pm \text{\tiny 0.37}}$ & $55.46_{\pm \text{\tiny 0.59}}$ & $0.3058_{\pm \text{\tiny 0.0028}}$ & $3.96_{\pm \text{\tiny 0.34}}$ \\
\bottomrule
\end{tabular}
\end{table}

\begin{table}[h!]
\centering
\caption{Results on the effect of pathway aggregation and late-fusion of neighbors on Tiny ImageNet.}
\vspace*{0.1cm}
\label{tab:aggregation-tinyimagenet}
\footnotesize
\setlength{\tabcolsep}{4pt}
\begin{tabular}{l l cccc>{\columncolor{lightgray}}c >{\columncolor{lightgray}}c}
\toprule
\textbf{Category} & \textbf{Model Variant} & \textbf{TA} & \textbf{RA} & \textbf{FA} & \textbf{MIA} & \textbf{$P_{s}$} & \textbf{Avg Gap $\downarrow$} \\
\midrule
\multirow{6}{*}{\rotatebox[origin=c]{90}{Retrained}} 
& Direct Key & $68.98_{\pm \text{\tiny 0.29}}$ & $99.98_{\pm \text{\tiny 0.00}}$ & $68.34_{\pm \text{\tiny 0.70}}$ & $49.93_{\pm \text{\tiny 0.51}}$ & -- & -- \\
& Deep Prompt & $67.75_{\pm \text{\tiny 0.53}}$ & $99.84_{\pm \text{\tiny 0.02}}$ & $67.68_{\pm \text{\tiny 0.59}}$ & $49.83_{\pm \text{\tiny 0.20}}$ & -- & -- \\
& Cross Attention & $70.41_{\pm \text{\tiny 0.16}}$ & $99.94_{\pm \text{\tiny 0.00}}$ & $69.95_{\pm \text{\tiny 0.47}}$ & $49.93_{\pm \text{\tiny 0.20}}$ & -- & -- \\
\cmidrule{2-8}
& Ensembling & $70.63_{\pm \text{\tiny 0.52}}$ & $99.94_{\pm \text{\tiny 0.00}}$ & $70.62_{\pm \text{\tiny 0.43}}$ & $50.10_{\pm \text{\tiny 0.42}}$ & -- & -- \\
& Softmax & $67.70_{\pm \text{\tiny 0.37}}$ & $99.83_{\pm \text{\tiny 0.01}}$ & $67.68_{\pm \text{\tiny 0.60}}$ & $49.82_{\pm \text{\tiny 0.16}}$ & -- & -- \\
& Rank & $67.46_{\pm \text{\tiny 0.47}}$ & $78.75_{\pm \text{\tiny 0.37}}$ & $67.35_{\pm \text{\tiny 0.67}}$ & $49.81_{\pm \text{\tiny 0.55}}$ & -- & -- \\
\midrule
\multirow{6}{*}{\rotatebox[origin=c]{90}{\textbf{MUNKEY}}} 
& Direct Key & $69.02_{\pm \text{\tiny 0.24}}$ & $99.98_{\pm \text{\tiny 0.00}}$ & $99.99_{\pm \text{\tiny 0.01}}$ & $77.30_{\pm \text{\tiny 0.39}}$ & $0.7265_{\pm \text{\tiny 0.0066}}$ & $14.77_{\pm \text{\tiny 0.26}}$ \\
& Deep Prompt & $67.36_{\pm \text{\tiny 0.39}}$ & $99.77_{\pm \text{\tiny 0.05}}$ & $72.19_{\pm \text{\tiny 0.37}}$ & $52.64_{\pm \text{\tiny 0.28}}$ & $0.5917_{\pm \text{\tiny 0.0125}}$ & $1.95_{\pm \text{\tiny 0.25}}$ \\
& Cross Attention & $70.48_{\pm \text{\tiny 0.16}}$ & $99.94_{\pm \text{\tiny 0.01}}$ & $72.05_{\pm \text{\tiny 0.40}}$ & $50.78_{\pm \text{\tiny 0.14}}$ & $0.7014_{\pm \text{\tiny 0.0079}}$ & $\underline{0.76}_{\pm \text{\tiny 0.18}}$ \\
\cmidrule{2-8}
& Ensembling & $70.86_{\pm \text{\tiny 0.42}}$ & $99.94_{\pm \text{\tiny 0.00}}$ & $71.73_{\pm \text{\tiny 0.51}}$ & $51.19_{\pm \text{\tiny 0.30}}$ & $0.2377_{\pm \text{\tiny 0.1053}}$ & \textbf{$\text{0.61}_{\pm \text{\tiny 0.27}}$} \\
& Softmax & $67.35_{\pm \text{\tiny 0.19}}$ & $99.80_{\pm \text{\tiny 0.01}}$ & $72.44_{\pm \text{\tiny 0.45}}$ & $52.74_{\pm \text{\tiny 0.31}}$ & $0.2276_{\pm \text{\tiny 0.1685}}$ & $2.02_{\pm \text{\tiny 0.23}}$ \\
& Rank & $66.79_{\pm \text{\tiny 0.13}}$ & $78.31_{\pm \text{\tiny 0.24}}$ & $71.84_{\pm \text{\tiny 0.40}}$ & $53.30_{\pm \text{\tiny 0.12}}$ & $0.5898_{\pm \text{\tiny 0.0126}}$ & $2.27_{\pm \text{\tiny 0.29}}$ \\
\bottomrule
\end{tabular}
\end{table}

\newpage
\vspace*{2cm}
\subsection{Retrieval Strategy Ablations}
\paragraph{Effect of the Key Encoder.}
\label{app:key-encoder}
We evaluate the impact of the frozen key encoder $g_\phi$ by comparing the standard ViT-Base pre-trained on ImageNet, used throughout the manuscript, against two alternative encoders: DINO-v2\footnote{https://huggingface.co/facebook/dinov2-base}, self-supervised, and trained on curated datasets like Google Landmarks; and CLIP\footnote{https://huggingface.co/openai/clip-vit-base-patch32}, multimodal and trained on broad internet-crawled image-caption pairs. This ablation demonstrates that MUNKEY does not require label-informed keys to effectively decouple and delete instance-specific information. In~\Cref{tab:combined-unlearning} we observe that across both natural and medical datasets, all three pre-training objectives of the key encoder yield robust and comparable results, consistently outperforming post-hoc baselines in both utility and forgetting.

To further isolate the effect of the key encoder and account for cases where the target dataset may have been part of the encoder’s pre-training data, we additionally train a ViT-Tiny $g_\phi$ on only 20\% of the original training set and train MUNKEY on the remaining 80\%. This ensures that the encoder has not seen the instances later used for MUNKEY training or unlearning. As shown in~\Cref{tab:key_encoder_subset}, MUNKEY remains effective in this setting, achieving a low Avg Gap. To contextualize the utility drop in this controlled setting, we additionally train a standard ViT-Tiny from scratch as baseline performance on the same 80\% split, after removing the corresponding forget set, i.e., Retrain (T). This baseline achieves $75.49 \pm 0.22$\% test accuracy on CIFAR-10 and $73.80 \pm 1.10$\% on DermaMNIST, both below our method's performance. Therefore, MUNKEY retains its main unlearning advantages in terms of membership inference and Avg Gap, even when utility is constrained by the weaker scratch-trained encoder. This suggests that strong pre-trained encoders improve utility, while the core unlearning mechanism does not depend on possible pre-training overlap.

\begin{table}[h!]
\centering
\caption{Performance under varying key encoders $g_\phi$ for the memory bank. Results on CIFAR-10 and DermaMNIST.}
\vspace*{0.1cm}
\label{tab:combined-unlearning}
\footnotesize
\setlength{\tabcolsep}{5.2pt}
\begin{tabular}{l l cccc >{\columncolor{lightgray}}c}
\toprule
\textbf{Dataset} & \textbf{Model Variant} & \textbf{TA} & \textbf{RA} & \textbf{FA} & \textbf{MIA} & \textbf{Avg Gap $\downarrow$} \\
\midrule
\multirow{6}{*}{\rotatebox[origin=c]{90}{CIFAR-10}} 
& Retrain (CLIP) & $92.61_{\pm \text{\tiny 0.04}}$ & $98.23_{\pm \text{\tiny 0.02}}$ & $93.19_{\pm \text{\tiny 0.15}}$ & $50.82_{\pm \text{\tiny 0.04}}$ & -- \\
& Retrain (DINO-v2) & $95.32_{\pm \text{\tiny 0.02}}$ & $99.96_{\pm \text{\tiny 0.01}}$ & $95.68_{\pm \text{\tiny 0.22}}$ & $50.71_{\pm \text{\tiny 0.34}}$ & -- \\
& Retrain (ViT/B) & $92.93_{\pm \text{\tiny 0.16}}$ & $99.93_{\pm \text{\tiny 0.02}}$ & $92.76_{\pm \text{\tiny 0.45}}$ & $49.97_{\pm \text{\tiny 0.25}}$ & -- \\
\cmidrule{2-7}
& Key Encoder CLIP & $92.69_{\pm \text{\tiny 0.08}}$ & $98.25_{\pm \text{\tiny 0.09}}$ & $93.84_{\pm \text{\tiny 0.16}}$ & $51.85_{\pm \text{\tiny 0.06}}$ & \underline{0.45}$_{\pm \text{\tiny 0.07}}$ \\
& Key Encoder DINO-v2 & $95.34_{\pm \text{\tiny 0.08}}$ & $99.95_{\pm \text{\tiny 0.02}}$ & $96.04_{\pm \text{\tiny 0.34}}$ & $51.98_{\pm \text{\tiny 0.12}}$ & \textbf{0.42}$_{\pm \text{\tiny 0.14}}$ \\
& Key Encoder Pre-trained-ViT/B & $93.12_{\pm \text{\tiny 0.08}}$ & $99.94_{\pm \text{\tiny 0.01}}$ & $93.41_{\pm \text{\tiny 0.58}}$ & $51.45_{\pm \text{\tiny 0.39}}$ & $0.58_{\pm \text{\tiny 0.22}}$ \\
\midrule
\multirow{6}{*}{\rotatebox[origin=c]{90}{DermaMNIST}} 
& Retrain (CLIP) & $81.08_{\pm \text{\tiny 0.02}}$ & $91.59_{\pm \text{\tiny 0.76}}$ & $79.62_{\pm \text{\tiny 1.41}}$ & $49.58_{\pm \text{\tiny 0.89}}$ & -- \\
& Retrain (DINO-v2) & $81.90_{\pm \text{\tiny 0.29}}$ & $95.66_{\pm \text{\tiny 0.56}}$ & $80.10_{\pm \text{\tiny 0.64}}$ & $50.17_{\pm \text{\tiny 0.91}}$ & -- \\
& Retrain (ViT/B) & $80.80_{\pm \text{\tiny 0.54}}$ & $97.47_{\pm \text{\tiny 0.15}}$ & $80.10_{\pm \text{\tiny 0.99}}$ & $50.53_{\pm \text{\tiny 0.89}}$ & -- \\
\cmidrule{2-7}
& Key Encoder CLIP & $81.43_{\pm \text{\tiny 0.20}}$ & $91.79_{\pm \text{\tiny 0.43}}$ & $84.05_{\pm \text{\tiny 1.87}}$ & $52.09_{\pm \text{\tiny 1.28}}$ & $1.87_{\pm \text{\tiny 0.74}}$ \\
& Key Encoder DINO-v2 & $82.16_{\pm \text{\tiny 0.08}}$ & $96.10_{\pm \text{\tiny 0.16}}$ & $84.52_{\pm \text{\tiny 1.12}}$ & $52.17_{\pm \text{\tiny 1.16}}$ & \underline{1.78}$_{\pm \text{\tiny 0.52}}$ \\
& Key Encoder Pre-trained-ViT/B & $81.06_{\pm \text{\tiny 0.41}}$ & $97.64_{\pm \text{\tiny 0.22}}$ & $84.10_{\pm \text{\tiny 1.58}}$ & $52.16_{\pm \text{\tiny 0.53}}$ & \textbf{1.52}$_{\pm \text{\tiny 0.56}}$ \\
\bottomrule
\end{tabular}
\end{table}

\begin{table}[h!]
\centering
\caption{Performance on encoder $g_\phi$ trained from scratch. We compare the default pre-trained encoder against a key encoder trained on only 20\% of the original training data, with MUNKEY trained on the remaining 80\%. Results are shown on CIFAR-10 and DermaMNIST.}
\vspace*{0.1cm}
\label{tab:key_encoder_subset}
\footnotesize
\renewcommand{\arraystretch}{1.15}
\setlength{\tabcolsep}{5.2pt}
\begin{tabular}{l l cccc >{\columncolor{lightgray}}c}
\toprule
\textbf{Dataset} & \textbf{Model Variant} & \textbf{TA} & \textbf{RA} & \textbf{FA} & \textbf{MIA} & \textbf{Avg Gap $\downarrow$} \\
\midrule

\multirow{4}{*}{\rotatebox[origin=c]{90}{CIFAR-10}}
& Retrain (20\% $\mathcal{D}$ ViT/T)
& $76.72_{\pm \text{\tiny 1.11}}$
& $99.40_{\pm \text{\tiny 0.05}}$
& $76.10_{\pm \text{\tiny 0.73}}$
& $50.34_{\pm \text{\tiny 0.23}}$
& -- \\

& Retrain (ViT/B)
& $92.14_{\pm \text{\tiny 0.09}}$
& $99.96_{\pm \text{\tiny 0.01}}$
& $92.07_{\pm \text{\tiny 0.70}}$
& $50.18_{\pm \text{\tiny 0.22}}$
& -- \\

\cmidrule{2-7}

& Key Encoder 20\% $\mathcal{D}$ ViT/T
& $76.42_{\pm \text{\tiny 0.73}}$
& $99.34_{\pm \text{\tiny 0.07}}$
& $79.11_{\pm \text{\tiny 0.09}}$
& $51.88_{\pm \text{\tiny 0.27}}$
& $1.23_{\pm \text{\tiny 0.39}}$ \\

& Key Encoder Pre-trained-ViT/B
& $92.28_{\pm \text{\tiny 0.16}}$
& $99.95_{\pm \text{\tiny 0.01}}$
& $92.64_{\pm \text{\tiny 0.72}}$
& $51.39_{\pm \text{\tiny 0.56}}$
& $0.48_{\pm \text{\tiny 0.30}}$ \\

\midrule

\multirow{4}{*}{\rotatebox[origin=c]{90}{DermaMNIST}}
& Retrain (20\% $\mathcal{D}$) ViT/T
& $74.11_{\pm \text{\tiny 0.43}}$
& $92.96_{\pm \text{\tiny 0.23}}$
& $73.87_{\pm \text{\tiny 0.22}}$
& $50.37_{\pm \text{\tiny 0.58}}$
& -- \\

& Retrain (ViT/B)
& $81.05_{\pm \text{\tiny 0.25}}$
& $99.81_{\pm \text{\tiny 0.04}}$
& $79.82_{\pm \text{\tiny 0.76}}$
& $49.68_{\pm \text{\tiny 0.80}}$
& -- \\

\cmidrule{2-7}

& Key Encoder 20\% $\mathcal{D}$ ViT/T
& $74.41_{\pm \text{\tiny 0.12}}$
& $93.48_{\pm \text{\tiny 0.41}}$
& $77.50_{\pm \text{\tiny 1.16}}$
& $52.41_{\pm \text{\tiny 0.66}}$
& $1.62_{\pm \text{\tiny 0.40}}$ \\

& Key Encoder Pre-trained-ViT/B
& $80.62_{\pm \text{\tiny 0.18}}$
& $99.80_{\pm \text{\tiny 0.03}}$
& $83.57_{\pm \text{\tiny 1.52}}$
& $51.86_{\pm \text{\tiny 1.33}}$
& $1.59_{\pm \text{\tiny 0.58}}$ \\

\bottomrule
\end{tabular}
\end{table}

\paragraph{Effect of the Neighborhood Size}
\label{app:neighbors}

\begin{figure}[h!]
    \centering
    \begin{overpic}[width=.9\linewidth]{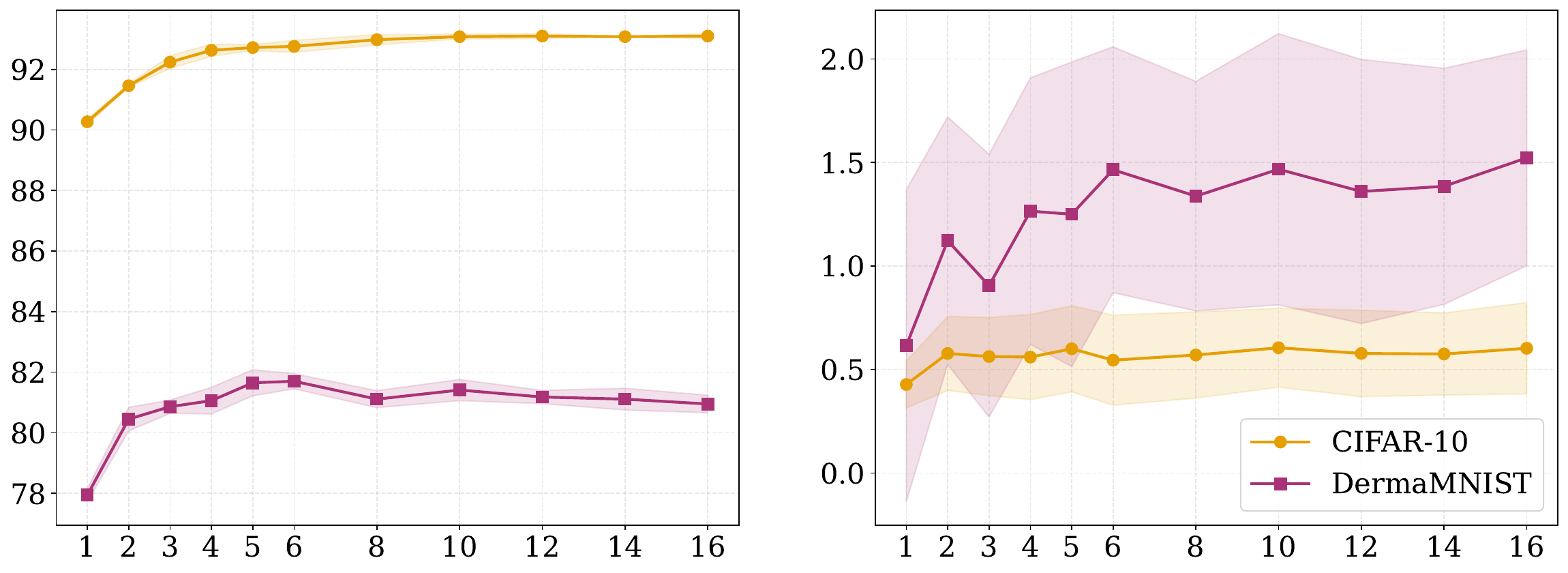} 
        \put(19.5, -2){\scalebox{1}{\# neighbors}}
        \put(71.5, -2){\scalebox{1}{\# neighbors}}
        \put(-3,7){\rotatebox{90}{\scalebox{1}{\texttt{Test Accuracy($\uparrow$)}}}}
        \put(49,11){\rotatebox{90}{\scalebox{1}{\texttt{Avg Gap($\downarrow$)}}}}
        \end{overpic}
    \vspace{0.2cm}
    \caption{Performance sensitivity to the number of neighbors on DermaMNIST and CIFAR-10. We report Test Accuracy (\%) (left) and Avg Gap (right) across different neighborhood sizes.}
    \label{fig:ablation-knn}
\end{figure}

As discussed in Appendix~\ref{app:variants}, the inference process involves fusing information from the $K$ retrieved neighbors $\mathcal{N}_K(\mathbf{x}_q)$. In this section, we investigate the impact of varying the neighborhood size, $K$, on model utility and unlearning efficacy.~\Cref{fig:ablation-knn} illustrates the relationship between the number of neighbors and two key metrics across the DermaMNIST and CIFAR-10 datasets: test accuracy, which represents the model's ability to perform standard classification, and Avg Gap, which serves as a proxy for unlearning performance. In both datasets, we observe that test accuracy increases slightly with the number of neighbors before reaching a plateau. This effect is notably more pronounced on DermaMNIST, likely due to the inherent complexity and fine-grained nature of medical image classification. A similar pattern is seen in Avg Gap, growing slightly before plateauing, all while remaining within small magnitudes. We attribute this to an edge case in model behavior where lower overall accuracy artificially reduces the gap. Consequently, we select $K=4$ for the results of this work, as it provides a robust balance between high classification performance and effective unlearning.\looseness-1

\vspace{0.4cm}

\subsection{Memory Ablations: Varying Exemplar Token Size}
\label{app:exemplarmemory}

\begin{figure}[h!]
    \centering
    \includegraphics[width=0.9\linewidth]{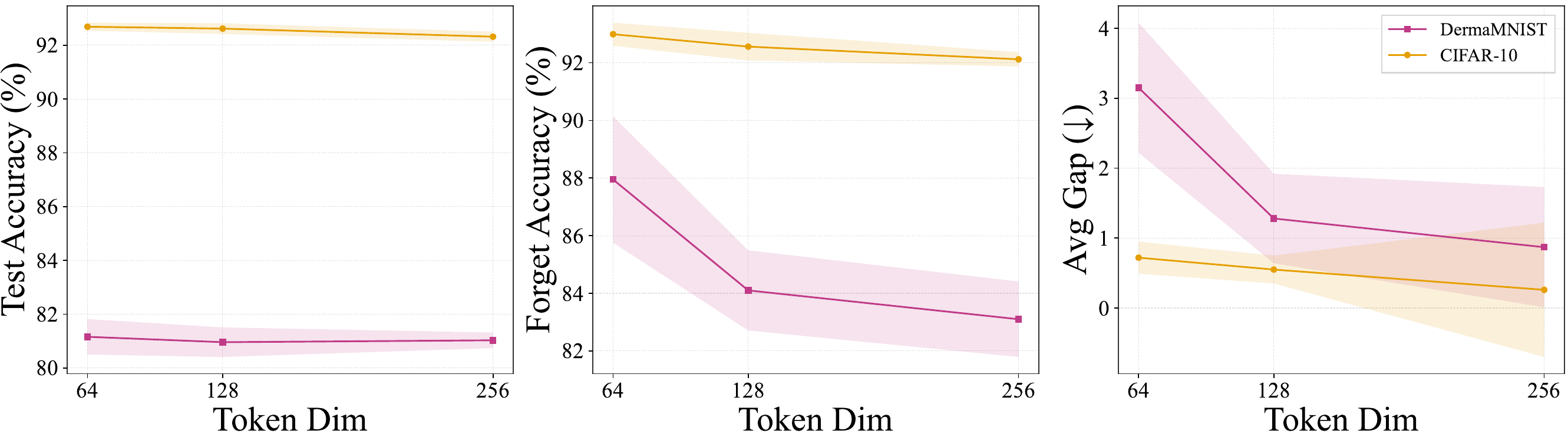}
    \caption{\textbf{Performance of MUNKEY under varying exemplar token size.} The impact of the exemplar token dimensionality ($d_e \in \{64, 128, 256\}$) on model utility and unlearning efficacy is illustrated through Test Accuracy, Forget Accuracy, and Avg Gap. Results on DermaMNIST and CIFAR-10 evaluated on three seeds with a forget fraction of 10\%.}
    \label{fig:token_size}
    \vspace*{-0.225cm}
\end{figure}

We study the effect of memory capacity by varying the dimensionality of the exemplar token, $d_e$, analyzing how the size of the external memory representation influences unlearning, shown in~\Cref{fig:token_size,fig:token_size2,fig:token_size3}.

\begin{table}[h!]
\centering
\caption{Extended comparison of MUNKEY across exemplar token size with random forget rate 10\% on DermaMNIST. We present the accuracy on the test (TA), retain (RA), and forget (FA) sets, as well as membership inference attack (MIA) and average gap values across three seeds.}
\vspace*{0.2cm}
\label{fig:token_size2}
\footnotesize
\begin{tabular}{l  l cccc >{\columncolor{lightgray}}c}
\toprule
\textbf{Model} & \textbf{Token Size} & \textbf{TA} & \textbf{RA} & \textbf{FA} & \textbf{MIA} & \textbf{Avg Gap $\downarrow$} \\
\midrule
Retrain \text{\tiny (M)} & 64  & $80.23_{\pm \text{\tiny 0.80}}$ & $99.76_{\pm \text{\tiny 0.08}}$ & $79.95_{\pm \text{\tiny 1.81}}$ & $49.90_{\pm \text{\tiny 1.08}}$ & -- \\
Retrain \text{\tiny (M)} & 128 & $81.25_{\pm \text{\tiny 0.66}}$ & $99.70_{\pm \text{\tiny 0.08}}$ & $80.29_{\pm \text{\tiny 1.41}}$ & $50.04_{\pm \text{\tiny 1.05}}$ & -- \\
Retrain \text{\tiny (M)} & 256 & $80.98_{\pm \text{\tiny 0.55}}$ & $99.87_{\pm \text{\tiny 0.03}}$ & $80.48_{\pm \text{\tiny 1.52}}$ & $50.08_{\pm \text{\tiny 0.97}}$ & -- \\
\midrule
\textbf{MUNKEY} & 64  & $81.16_{\pm \text{\tiny 0.66}}$ & $99.73_{\pm \text{\tiny 0.04}}$ & $87.95_{\pm \text{\tiny 2.19}}$ & $53.52_{\pm \text{\tiny 1.91}}$ & $3.15_{\pm \text{\tiny 0.93}}$ \\
\textbf{MUNKEY} & 128 & $81.06_{\pm \text{\tiny 0.44}}$ & $99.71_{\pm \text{\tiny 0.08}}$ & $84.00_{\pm \text{\tiny 1.22}}$ & $51.19_{\pm \text{\tiny 1.19}}$ & $1.27_{\pm \text{\tiny 0.64}}$ \\
\textbf{MUNKEY} & 256 & $81.03_{\pm \text{\tiny 0.29}}$ & $99.84_{\pm \text{\tiny 0.05}}$ & $83.10_{\pm \text{\tiny 1.31}}$ & $50.84_{\pm \text{\tiny 1.05}}$ & $0.87_{\pm \text{\tiny 0.86}}$ \\
\bottomrule
\end{tabular}
\end{table}

\begin{table}[h!]
\centering
\caption{Extended comparison of MUNKEY across exemplar token size with random forget rate 10\% on CIFAR-10. We present the accuracy on the test (TA), retain (RA), and forget (FA) sets, as well as membership inference attack (MIA) and average gap values across three seeds.}
\vspace*{0.2cm}
\label{fig:token_size3}
\footnotesize
\begin{tabular}{l  l cccc >{\columncolor{lightgray}}c}
\toprule
\textbf{Model} & \textbf{Token Size} & \textbf{TA} & \textbf{RA} & \textbf{FA} & \textbf{MIA} & \textbf{Avg Gap $\downarrow$} \\
\midrule
Retrain \text{\tiny (M)} & 64  & $92.55_{\pm \text{\tiny 0.18}}$ & $99.82_{\pm \text{\tiny 0.05}}$ & $92.06_{\pm \text{\tiny 0.55}}$ & $49.93_{\pm \text{\tiny 0.35}}$ & -- \\
Retrain \text{\tiny (M)} & 128 & $92.38_{\pm \text{\tiny 0.06}}$ & $99.93_{\pm \text{\tiny 0.03}}$ & $92.05_{\pm \text{\tiny 0.23}}$ & $49.90_{\pm \text{\tiny 0.34}}$ & -- \\
Retrain \text{\tiny (M)} & 256 & $92.28_{\pm \text{\tiny 0.10}}$ & $99.99_{\pm \text{\tiny 0.00}}$ & $91.86_{\pm \text{\tiny 0.21}}$ & $49.84_{\pm \text{\tiny 0.22}}$ & -- \\
\midrule
\textbf{MUNKEY} & 64  & $92.69_{\pm \text{\tiny 0.15}}$ & $99.82_{\pm \text{\tiny 0.03}}$ & $92.99_{\pm \text{\tiny 0.40}}$ & $51.73_{\pm \text{\tiny 0.48}}$ & $0.72_{\pm \text{\tiny 0.23}}$ \\
\textbf{MUNKEY} & 128 & $92.63_{\pm \text{\tiny 0.20}}$ & $99.94_{\pm \text{\tiny 0.01}}$ & $92.53_{\pm \text{\tiny 0.52}}$ & $51.40_{\pm \text{\tiny 0.44}}$ & $0.56_{\pm \text{\tiny 0.21}}$ \\
\textbf{MUNKEY} & 256 & $92.32_{\pm \text{\tiny 0.19}}$ & $100.00_{\pm \text{\tiny 0.00}}$ & $92.12_{\pm \text{\tiny 0.25}}$ & $50.57_{\pm \text{\tiny 0.19}}$ & $0.26_{\pm \text{\tiny 0.12}}$ \\
\bottomrule
\end{tabular}
\end{table}

\subsection{Architecture Scaling}
\label{app:scaling}

We examine whether the benefits of MUNKEY persist under larger transformer architectures, assessing how model capacity interacts with utility and unlearning performance.

\paragraph{Scaling Transformer Depth.} To isolate the role of representational depth, we vary the number of transformer blocks while keeping the overall framework unchanged, analyzing how the unlearning behavior scales with the architecture backbone parameters in~\Cref{tab:unlearning-dermamnist-layers,tab:unlearning-cifar10-layers}.

\begin{table}[h!]
\centering
\caption{Comparison of MUNKEY across number of hidden layers with random forget rate 10\% on DermaMNIST. We present the accuracy on the test (TA), retain (RA), and forget (FA) sets, membership inference attack (MIA), and average gap values across three seeds.}
\label{tab:unlearning-dermamnist-layers}
\vspace*{0.2cm}
\footnotesize
\begin{tabular}{l l l cccc >{\columncolor{lightgray}}c}
\toprule
\textbf{Model} & \textbf{\# Layers} & \textbf{\# Parameters} & \textbf{TA} & \textbf{RA} & \textbf{FA} & \textbf{MIA} & \textbf{Avg Gap $\downarrow$} \\
\midrule
Retrain \text{\tiny (M)} & 10 & 4.6M & $81.15_{\pm \text{\tiny 0.86}}$ & $99.91_{\pm \text{\tiny 0.01}}$ & $80.38_{\pm \text{\tiny 1.30}}$ & $50.49_{\pm \text{\tiny 0.74}}$ & -- \\
Retrain \text{\tiny (M)} & 11 & 5.1M & $81.33_{\pm \text{\tiny 0.36}}$ & $99.93_{\pm \text{\tiny 0.01}}$ & $80.86_{\pm \text{\tiny 1.69}}$ & $50.55_{\pm \text{\tiny 0.89}}$ & -- \\
Retrain \text{\tiny (M)} & 12 \text{\tiny (ViT-T)} & 5.5M & $81.25_{\pm \text{\tiny 0.66}}$ & $99.70_{\pm \text{\tiny 0.08}}$ & $80.29_{\pm \text{\tiny 1.41}}$ & $50.04_{\pm \text{\tiny 1.05}}$ & -- \\
Retrain \text{\tiny (M)} & 13 & 6.0M & $81.25_{\pm \text{\tiny 0.55}}$ & $99.92_{\pm \text{\tiny 0.02}}$ & $80.29_{\pm \text{\tiny 0.73}}$ & $50.50_{\pm \text{\tiny 0.45}}$ & -- \\
Retrain \text{\tiny (M)} & 14 & 6.4M & $81.01_{\pm \text{\tiny 0.12}}$ & $99.89_{\pm \text{\tiny 0.03}}$ & $80.38_{\pm \text{\tiny 1.59}}$ & $50.32_{\pm \text{\tiny 1.22}}$ & -- \\
\midrule
\textbf{MUNKEY} & 10 & 4.6M & $81.80_{\pm \text{\tiny 0.45}}$ & $99.78_{\pm \text{\tiny 0.03}}$ & $84.86_{\pm \text{\tiny 1.53}}$ & $51.53_{\pm \text{\tiny 0.90}}$ & $1.58_{\pm \text{\tiny 0.63}}$ \\
\textbf{MUNKEY} & 11 & 5.1M & $81.68_{\pm \text{\tiny 0.18}}$ & $99.76_{\pm \text{\tiny 0.06}}$ & $85.19_{\pm \text{\tiny 0.66}}$ & $51.86_{\pm \text{\tiny 0.66}}$ & $1.54_{\pm \text{\tiny 0.54}}$ \\
\textbf{MUNKEY} & 12 \text{\tiny (ViT-T)} & 5.5M & $81.06_{\pm \text{\tiny 0.44}}$ & $99.71_{\pm \text{\tiny 0.08}}$ & $84.00_{\pm \text{\tiny 1.22}}$ & $51.19_{\pm \text{\tiny 1.19}}$ & $1.27_{\pm \text{\tiny 0.64}}$ \\
\textbf{MUNKEY} & 13 & 6.0M & $81.88_{\pm \text{\tiny 0.19}}$ & $99.76_{\pm \text{\tiny 0.10}}$ & $84.05_{\pm \text{\tiny 1.79}}$ & $51.48_{\pm \text{\tiny 1.30}}$ & $1.38_{\pm \text{\tiny 0.61}}$ \\
\textbf{MUNKEY} & 14 & 6.4M & $81.56_{\pm \text{\tiny 0.50}}$ & $99.74_{\pm \text{\tiny 0.13}}$ & $83.57_{\pm \text{\tiny 0.88}}$ & $51.29_{\pm \text{\tiny 0.55}}$ & $1.22_{\pm \text{\tiny 0.58}}$ \\
\bottomrule
\end{tabular}
\end{table}

\begin{table}[h!]
\centering
\caption{Comparison of MUNKEY across number of hidden layers with random forget rate 10\% on CIFAR-10. We present the accuracy on the test (TA), retain (RA), and forget (FA) sets, membership inference attack (MIA), and average gap values across three seeds.}
\label{tab:unlearning-cifar10-layers}
\vspace*{0.2cm}
\footnotesize
\begin{tabular}{l l l cccc >{\columncolor{lightgray}}c}
\toprule
\textbf{Model} & \textbf{\# Layers} & \textbf{\# Parameters} & \textbf{TA} & \textbf{RA} & \textbf{FA} & \textbf{MIA} & \textbf{Avg Gap $\downarrow$} \\
\midrule
Retrain \text{\tiny (M)} & 10 & 4.6M & $91.95_{\pm \text{\tiny 0.08}}$ & $100.00_{\pm \text{\tiny 0.00}}$ & $91.52_{\pm \text{\tiny 0.13}}$ & $50.08_{\pm \text{\tiny 0.31}}$ & -- \\
Retrain \text{\tiny (M)} & 11 & 5.1M & $91.94_{\pm \text{\tiny 0.06}}$ & $100.00_{\pm \text{\tiny 0.00}}$ & $91.53_{\pm \text{\tiny 0.33}}$ & $49.89_{\pm \text{\tiny 0.22}}$ & -- \\
Retrain \text{\tiny (M)} & 12 \text{\tiny (ViT-T)} & 5.5M & $92.38_{\pm \text{\tiny 0.06}}$ & $99.93_{\pm \text{\tiny 0.03}}$ & $92.05_{\pm \text{\tiny 0.23}}$ & $49.90_{\pm \text{\tiny 0.34}}$ & -- \\
Retrain \text{\tiny (M)} & 13 & 6.0M & $91.91_{\pm \text{\tiny 0.05}}$ & $100.00_{\pm \text{\tiny 0.00}}$ & $91.57_{\pm \text{\tiny 0.13}}$ & $50.02_{\pm \text{\tiny 0.08}}$ & -- \\
Retrain \text{\tiny (M)} & 14 & 6.4M & $91.91_{\pm \text{\tiny 0.10}}$ & $100.00_{\pm \text{\tiny 0.00}}$ & $91.61_{\pm \text{\tiny 0.10}}$ & $50.13_{\pm \text{\tiny 0.11}}$ & -- \\
\midrule
\textbf{MUNKEY} & 10 & 4.6M & $92.46_{\pm \text{\tiny 0.08}}$ & $99.96_{\pm \text{\tiny 0.01}}$ & $92.51_{\pm \text{\tiny 0.34}}$ & $51.11_{\pm \text{\tiny 0.42}}$ & $0.64_{\pm \text{\tiny 0.16}}$ \\
\textbf{MUNKEY} & 11 & 5.1M & $92.60_{\pm \text{\tiny 0.19}}$ & $99.95_{\pm \text{\tiny 0.01}}$ & $92.60_{\pm \text{\tiny 0.29}}$ & $51.27_{\pm \text{\tiny 0.25}}$ & $0.79_{\pm \text{\tiny 0.15}}$ \\
\textbf{MUNKEY} & 12 \text{\tiny (ViT-T)} & 5.5M & $92.63_{\pm \text{\tiny 0.20}}$ & $99.94_{\pm \text{\tiny 0.01}}$ & $92.53_{\pm \text{\tiny 0.52}}$ & $51.40_{\pm \text{\tiny 0.44}}$ & $0.56_{\pm \text{\tiny 0.21}}$ \\
\textbf{MUNKEY} & 13 & 6.0M & $92.54_{\pm \text{\tiny 0.08}}$ & $99.95_{\pm \text{\tiny 0.02}}$ & $92.72_{\pm \text{\tiny 0.26}}$ & $51.38_{\pm \text{\tiny 0.30}}$ & $0.80_{\pm \text{\tiny 0.11}}$ \\
\textbf{MUNKEY} & 14 & 6.4M & $92.59_{\pm \text{\tiny 0.12}}$ & $99.96_{\pm \text{\tiny 0.01}}$ & $92.67_{\pm \text{\tiny 0.25}}$ & $51.09_{\pm \text{\tiny 0.63}}$ & $0.69_{\pm \text{\tiny 0.18}}$ \\
\bottomrule
\end{tabular}
\end{table}

\newpage
\paragraph{Additional Backbone: ViT-Base.}

Beyond the ViT-Tiny backbone used in the main experiments, we evaluate all methods with ViT-Base and verify that the observed trends transfer to larger-scale models in~\Cref{tab:unlearning-derma-cifar-vitbase}.

\begin{table}[h!]
\centering
\caption{Comparison of MUNKEY and studied baselines using ViT-B/16 backbone (85.8M parameters) on a random forget set of 10\% and three seeds on DermaMNIST and CIFAR-10.}
\label{tab:unlearning-derma-cifar-vitbase}
\vspace*{0.2cm}
\setlength{\tabcolsep}{2.4pt}
\fontsize{6}{6.7}\selectfont
\begin{tabular}{@{}l cccc >{\columncolor{lightgray}[0pt][0pt]}c cccc >{\columncolor{lightgray}[0pt][0pt]}c@{}}
\toprule
& \multicolumn{5}{c}{\textbf{DermaMNIST}} & \multicolumn{5}{c}{\textbf{CIFAR-10}} \\
\cmidrule(lr){2-6} \cmidrule(lr){7-11}
\textbf{Model} 
& \textbf{TA} & \textbf{RA} & \textbf{FA} & \textbf{MIA} & \textbf{Avg Gap $\downarrow$}
& \textbf{TA} & \textbf{RA} & \textbf{FA} & \textbf{MIA} & \textbf{Avg Gap $\downarrow$} \\
\midrule
Retrain \text{\tiny (B)} 
& $76.74_{\err{0.20}}$ & $100.0_{\err{0.00}}$ & $75.71_{\err{0.84}}$ & $49.54_{\err{0.22}}$ & -- 
& $75.58_{\err{0.19}}$ & $99.64_{\err{0.02}}$ & $75.22_{\err{0.14}}$ & $50.30_{\err{0.07}}$ & -- \\

Retrain \text{\tiny (M)} 
& $79.25_{\err{0.75}}$ & $99.01_{\err{0.26}}$ & $78.05_{\err{2.12}}$ & $49.75_{\err{0.93}}$ & -- 
& $91.99_{\err{0.06}}$ & $99.95_{\err{0.02}}$ & $92.16_{\err{0.50}}$ & $50.02_{\err{0.49}}$ & -- \\
\midrule

KNN 
& $74.60_{\err{0.24}}$ & $85.54_{\err{0.14}}$ & $74.81_{\err{1.23}}$ & $49.94_{\err{0.90}}$ & --
& $91.06_{\err{0.05}}$ & $94.56_{\err{0.06}}$ & $91.02_{\err{0.13}}$ & $50.16_{\err{0.10}}$ & -- \\

CF 
& $75.99_{\err{0.82}}$ & $98.11_{\err{0.73}}$ & $95.00_{\err{1.85}}$ & $64.09_{\err{1.53}}$ & $9.12_{\err{0.70}}$ 
& $74.26_{\err{0.83}}$ & $95.55_{\err{0.15}}$ & $88.49_{\err{0.58}}$ & $59.22_{\err{0.38}}$ & $6.90_{\err{0.28}}$ \\

CF-k 
& $77.64_{\err{0.52}}$ & $100.0_{\err{0.00}}$ & $100.0_{\err{0.00}}$ & $64.44_{\err{0.47}}$ & $10.02_{\err{0.28}}$ 
& $77.22_{\err{0.30}}$ & $99.70_{\err{0.05}}$ & $99.63_{\err{0.13}}$ & $69.50_{\err{0.23}}$ & $11.33_{\err{0.12}}$ \\

EU-k 
& $77.61_{\err{0.53}}$ & $100.0_{\err{0.00}}$ & $100.0_{\err{0.00}}$ & $62.58_{\err{0.40}}$ & $9.55_{\err{0.28}}$ 
& $77.22_{\err{0.30}}$ & $99.69_{\err{0.05}}$ & $99.63_{\err{0.11}}$ & $69.43_{\err{0.16}}$ & $11.31_{\err{0.11}}$ \\

$\ell_1$-sparse 
& $73.83_{\err{1.04}}$ & $83.35_{\err{1.32}}$ & $77.90_{\err{1.66}}$ & $53.17_{\err{1.02}}$ & $6.35_{\err{0.68}}$ 
& $45.26_{\err{0.56}}$ & $45.63_{\err{0.56}}$ & $45.85_{\err{0.78}}$ & $50.41_{\err{0.89}}$ & $28.45_{\err{0.36}}$ \\

NegGrad+ 
& $74.01_{\err{0.22}}$ & $96.08_{\err{1.65}}$ & $78.71_{\err{1.72}}$ & $53.01_{\err{1.60}}$ & $3.28_{\err{0.75}}$ 
& $72.53_{\err{0.24}}$ & $93.96_{\err{0.40}}$ & $75.71_{\err{0.67}}$ & $52.79_{\err{0.18}}$ & \underline{$2.93_{\err{0.22}}$} \\

Amnesiac 
& $75.56_{\err{0.23}}$ & $98.30_{\err{0.85}}$ & $82.00_{\err{2.38}}$ & $47.01_{\err{1.11}}$ &$2.93_{\err{0.73}}$
& $73.08_{\err{0.05}}$ & $95.11_{\err{0.15}}$ & $79.64_{\err{0.82}}$ & $48.81_{\err{0.21}}$ & $3.24_{\err{0.22}}$ \\

SCRUB 
& $73.58_{\err{0.20}}$ & $80.28_{\err{1.66}}$ & $76.52_{\err{0.54}}$ & $51.97_{\err{1.28}}$ & $6.53_{\err{0.59}}$ 
& $74.48_{\err{0.00}}$ & $97.42_{\err{0.00}}$ & $94.49_{\err{0.00}}$ & $62.69_{\err{0.00}}$ & $8.75_{\err{0.06}}$ \\

SSD 
& $77.42_{\err{0.41}}$ & $100.0_{\err{0.00}}$ & $100.0_{\err{0.00}}$ & $64.37_{\err{0.05}}$ & $9.95_{\err{0.25}}$ 
& $77.23_{\err{0.27}}$ & $99.70_{\err{0.06}}$ & $99.65_{\err{0.10}}$ & $69.55_{\err{0.26}}$ & $11.34_{\err{0.11}}$ \\
AMUN 
& $75.08_{\err{0.72}}$ & $97.95_{\err{0.33}}$ & $92.81_{\err{1.48}}$ & $62.08_{\err{1.33}}$ & $8.34_{\err{0.58}}$ 
& $73.14_{\err{0.46}}$ & $94.63_{\err{0.37}}$ & $86.89_{\err{1.13}}$ & $58.47_{\err{0.49}}$ & $6.82_{\err{0.34}}$ \\
CoUn
& $78.39_{\err{0.76}}$ & $100.0_{\err{0.00}}$ & $82.95_{\err{1.92}}$ & $51.81_{\err{0.96}}$ & \underline{$2.79_{\err{0.61}}$}
& $76.52_{\err{0.77}}$ & $99.74_{\err{0.01}}$ & $85.21_{\err{0.39}}$ & $51.10_{\err{0.28}}$ & $2.96_{\err{0.24}}$ \\
\midrule

\textbf{MUNKEY} 
& $80.15_{\err{0.59}}$ & $99.47_{\err{0.10}}$ & $82.00_{\err{1.74}}$ & $52.79_{\err{0.68}}$ & $\mathbf{2.09}_{\err{0.75}}$ 
& $92.05_{\err{0.18}}$ & $99.92_{\err{0.04}}$ & $95.13_{\err{0.13}}$ & $55.63_{\err{0.17}}$ & $\mathbf{2.17}_{\err{0.15}}$ \\
\bottomrule
\end{tabular}
\end{table}

\subsection{Membership Signals on Model Components}
\label{app:privacy_components}
Throughout the paper and following the literature, we report membership inference attacks on model outputs under the standard MUNKEY inference setting with retrieved exemplar tokens. In this section, we evaluate whether the isolated image backbone retains a membership signal. To this end, we remove the retrieval branch entirely at evaluation time, so that the model processes only image patches without retrieved exemplar tokens. Removing this pathway eliminates the instance-specific information supplied by the memory. As shown in Table~\ref{tab:privacy_components}, this substantially reduces predictive performance, confirming that the exemplar pathway is actively used by the model. Nevertheless, performance remains above chance, indicating that the image backbone retains meaningful global structure and making this setting a useful stress test for membership signal in the backbone alone. We note that this no-retrieval evaluation also changes the evaluation distribution relative to standard inference, and should therefore be interpreted as a controlled stress test rather than a distribution-preserving deployment setting. We compare MUNKEY against the corresponding Retrain (M) oracle under the same no-retrieval protocol, and include the standard retrieval-based results for reference. The MIA scores remain close to random guessing, indicating that the isolated backbone does not retain a measurable membership signal for forgotten samples.\looseness-1
\begin{table}[h!]
\centering
\caption{Membership inference on isolated model subcomponents: Standard MUNKEY and removing retrieval from the memory bank (w/o $\mathcal{M}$) across three seeds.}
\label{tab:privacy_components}
\vspace*{0.2cm}
\footnotesize
\setlength{\tabcolsep}{4.5pt}
\begin{tabular}{l l c c c >{\columncolor{lightgray}[3pt][3pt]}c}
\toprule
\textbf{Dataset} & \textbf{Model} & \textbf{TA} & \textbf{RA} & \textbf{FA} & \textbf{MIA} \\
\midrule

\multirow{4}{*}{DermaMNIST}
& Retrain \text{\tiny (M)}
& $81.25_{\pm \text{\tiny 0.66}}$
& $99.70_{\pm \text{\tiny 0.08}}$
& $80.29_{\pm \text{\tiny 1.41}}$
& $50.04_{\pm \text{\tiny 1.05}}$ \\

& Retrain \text{\tiny (M)} w/o $\mathcal{M}$
& $33.25_{\pm \text{\tiny 6.73}}$
& $35.41_{\pm \text{\tiny 6.96}}$
& $33.57_{\pm \text{\tiny 5.94}}$
& $50.23_{\pm \text{\tiny 1.68}}$ \\
\cmidrule(lr){2-6}

& \textbf{MUNKEY}
& $81.06_{\pm \text{\tiny 0.44}}$
& $99.71_{\pm \text{\tiny 0.08}}$
& $84.00_{\pm \text{\tiny 1.22}}$
& $51.19_{\pm \text{\tiny 1.19}}$ \\

& \textbf{MUNKEY} w/o $\mathcal{M}$
& $37.68_{\pm \text{\tiny 8.40}}$
& $39.34_{\pm \text{\tiny 8.20}}$
& $38.29_{\pm \text{\tiny 8.52}}$
& $50.81_{\pm \text{\tiny 0.51}}$ \\
\midrule

\multirow{4}{*}{CIFAR-10}
& Retrain \text{\tiny (M)}
& $92.38_{\pm \text{\tiny 0.06}}$
& $99.93_{\pm \text{\tiny 0.03}}$
& $92.05_{\pm \text{\tiny 0.23}}$
& $49.90_{\pm \text{\tiny 0.34}}$ \\

& Retrain \text{\tiny (M)} w/o $\mathcal{M}$
& $34.43_{\pm \text{\tiny 7.37}}$
& $35.84_{\pm \text{\tiny 7.76}}$
& $34.74_{\pm \text{\tiny 7.74}}$
& $50.26_{\pm \text{\tiny 0.21}}$ \\
\cmidrule(lr){2-6}

& \textbf{MUNKEY}
& $92.63_{\pm \text{\tiny 0.20}}$
& $99.94_{\pm \text{\tiny 0.01}}$
& $92.53_{\pm \text{\tiny 0.52}}$
& $51.40_{\pm \text{\tiny 0.44}}$ \\

& \textbf{MUNKEY} w/o $\mathcal{M}$
& $39.12_{\pm \text{\tiny 9.28}}$
& $40.89_{\pm \text{\tiny 9.52}}$
& $39.55_{\pm \text{\tiny 9.55}}$
& $51.03_{\pm \text{\tiny 0.42}}$ \\

\bottomrule
\end{tabular}
\end{table}

\subsection{Robustness to Neighbor Deletion}

\label{app:neighbor-deletion}

We further investigate whether deleting a key affects the predictions of similar retained samples. While removing a key does not prevent retrieving $K$ neighbors, as the next nearest entries are selected instead, it may affect the predictions of samples that previously relied on the deleted entry. To quantify this effect, we evaluate CIFAR-10 and DermaMNIST under 10\% random forgetting. For each retained sample, we identify forgotten keys within its top-$4$ neighborhood and compare its accuracy before and after their removal, excluding the sample's own key in both cases to prevent self-retrieval. We group samples by the fraction of their original retrieval weight assigned to the deleted neighbors, as reported in Table~\ref{tab:neighbor-deletion}.  MUNKEY's training mechanisms encourage localized information storage and reduce dependence on individual neighboring tokens, potentially limiting interference from key deletion. We see how retained samples with no deleted neighbors remain unaffected,  which we include as a reference, while those losing up to half their retrieval weight show accuracy changes of at most 2.83 percentage points. 
As an additional stress test, deleting each sample's nearest neighbor reduces accuracy by 1.50 and 3.00 percentage points on CIFAR-10 and DermaMNIST, respectively. These results suggest that MUNKEY is robust to neighboring key deletions, limiting unlearning interference with retained samples.

\begin{table}[h!]
\centering
\caption{Effect of neighbor deletion on retained samples. Accuracy before and after removing forgotten keys from the top-$4$ neighborhood, grouped by the fraction of retrieval weight removed.}
\label{tab:neighbor-deletion}
\vspace*{0.2cm}
\footnotesize
\setlength{\tabcolsep}{4.5pt}
\begin{tabular}{l l c c >{\columncolor{lightgray}[3pt][3pt]}c}
\toprule
\textbf{Dataset} & \textbf{Deleted retrieval mass} & \textbf{Acc. before (\%)} & \textbf{Acc. after (\%)} & \textbf{$\Delta$} \\
\midrule

\multirow{3}{*}{DermaMNIST}
& No deleted neighbor
& $83.69_{\pm \text{\tiny 0.71}}$
& $83.69_{\pm \text{\tiny 0.71}}$
& $0.00$ \\

& 0--25\%
& $83.97_{\pm \text{\tiny 1.43}}$
& $84.58_{\pm \text{\tiny 1.41}}$
& $+0.61$ \\

& 25--50\%
& $85.09_{\pm \text{\tiny 1.55}}$
& $82.26_{\pm \text{\tiny 1.66}}$
& $-2.83$ \\

\midrule

\multirow{3}{*}{CIFAR-10}
& No deleted neighbor
& $93.65_{\pm \text{\tiny 0.48}}$
& $93.65_{\pm \text{\tiny 0.48}}$
& $0.00$ \\

& 0--25\%
& $92.77_{\pm \text{\tiny 0.98}}$
& $92.62_{\pm \text{\tiny 0.98}}$
& $-0.14$ \\

& 25--50\%
& $95.10_{\pm \text{\tiny 0.94}}$
& $95.29_{\pm \text{\tiny 0.92}}$
& $+0.19$ \\

\bottomrule
\end{tabular}
\end{table}

\subsection{Retrieval-Augmented Study}
\label{app:retrieval-fusion}

We further investigate whether an editable retrieval component alone is sufficient for unlearning by comparing different retrieval-based approaches. In addition to the existing baselines, we implement a RAC-inspired retrieval-fusion variant~\citep{long2022retrieval}, combining a normally trained ViT classifier with an editable memory of frozen key-label pairs. Classifier logits $\mathbf{s}_{\mathrm{cls}}$ and retrieval probabilities $\mathbf{p}_{\mathrm{ret}}$ are fused as $\mathbf{s}_{\mathrm{cls}} + \alpha \log \mathbf{p}_{\mathrm{ret}}$, where $\alpha > 0$ is learned jointly with the classifier. We use the same frozen key encoder, optimizer, and training schedule as MUNKEY, but without its deletion-oriented training mechanisms. Unlearning consists of removing memory entries without updating the model parameters.\looseness-1

Table~\ref{tab:retrieval-fusion} compares the RAC-inspired baseline, Direct Key, KNN, and MUNKEY on DermaMNIST under 10\% random forgetting. Each trainable method is evaluated against its corresponding retain-only retraining oracle over three seeds. While KNN enables deletion through purely non-parametric retrieval, its predictive utility is lower. Both RAC-inspired retrieval fusion and Direct Key retain substantial membership signals after deletion, whereas MUNKEY remains considerably closer to its retraining oracle. These results highlight the importance of deletion-oriented training for encouraging localized information storage.
\begin{table}[h!]
\centering
\caption{Comparison of retrieval-augmented approaches on DermaMNIST under 10\% random forgetting. Trainable methods are compared against their corresponding retraining oracles across three seeds.\looseness-1}
\label{tab:retrieval-fusion}
\vspace*{0.2cm}
\footnotesize
\setlength{\tabcolsep}{4.5pt}
\begin{tabular}{l c c c c >{\columncolor{lightgray}[3pt][3pt]}c}
\toprule
\textbf{Model} & \textbf{TA} & \textbf{RA} & \textbf{FA} & \textbf{MIA} & \textbf{Avg Gap} \\
\midrule

RAC-inspired
& $76.28_{\pm \text{\tiny 0.22}}$
& $100.00_{\pm \text{\tiny 0.00}}$
& $100.00_{\pm \text{\tiny 0.00}}$
& $61.51_{\pm \text{\tiny 0.40}}$
& $9.29_{\pm \text{\tiny 0.33}}$ \\

Retrain \text{\tiny (RAC)}
& $76.77_{\pm \text{\tiny 0.48}}$
& $100.00_{\pm \text{\tiny 0.00}}$
& $75.52_{\pm \text{\tiny 1.09}}$
& $49.30_{\pm \text{\tiny 1.40}}$
& -- \\
\midrule

Direct Key
& $81.70_{\pm \text{\tiny 0.74}}$
& $99.30_{\pm \text{\tiny 0.45}}$
& $99.33_{\pm \text{\tiny 0.64}}$
& $61.76_{\pm \text{\tiny 2.00}}$
& $8.23_{\pm \text{\tiny 0.70}}$ \\

Retrain \text{\tiny (Direct Key)}
& $81.83_{\pm \text{\tiny 0.68}}$
& $99.25_{\pm \text{\tiny 0.22}}$
& $80.05_{\pm \text{\tiny 1.10}}$
& $48.30_{\pm \text{\tiny 0.98}}$
& -- \\
\midrule

KNN
& $74.60_{\pm \text{\tiny 0.24}}$
& $85.54_{\pm \text{\tiny 0.14}}$
& $74.81_{\pm \text{\tiny 1.23}}$
& $49.94_{\pm \text{\tiny 0.90}}$
& -- \\
\midrule

\textbf{MUNKEY}
& $81.06_{\pm \text{\tiny 0.41}}$
& $97.64_{\pm \text{\tiny 0.22}}$
& $84.10_{\pm \text{\tiny 1.58}}$
& $52.16_{\pm \text{\tiny 0.53}}$
& $\mathbf{1.52}_{\pm \text{\tiny 0.56}}$ \\

Retrain \text{\tiny (M)}
& $80.80_{\pm \text{\tiny 0.54}}$
& $97.47_{\pm \text{\tiny 0.15}}$
& $80.10_{\pm \text{\tiny 0.99}}$
& $50.53_{\pm \text{\tiny 0.89}}$
& -- \\

\bottomrule
\end{tabular}
\end{table}

\newpage
\section{Sensitivity Score and Model Selection}
\label{app:sensitivity}

\begin{figure}[h!]
    \centering
    \includegraphics[width=.9\linewidth]{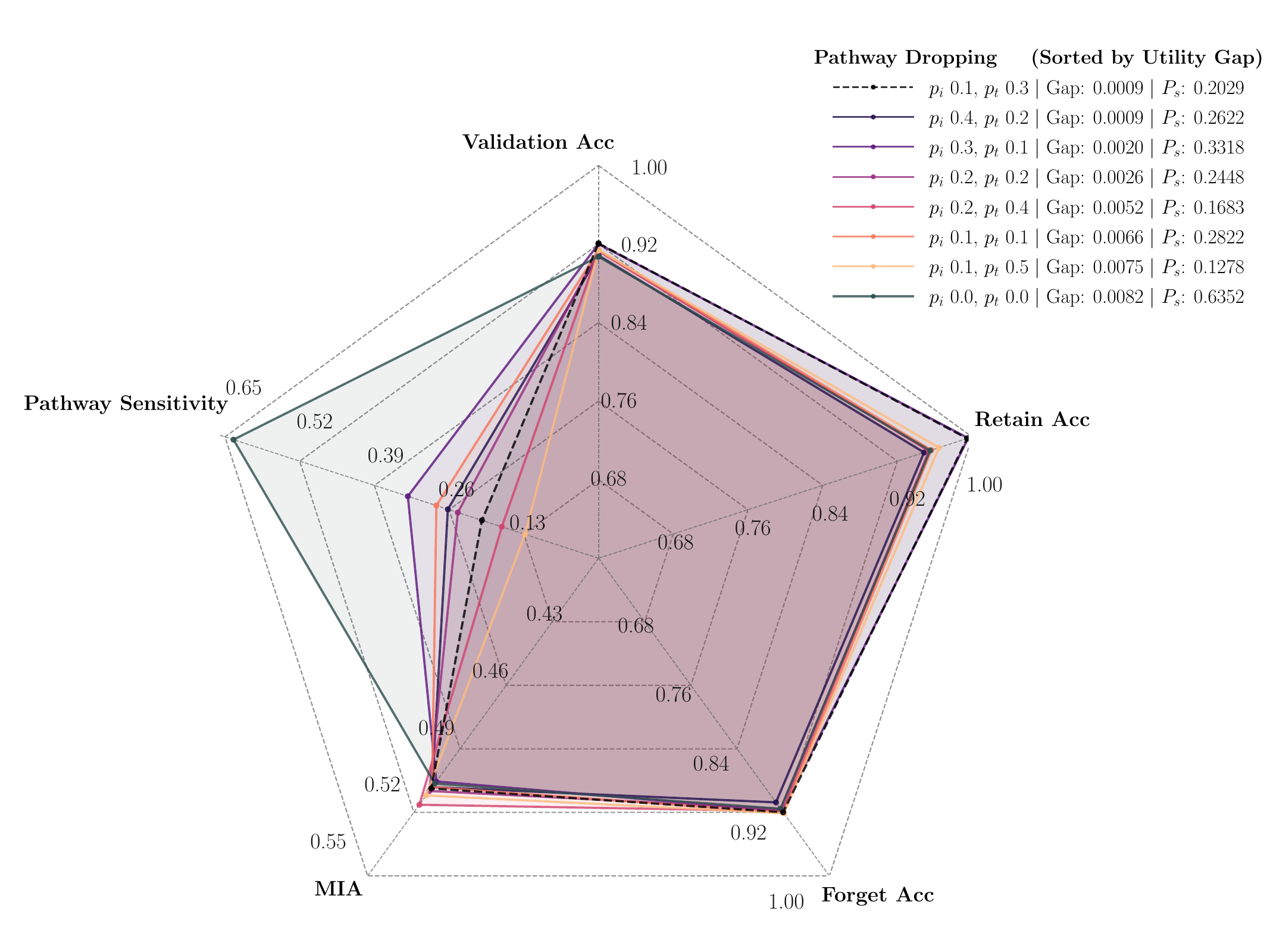}
    \caption{Overview of the performance metrics under varying image and token dropout probabilities in MUNKEY training in CIFAR-10 for visualization of model selection.}
    \label{fig:radarcifar10}
\end{figure}

\begin{figure}[h!]
    \centering
    \includegraphics[width=.9\linewidth]{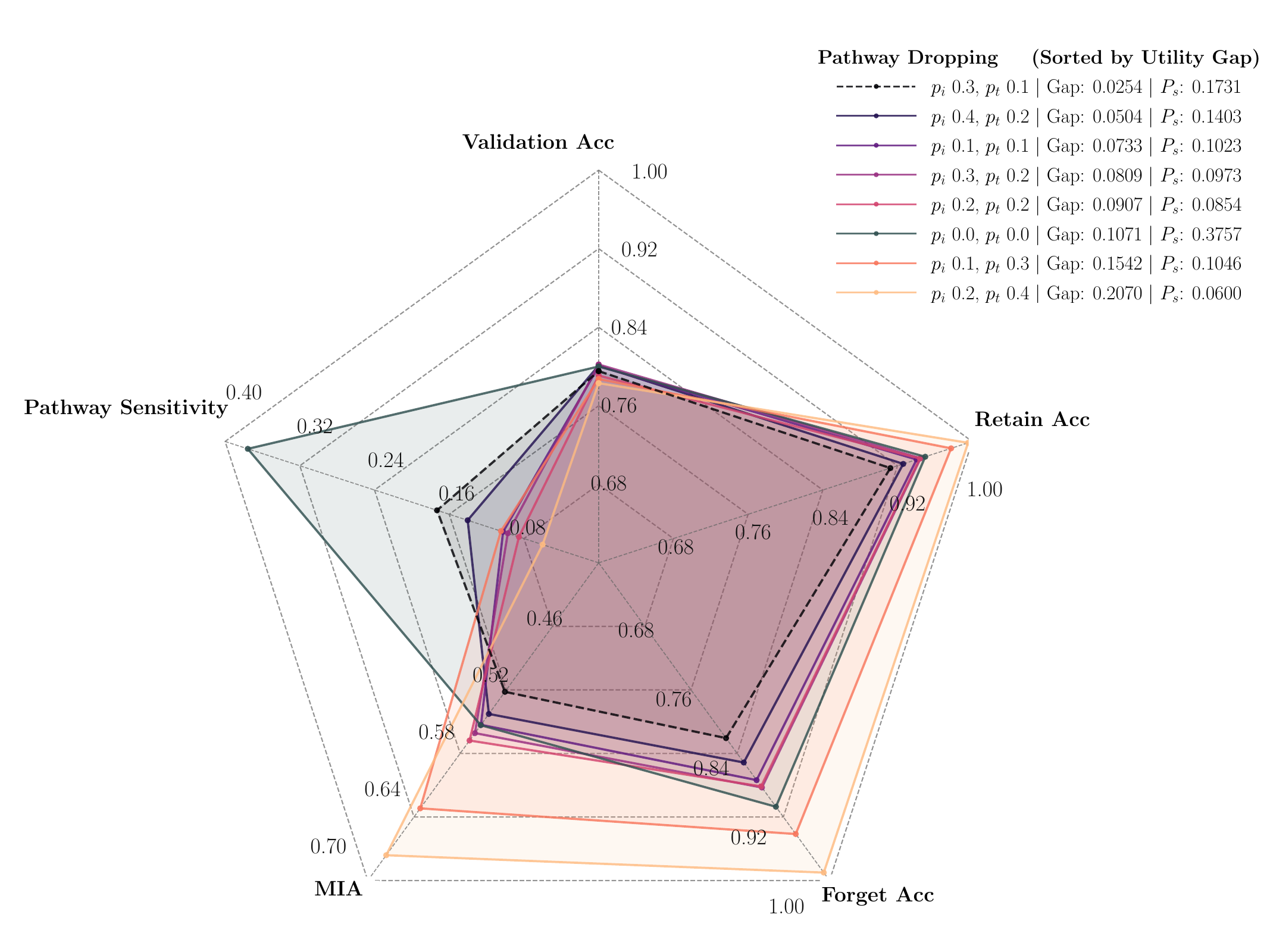}
    \caption{Overview of the resulting performance metrics under varying image and token dropout probabilities in MUNKEY training in DermaMNIST for visualization of model selection.}
    \label{fig:radar-dermamnist}
\end{figure}

As detailed in~\Cref{sec:methods}, we introduce the Pathway Sensitivity Score ($P_s$) as a quantitative measure to assess the informational balance between the image and token pathways in MUNKEY. This metric is used in our model selection process, particularly to determine the stochastic dropout probabilities $p_i$ and $p_t$ used in training, which are tuned to account for dataset-specific characteristics. In certain datasets, the model may tend to ``shortcut'' information by over-relying on either the raw image patches or the exemplar tokens, a behavior that risks pathway collapse and can hinder unlearning and generalization. With this regularization strategy, we aim to push the decoupling of global image features from instance-specific memories. To ensure a robust configuration, we first implement a stability constraint by filtering out models where $P_s \ge \xi$, with $\xi = 0.3$ serving as a global threshold to prevent disproportionate reliance on a single modality. This accounts for the inherent asymmetry between image patches and optimized tokens while ensuring that neither pathway becomes a ``dead branch'' during training. After this filtering, we apply a selection strategy following the ``rewind'' procedure in~\citet{kurmanji2023towards}, prioritizing models with the smallest validation utility gap, i.e., the difference between validation accuracy and disjoint proxy forget accuracy. This proxy forget split is sampled only for model selection and is disjoint from the evaluation forget set used in all reported unlearning results. In the event of comparable utility gaps, we select the model with the lower $P_s$ score to favor configurations that maintain a more balanced modality equilibrium.~\Cref{fig:radarcifar10} and~\Cref{fig:radar-dermamnist} illustrate these dynamics across various parameter sets in CIFAR-10 and DermaMNIST, with their corresponding utility gaps and $P_s$ in the legend. In CIFAR-10, lower token dropout generally leads to higher pathway divergence, as the model defaults to over-relying on the token pathway. Moreover, the two best parameter configurations have comparable gaps, and we rely on $P_s$ to select the optimal set. Conversely, DermaMNIST displays a wider diversity of dropout patterns, where the best configuration is identified directly through the utility gap. In both benchmarks, the best-performing parameter set is marked with a dashed black line, while the baseline corner case ($p_i=0, p_t=0$) is shown in gray. These results highlight that without the introduced $p_i > 0$ and $p_t > 0$, pathway sensitivity is strongly degraded and utility gaps remain sub-optimal, motivating the necessity of our pathway dropout approach.

\end{document}